\documentclass[runningheads]{llncs}
\usepackage{eccvabbrv}

\usepackage{graphicx}
\usepackage{booktabs}
\usepackage{subcaption}
\usepackage{wrapfig}
\usepackage[normalem]{ulem}

\usepackage{longtable}
\usepackage{array}
\usepackage{tcolorbox}
\usepackage[ruled,vlined]{algorithm2e}
\usepackage{listings}
\usepackage[accsupp]{axessibility}  % Improves PDF readability for those with disabilities.

\usepackage[hypertexnames=false]{hyperref}

\usepackage{orcidlink}

\begin{document}

\title{From Plans to Pixels: Learning to Plan and Orchestrate for Open-Ended Image Editing} 
%\title{From Plans to Pixels: Experiential Planning and Orchestration for Open-Ended Image Editing}

% Experiential Planning and Tool Orchestration for Long-Horizon Image Editing
% Learning to Plan and Orchestrate for Long-Horizon, Open-Ended Image Editing
% An Experiential Learning Framework for Long-Horizon Image Editing
% From Plans to Pixels: Experiential Planning and Tool Orchestration for Open-Ended Image Editing
% Beyond Single-Step Editing: Planning and Outcome-Driven Tool Selection for Open-Ended Image Transformation
% Plan, Execute, Refine: Experiential Learning for Long-Horizon Image Editing

\titlerunning{Learning to Plan and Orchestrate for Open-Ended Image Editing}
% \orcidlink{0000-1111-2222-3333}\orcidlink{1111-2222-3333-4444}\orcidlink{2222--3333-4444-5555}
\author{
Anirudh Sundara Rajan\inst{1} \and
Krishna Kumar Singh\inst{2} \and
Yong Jae Lee\inst{2}
}

\authorrunning{A.~S.~Rajan et al.}

\institute{
University of Wisconsin--Madison, WI, USA\\
\email{asundararaj2@wisc.edu}
\and
Adobe Research, San Jose, CA, USA\\
\email{\{krishin, yongl\}@adobe.com}
}
\maketitle

\begin{abstract}

Modern image editing models produce realistic results but struggle with abstract, multi-step instructions (e.g., ``make this advertisement more vegetarian-friendly''). Prior agent-based methods decompose such tasks but rely on handcrafted pipelines or teacher-imitation, limiting flexibility and decoupling learning from actual editing outcomes. We propose an \emph{experiential framework for abstract, long-horizon image editing}, where a planner generates structured atomic decompositions and an orchestrator selects tools and regions to execute each step. A vision–language judge provides outcome-based rewards for instruction adherence and visual quality. The orchestrator is trained to maximize these rewards, and successful trajectories are used to refine the planner. By tightly coupling planning with reward-driven execution, our approach yields more coherent and reliable edits than single-step or rule-based multi-step baselines. \noindent\textbf{Project Page:} \url{https://anisundar18.github.io/Plan2Pix.github.io/}

\end{abstract}

\section{Introduction}
\label{sec:intro}

%Recent advances in diffusion-based image editing have significantly improved the fidelity and controllability of instruction-based visual modifications. Methods such as InstructPix2Pix~\cite{brooks2023instructpix2pix}, Prompt-to-Prompt~\cite{hertz2022prompt}, and more recent large-scale instruction-tuned editors (e.g., Flux Kontext~\cite{labs2025flux1kontextflowmatching}, Qwen-Image-Edit~\cite{wu2025qwenimagetechnicalreport}) demonstrate impressive performance on localized and well-specified edits, such as \emph{“add a hat to the man”} or \emph{“change the car color to red.”} These systems excel when the instruction maps directly to a single, visually grounded transformation.

Recent advances in diffusion-based image editing have significantly improved the fidelity and controllability of instruction-based visual modifications. Methods such as InstructPix2Pix~\cite{brooks2023instructpix2pix}, Prompt-to-Prompt~\cite{hertz2022prompt}, and large-scale editors like Flux Kontext~\cite{labs2025flux1kontextflowmatching} and Qwen-Image-Edit~\cite{wu2025qwenimagetechnicalreport} perform well on well-specified edits (e.g., \emph{``add a hat to the man''}, \emph{``change the car color to red''}), where the instruction corresponds to a simple concrete transformation.

\begin{figure*}[t]
\centering
\includegraphics[width=\linewidth]{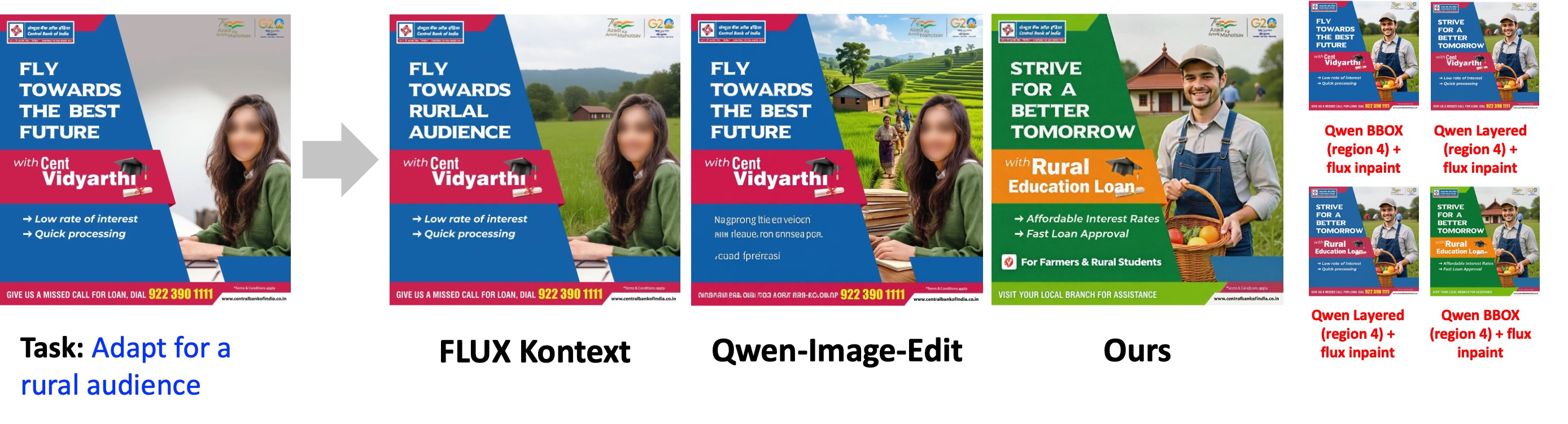}
\caption{Given a high-level instruction such as \emph{“Adapt for a rural audience,”} single-step editors (e.g., Flux Kontext~\cite{labs2025flux1kontextflowmatching} and Qwen-Image-Edit~\cite{wu2025qwenimagetechnicalreport}) struggle to jointly adapt visual themes, textual content, and audience-specific context while preserving the original advertisement layout and identity. 
In contrast, our framework decomposes the task into structured subtasks and orchestrates multiple tools using outcome-based feedback. 
The four images to the right of our result illustrate intermediate edits produced by automatically selected tool--region combinations during orchestration (corresponding subtasks are: (1) \emph{Replace the background with a rural scene featuring a farmer holding a basket of produce, while keeping the existing text elements intact}; (2) \emph{Replace the text `FLY TOWARDS THE BEST FUTURE' with `STRIVE FOR A BETTER TOMORROW'}; (3) \emph{Replace `Cent Vidhyarthi' with `Rural Education Loan'}; and (4) \emph{Add a village house or temple in the background to reinforce the rural setting}).}
\label{fig:teaser}
\end{figure*}

However, many real-world editing tasks are abstract, open-ended, and long-horizon. For example, adapting a student-focused loan advertisement into a campaign targeting rural audiences (Fig.~\ref{fig:teaser}) requires coordinated changes to imagery, slogans, audience-specific messaging, and environmental context—far beyond a single atomic edit. Different subtasks may also require different tools (e.g., object replacement vs.\ text modification). Prior agent-based systems attempt multi-step orchestration but often rely on handcrafted pipelines or teacher-imitation~\cite{yang2024mastering,wang2024div,ji2025iccv,yeh2025beyond}, fixing execution order and heuristics. These approaches do not train the planner on its own distribution and do not optimize tool selection based on actual editing outcomes, which can lead to distribution shift, limited generalization, and poor scalability to open-ended instructions.

To address these limitations, we decouple long-horizon image editing into \emph{planning} and \emph{orchestration}. Given a high-level abstract instruction, the planner produces a checklist-guided decomposition into atomic subtasks and is trained on its own sampled plans to reduce distribution shift and improving stability relative to teacher imitation. Conditioned on the plan, the orchestrator selects tools and regions, executes edits, and receives outcome-based feedback from a VLM judge evaluating instruction adherence, identity preservation, and visual quality. These rewards directly supervise tool selection, grounding decisions in empirical performance. A refinement stage prunes infeasible subtasks, aligning plans with executable actions. Together, this forms an experiential learning framework that improves through interaction with editing tools and judged outcomes.

Training this system, however, poses challenges beyond standard supervision: there is no large-scale dataset of abstract multi-step plans, tool selection is context-dependent and ambiguous, and multiple edited outputs can validly satisfy the same instruction. In addition, invoking modern image editing tools is computationally expensive making exploration intractable. These factors make fixed-label standard supervised training challenging. We therefore adopt an experiential learning paradigm grounded in observed editing outcomes. To keep training tractable, we approximate trajectory reward as the sum of independently evaluated sub-task rewards, enabling precomputation over tool–region pairs. The planner learns structured decompositions via checklist-guided self-supervision, while the orchestrator learns tool and region selection directly from judged edits rather than prompts or teacher traces. This design removes handcrafted rules, aligns training with inference, and improves generalization to open-ended instructions.

%\kr{Training this system poses challenges not addressed by standard supervised pipelines: there is no large scale dataset with multi step decompositions or definitive target images, tool performance and region choice are context dependent and ambiguous, and invoking modern editors is computationally costly. We adopt an experiential learning paradigm that provides efficient credit assignment while grounding both components in observed editing outcomes. To keep training tractable, we approximate the reward of a full editing trajectory by summing rewards for individual sub tasks applied independently to the input image, which allows precomputation of sub task rewards across candidate tools and regions. The planner learns structured and complete decompositions through checklist based supervision while remaining on policy by imitating its own sampled plans, and the orchestrator learns to jointly select tools and regions from judged edited outcomes rather than from prompts or teacher traces. These choices remove reliance on handcrafted rules, align training with inference, and improve generalization to open ended instructions.}

%on a newly curated advertisement editing benchmark (based on images from~\cite{sagar2024madverse}) and the G-Edit image editing benchmark~\cite{liu2025step1x-edit}
Extensive experiments demonstrate that our framework produces more reliable, coherent, and instruction-faithful results than both single-step generation approaches and multi-step agent baselines. Our key contributions are:
%rule-based 
%\noindent
\begin{itemize}

\item \textbf{Long-horizon, high-level image editing framework.} 
We cast abstract, open-ended editing as a coordinated planning-and-orchestration problem, enabling multi-step reasoning beyond single-step generation.

\item \textbf{Self-Supervised checklist-guided plan generation.} 
A structured planner learns multi-step decompositions from its own checklist-guided samples, reducing distribution shift.

\item \textbf{Experiential orchestrator.} 
A reward-driven policy jointly selects tools and regions based on judged executed edits, grounding decisions in empirical outcomes rather than handcrafted rules.

\item \textbf{Closed-loop refinement and strong results.} 
We prune infeasible sub-tasks using orchestration feedback and achieve state-of-the-art performance for open-ended image editing.

\end{itemize}

% \end{itemize}
% \begin{itemize}
% \yj{I think we need to emphasize the long-term, high-level instructions aspect.}

%   \item \textbf{On-policy planner:} Learns from its own sampled, checklist-guided decompositions.
%   \item \textbf{Experiential orchestrator:} Learns tool \emph{and} region selection directly from judged real edits.
%   \item \textbf{Planner refinement:} Removes infeasible subplans using orchestrator feasibility feedback.
%   \item \textbf{State-of-the-art performance:} Achieves superior instruction adherence and visual quality on both the advertisement editing benchmark and G-Edit.
  
% \end{itemize}

\section{Related Work}

% \paragraph{Controllable Image Editing with Diffusion Models.}
% Recent diffusion-based models have demonstrated remarkable performance in text-guided image generation and editing~\cite{rombach2022high,podell2023sdxl}. 
% Early training-free approaches, such as SDEdit~\cite{meng2021sdedit}, Prompt-to-Prompt~\cite{hertz2022prompt}, and related methods~\cite{parmar2023zero,cao2023masactrl,hertz2024style}, manipulate the denoising process to align an input image with modified textual prompts. 
% While flexible, these approaches are typically limited to relatively direct and localized edits and may suffer from over-editing or insufficient instruction faithfulness.  Training-based approaches, including InstructPix2Pix~\cite{brooks2023instructpix2pix} and MagicBrush~\cite{zhang2024magicbrush}, improve robustness by fine-tuning diffusion models on paired before–after data. 
% Subsequent methods incorporate additional control signals—such as segmentation masks, bounding boxes, or drag-based interactions—to improve spatial precision and controllability~\cite{li2023gligen,wang2024instancediffusion,mou2023dragondiffusion,shi2024dragdiffusion,nie2024compositional}. 
% However, these systems typically assume well-specified, low-level instructions (e.g., “add a hat,” “change color to blue”) and often require manual specification of control inputs. 
% In contrast, we focus on \emph{abstract, open-ended} instructions that require multi-step reasoning and coordinated use of heterogeneous editing tools.

\paragraph{Controllable Image Editing with Diffusion Models.}
Diffusion-based models have achieved strong performance in text-guided image editing~\cite{rombach2022high,podell2023sdxl}. Training-free methods such as SDEdit and Prompt-to-Prompt~\cite{meng2021sdedit,hertz2022prompt,parmar2023zero,cao2023masactrl,hertz2024style} manipulate the denoising process for prompt-aligned edits, but are typically limited to localized changes and may over-edit or under-follow instructions. Training-based approaches, including InstructPix2Pix and MagicBrush~\cite{brooks2023instructpix2pix,zhang2024magicbrush}, improve robustness via paired supervision. Later methods add control signals (e.g., masks, boxes, drag-based inputs) to enhance spatial precision~\cite{li2023gligen,wang2024instancediffusion,mou2023dragondiffusion,shi2024dragdiffusion,nie2024compositional}. However, these systems assume well-specified, low-level instructions and often require manual controls. In contrast, we target abstract, open-ended instructions requiring multi-step reasoning and coordinated tool use.

% \paragraph{Multimodal LLMs for Image Editing and Planning.}
% Multimodal large language models (MLLMs) extend LLMs with visual inputs, enabling joint reasoning over text and images~\cite{liu2023visual,zhu2023minigpt,liu2024improved}. 
% Recent works have explored leveraging MLLMs to assist image editing. 
% For example, MGIE~\cite{fu2023guiding} refines editing instructions using an MLLM before passing them to a diffusion editor, while other systems employ external LLM agents to decompose complex editing requests into simpler instructions.  These approaches primarily use MLLMs as instruction rewriters or high-level decomposers, with execution handled by a fixed editing model. 
% In contrast, our framework tightly integrates planning and execution via experiential learning: the planner generates a structured multi-step plan, and an orchestrator learns a policy over multiple editing tools and spatial regions. 
%Moreover, our planner is trained via checklist-guided self-training and operates independently at inference time, without reliance on closed-source external agents.

\vspace{-5pt}

\paragraph{Multimodal LLMs for Image Editing and Planning.}
In vision, recent work generates code to invoke specialized modules, decomposing tasks into tool-executable subproblems~\cite{gupta2023visual,suris2023vipergpt,hu2024visual,vista2026huang}. 
These systems treat pretrained models as callable tools and use LLMs to orchestrate their composition for complex visual reasoning. 
Building on this paradigm of task decomposition and tool invocation, multimodal LLMs (MLLMs) extend language models with visual inputs for joint text–image reasoning~\cite{liu2023visual,zhu2023minigpt,liu2024improved}, and have recently been applied to image editing. 
For example, MGIE~\cite{fu2023guiding} rewrites instructions before passing them to a diffusion editor, while other systems use VLM agents to decompose complex editing requests into simpler steps executed by a fixed editor~\cite{yang2024mastering,wang2024div,ji2025iccv,yeh2025beyond}. These approaches are typically training-free or rely on imitation of teacher plans, and do not learn from the outcomes of real edits—planners are not trained on their own plan distributions, and tool selection is not policy-optimized. 
In contrast, our framework couples checklist-guided planning with experiential orchestration, learning tool and region selection directly from judged editing outcomes.

\vspace{-5pt}
\paragraph{Experiential Learning for Long-Horizon Reasoning.}
Reinforcement learning (RL) has recently been used to enhance long-horizon reasoning in language models, enabling step decomposition, iterative refinement, and improved robustness~\cite{openai2024learning,guo2025deepseek,wei2022chain}. 
Several works extend such ideas to multimodal reasoning by training models to generate chain-of-thought explanations grounded in visual inputs~\cite{liu2025visual,huang2025vision}. While these approaches primarily refine the reasoning model itself for end-to-end prediction, we adopt a complementary perspective. 
Instead of modifying the internal reasoning dynamics of a single editor, we learn a policy that selects among multiple editing tools and spatial regions to maximize a reward signal from a learned judge. 
Furthermore, because diffusion-based editors are computationally intensive, direct online RL over full trajectories is impractical. 
We therefore introduce structured reward approximations that enable tractable policy optimization while preserving meaningful credit assignment.

% \vspace{10pt}
% In summary, prior work on image editing largely
% focuses on single-step, directly specified transformations, or on planning without learned execution policies. 
% Our framework addresses a distinct setting: long-horizon, abstract image editing requiring coordinated planning and reward-driven orchestration over heterogeneous tools. 
% By combining checklist-guided plan generation with efficient policy optimization, we move beyond handcrafted pipelines and enable scalable, open-ended visual editing.

\section{Approach}

%\subsection{Overview}

We propose an \emph{experiential learning framework for long-horizon, open-ended image editing}. Abstract editing tasks require both high-level reasoning and low-level tool execution, which we learn through interaction with editing tools and feedback from a learned judge.

Given an input image $x$ and instruction $I$, our goal is to produce an edited image $\hat{x}$ which fulfills the instruction while maintaining high visual quality and preserving essential details from the original image. We decompose this into two stages: a \textbf{Planner} that generates a structured sequence of sub-tasks, and an \textbf{Orchestrator} that selects tools and/or regions to execute each step. Training is guided by rewards from an MLLM-based judge evaluating correctness, visual quality, and consistency with the original image.

This design is motivated by two observations: abstract instructions require multi-step, heterogeneous operations, and direct end-to-end optimization over full editing trajectories is computationally expensive. We address both via structured decomposition and efficient reward approximation.

\subsection{Stage 1: Planner via Checklist-Guided Self-Training}\label{subsec:planner}

%Given an input image $x$ and a high-level instruction $I$, the planner (implemented as a multimodal LLM) generates an ordered sequence of $T$ sub-tasks: $\mathcal{P} = \{s_1, s_2, \dots, s_T\},$ where each $s_t$ is a structured editing instruction (e.g., ``replace beef patty with plant-based alternative,'' ``update slogan to vegetarian-friendly message,'' or ``adjust color palette to green tones''). 
%This decomposition transforms an abstract, open-ended objective into a sequence of concrete, executable atomic steps. 
%Such explicit factorization reduces the cognitive burden on downstream execution, promotes interpretability, and enables modular reasoning over heterogeneous editing operations.

Given an input image $x$ and high-level instruction $I$, the planner (a multimodal LLM) generates an ordered sequence of sub-tasks $\mathcal{P}=\{s_1,\dots,s_T\}$, where each $s_t$ is a structured editing step (e.g., ``add a laptop and organized business supplies to the bedside table,''). This decomposition converts an abstract objective into executable atomic operations, enabling modular reasoning and interpretable multi-step editing.

\begin{figure}[t]
\centering
\includegraphics[width=\linewidth]{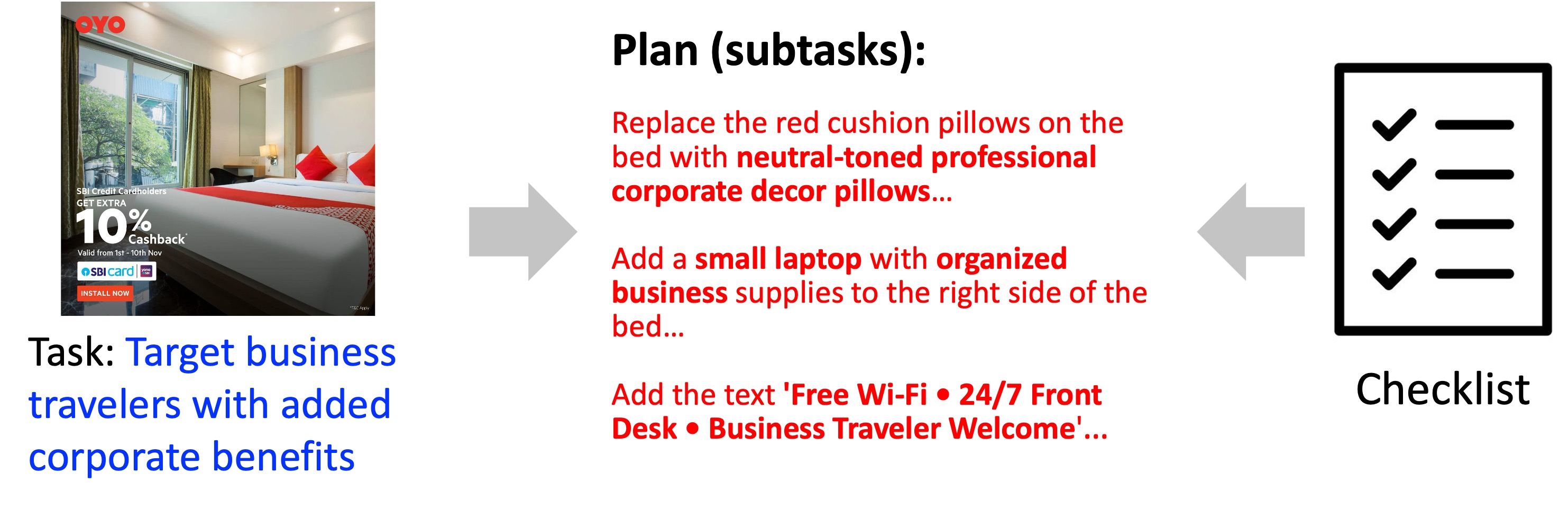}
\caption{\textbf{Checklist-guided planner (Stage 1).} Given an input advertisement and a high-level instruction (left, e.g., ``Target business travelers with added corporate benefits''), the planner generates a structured, ordered sequence of subtasks (center) that explicitly address checklist items (right). In this example, the generated plan includes adapting the room aesthetics for a professional audience, adding work-related objects, and introducing business-oriented promotional text. Checklists can be provided by human or LLM annotators, helping enforce coverage and interpretability while enabling the planner to produce modular plans for downstream orchestration.}
\label{fig:planner-checklist}
\end{figure}

\vspace{-5pt}
\paragraph{Checklist Construction.}
Rather than imitating a teacher model~\cite{yeh2025beyond}, we introduce a \emph{checklist} $\mathcal{C}=\{c_1,\dots,c_K\}$ specifying criteria a satisfactory edit must meet (e.g., product substitution, semantic alignment, layout coherence). During data construction, the planner is prompted with $(x,I,\mathcal{C})$ to generate plans that explicitly satisfy all checklist items (Fig.~\ref{fig:planner-checklist}). Unlike loosely related prior checklist-based reward alignment for LLMs~\cite{checklists-are-better}, we use checklists for structured plan generation for long-horizon image editing.

%Instead of directly imitating plans produced by a separate teacher model as in prior work (e.g.,~\cite{yeh2025beyond}), we introduce a \emph{checklist} $\mathcal{C} = \{c_1, \dots, c_K\}$, which enumerates criteria that a satisfactory edit should satisfy (e.g., product substitution accuracy, semantic alignment of text, cultural consistency, and layout coherence). 
%During data construction, the planner is prompted with $(x, I, \mathcal{C})$ and required to generate a plan that explicitly addresses all checklist items; see Fig.~\ref{fig:planner-checklist}. While our approach is loosely related to~\cite{checklists-are-better}, it differs fundamentally in its role: we use checklists as a mechanism for structured planning in long-horizon image editing, rather than as a reward-shaping signal for LLM response alignment.

% \begin{figure}[t]
% \centering
% \includegraphics[width=\linewidth]{figs/planner_fig.jpg}
% \caption{\textbf{Checklist-guided planner (Stage 1).} Given an input advertisement and a high-level instruction (left, e.g., ``Adapt for students and educational use''), the planner generates a structured, ordered sequence of subtasks (center) that explicitly address items from a checklist (right). In this example, the plan includes updating the headline for a student audience, adding supporting text about academic use cases, and introducing subtle educational visual elements. Checklists can be provided by human or LLM annotators, and can enforce coverage and interpretability, enabling the planner to produce comprehensive, modular plans for downstream orchestration.}
% \label{fig:planner-checklist}
% \end{figure}

This checklist-guided prompting serves two purposes. First, it enforces coverage, ensuring the planner addresses all relevant aspects rather than producing partial plans. Second, it provides modular, human-interpretable supervision without requiring gold-standard plans. Compared to hard-coded templates, it avoids brittle heuristics while retaining structured guidance. Our experiments in Appendix~\ref{supp:clist_comp} demonstrate that plans generated with checklist guidance provide greater coverage and suggest more contextual edits compared to plans generated without a checklist.

% This checklist-guided prompting serves two purposes. 
% First, it enforces coverage: the planner must account for all relevant aspects of the task rather than producing partial or underspecified plans. 
% Second, it provides modular and human-interpretable supervision—checklists can be authored or refined by annotators without requiring fully specified gold-standard plans. 
% Compared to hard-coded planning templates, this approach avoids brittle heuristics while retaining structural guidance.

\vspace{-5pt}
\paragraph{Self-Supervised Fine-Tuning.}
Let $\mathcal{P}^*=\{s_1^*,\dots,s_T^*\}$ denote the checklist-guided plan produced by the planner, where each sub-task $s_t^*$ is a token sequence $s_t^*=(s_{t,1}^*,\dots,s_{t,N_t}^*)$. The planner outputs a structured list of sub-tasks, with each list element corresponding to a distinct operation.

% Let $\mathcal{P}^* = \{s_1^*, \dots, s_T^*\}$ denote the checklist-guided plan produced by the planner, where each sub-task $s_t^*$ is itself a sequence of tokens
% $
% s_t^* = (s_{t,1}^*, \dots, s_{t,N_t}^*).
% $

% The planner generates the plan as a structured list of sub-tasks. Each list element corresponds to one sub-task, which naturally defines the boundaries between consecutive operations.

% Let $\mathcal{P}^* = \{s_1^*, \dots, s_T^*\}$ denote the checklist-guided plan produced by the planner, where each sub-task $s_t^*$ is itself a sequence of tokens
% $
% s_t^* = (s_{t,1}^*, \dots, s_{t,N_t}^*).
% $
% We include special separator tokens (e.g., \texttt{<SUBTASK>}) between consecutive sub-tasks to preserve structure and make task boundaries explicit. \kr{HOW YOU DETERMINE THESE BOUNDARIES?}

We then fine-tune the planner to reproduce the entire plan conditioned only on $(x, I)$ via autoregressive likelihood maximization:
\begin{equation}
\label{eq:planner_loss_exp}
\mathcal{L}_{\mathrm{planner}}
= \mathbb{E}_{(x,I)\sim\mathcal{D}}\Bigg[
- \sum_{t=1}^{T}
  \sum_{j=1}^{N_t}
  \log p_{\theta}\big(s_{t,j}^* \mid x, I, s_{<t}^*, s_{t,<j}^*\big)
\Bigg],
\end{equation}
where $\mathcal{D}$ is the training distribution, $s_{<t}^*$ denotes all tokens from preceding sub-tasks $\{s_1^*, \dots, s_{t-1}^*\}$ and $s_{t,<j}^*$ denotes the tokens preceding position $j$ within sub-task $t$.

Autoregressive modeling over the full plan captures dependencies across subtasks, which is crucial for long-horizon editing (e.g., in advertisement redesign, slogan changes may depend on prior object substitutions). Modeling the plan as an ordered list of subtasks enables coherent sequencing and global consistency while avoiding contradictory operations.

% Autoregressive modeling over the full plan captures structural dependencies across subtasks, which is essential for long-horizon editing. 
% For instance, in advertisement redesign, later steps such as slogan modification may depend on earlier object substitutions. 
% Joint sequence modeling enables the planner to learn coherent ordering, maintain global consistency, and avoid contradictory operations. The plan is represented as a list of subtasks, naturally defining a sequence of operations.

Importantly, supervision is derived from plans sampled from the planner itself under checklist prompting. The model is thus trained via self-distillation, keeping supervision close to its native generation distribution rather than relying on external demonstrations. This has been shown to reduce distribution shift at inference and improves robustness and generalization compared to pure off-policy imitation~\cite{zhao2026self,hubotter2026reinforcement,shenfeld2026selfdistillationenablescontinuallearning}. 

% Importantly, the supervision signal is derived from plans sampled from the planner itself under checklist prompting. 
% Consequently, the model is trained through self-distillation on plans generated by itself, keeping the supervision closer to its native generation distribution rather than relying on external demonstrations.
% % Consequently, training remains \emph{on-policy}: the model is optimized to reproduce plans drawn from its own distribution rather than mimicking an external teacher. 
% This reduces distributional shift at inference time and has been shown to improve robustness and generalization compared to pure off-policy imitation (e.g.,~\cite{zhao2026self,hubotter2026reinforcement,shenfeld2026selfdistillationenablescontinuallearning}). 

At inference, the checklist is no longer needed; the fine-tuned planner directly generates a structured multi-step plan from $(x,I)$. 

%At inference time, the checklist is no longer required; the fine-tuned planner directly generates a structured multi-step plan from $(x, I)$ alone.

%\as{\subsection{Stage 2: Learning Tool Selection from AI Feedback}}

\subsection{Stage 2: Orchestrator via Reward-Driven Tool Selection}

Given $(x, I, \mathcal{P})$, the orchestrator (a multimodal LLM with parameters $\phi$) selects, for each sub-task $s_t$, a tool $a_t$ and a region $r_t$. Tools (detailed in Sec.~\ref{sec:tools}) are represented as token sequences describing editing operations (e.g., object replacement, style transfer, text editing), while regions correspond to either the full image or candidate object/text areas proposed by segmentation or bounding-box models. This discrete representation frames tool and region selection as a language-generation problem, enabling seamless integration with the LLM architecture without task-specific control logic.

%Given $(x, I, \mathcal{P})$, the orchestrator (a multimodal LLM with parameters $\phi$) selects, for each sub-task $s_t$, a tool $a_t$ and a region $r_t$. Tools (which will be described in detail in Sec.~\ref{sec:tools}) are represented as a sequence of tokens corresponding to concise textual descriptions (e.g., tool name, object replacement, style transfer, text editing), while regions correspond either to the full image or to candidate object/text regions generated by a segmentation or bounding-box model tool. This discrete parameterization allows tool selection to be modeled as a language-generation problem, enabling seamless integration with the LLM architecture and avoiding task-specific control logic.

\begin{figure}[t]
\centering
\includegraphics[width=\linewidth]{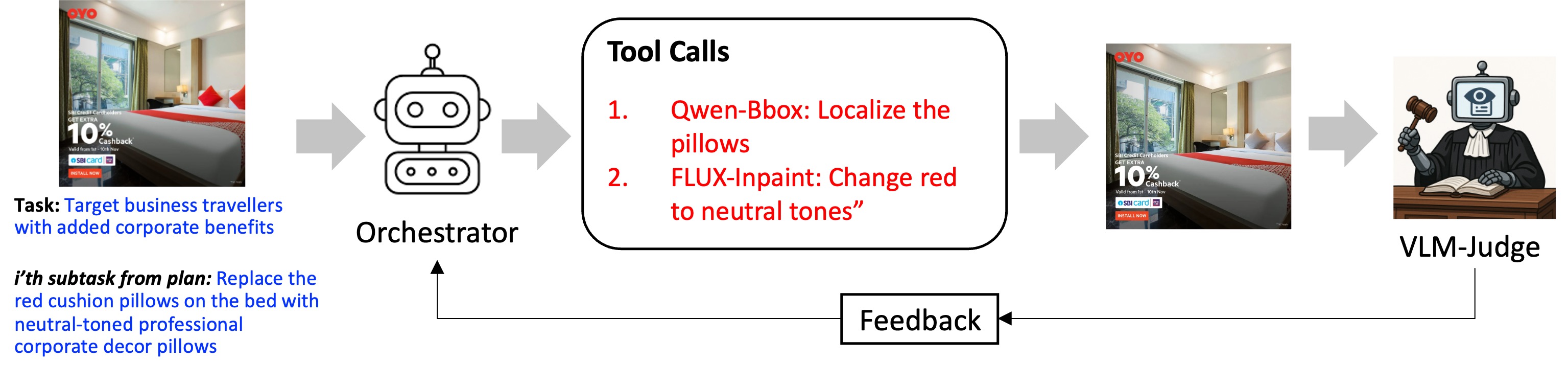}
\caption{\textbf{Reward-driven orchestration (Stage 2).} 
Given an input image and a high-level instruction (e.g., ``Target business travelers with added corporate benefits''), 
the orchestrator selects appropriate tools and spatial regions to execute a specific subtask (e.g., replacing the red pillows with neutral-toned professional decor pillows). 
The edited result is then evaluated by a VLM judge, which assesses instruction adherence, identity preservation, and visual quality. 
The feedback from the judge is used to improve future tool/region selection decisions.}
\label{fig:orchestrator-feedback}
\end{figure}

% \begin{figure}[t]
% \centering
% \includegraphics[width=\linewidth]{figs/orch-tease.png}
% \caption{\textbf{Reward-driven orchestration (Stage 2).} 
% Given an input image and a high-level instruction (e.g., ``Adapt for students and educational use''), 
% the orchestrator selects appropriate tools and spatial regions to execute a specific subtask (e.g., adding ``Perfect for Students'' at the top). 
% The edited result is then evaluated by a VLM judge, which assesses instruction adherence, identity preservation, and overall visual quality. 
% The feedback from the judge is used to improve future tool and region selection decisions.}
% \label{fig:orchestrator-feedback}
% \end{figure}

Executing the selected sequence yields the final edited image:
\begin{equation}
    \hat{x} 
    = 
    f_{a_T, r_T} \circ \cdots \circ f_{a_1, r_1}(x),
\end{equation}
where $f_{a_t, r_t}$ applies tool $a_t$ to region $r_t$ (when applicable). Sequential composition allows later edits to refine or build upon earlier ones, which is essential for long-horizon tasks.
% Executing the selected sequence produces the final edited image:
% \begin{equation}
%     \hat{x} 
%     = 
%     f_{a_T, r_T} \circ \cdots \circ f_{a_1, r_1}(x),
% \end{equation}
% where $f_{a_t, r_t}$ denotes applying tool $a_t$ to region $r_t$ (when applicable). Sequential composition is necessary because later edits may refine or build upon earlier modifications, especially in long-horizon tasks.

\vspace{-5pt}
\paragraph{Reward Function.}
We use a strong MLLM-based judge~\cite{chatgpt} to assign a scalar reward $R(\hat{x}, x, I)$ conditioned on the edited image $\hat{x}$, the original image $x$, and the instruction $I$. The judge evaluates instruction adherence, identity preservation, and overall visual quality (e.g., layout fidelity and realism; see Fig.~\ref{fig:orchestrator-feedback}). Since multiple outputs may satisfy the same instruction, a scalar reward provides flexible supervision without requiring pixel-level alignment. Importantly, the judge is used only to provide outcome signals, rather than dense token-level supervision. Implementation details of the judge are included in Appendix~\ref{app:orchestrator-details}. As demonstrated in our user studies (Sec.~\ref{sec:mainresults}), improvements transfer to human preference, suggesting that the learned policy is not merely overfitting to the judge's scoring function.

%To evaluate the quality of the final output, we employ a strong MLLM-based judge~\cite{chatgpt} that assigns a scalar reward $R(\hat{x}, x, I),$ conditioned on the edited image $\hat{x}$, the original image $x$, and the instruction $I$. The judge evaluates instruction adherence, identity consistency with respect to $x$, and overall presentability (e.g., layout fidelity, visual realism, and typographic clarity); see Fig.~\ref{fig:orchestrator-feedback}. Using a judge is appropriate in this setting because there is no unique ground-truth target image; multiple edited results may satisfy the instruction. A scalar reward therefore provides flexible supervision that captures these factors without relying on pixel-level alignment.

%Because the quality of an edit can only be assessed after all modifications have been applied, the reward is computed only on the final edited image.
%Although the reward is computed only at the end, it depends on the entire sequence of intermediate edits that produced $\hat{x}$. 

\vspace{-5pt}
\paragraph{Policy Optimization Objective.}
Our objective is to maximize the reward of the full editing trajectory. 
Given tool–region decisions $(a_{1:T}, r_{1:T})$, executing the edits produces a final image $\hat{x}$, which is evaluated by the VLM judge with reward $R(\hat{x}, x, I)$. We therefore optimize the expected trajectory reward:
\begin{equation}
    \max_{\phi} \;
    \mathbb{E}_{(a_{1:T}, r_{1:T}) \sim \pi_\phi}
    \big[
        R(\hat{x}, x, I)
    \big].
\end{equation}
Optimizing the trajectory-level reward encourages coordinated decisions across steps, since the quality of later edits depends on earlier tool and region selections.

In practice, we sample candidate trajectories and select high-reward ones as supervision signals: $(a_{1:T}^*, r_{1:T}^*) = \arg\max_{(a_{1:T}, r_{1:T})} R(\hat{x}, x, I).$ When multiple trajectories achieve comparable rewards, all can be used for training.  We then train the orchestrator to reproduce these high-reward trajectories by maximizing their likelihood:
\begin{equation}
\mathcal{L}_{\mathrm{orch}}
=
-
\mathbb{E}_{(x, I, \mathcal{P}, a_{1:T}^*, r_{1:T}^*)}
\left[
\sum_{t=1}^{T}
\log 
\pi_\phi
\big(
a_t^*, r_t^*
\mid
x, I, \mathcal{P}, a_{<t}^*, r_{<t}^*
\big)
\right].
\end{equation}
This aligns training with inference-time behavior, grounding tool and region selection in empirically successful trajectories while remaining computationally tractable.

\vspace{-5pt}
\paragraph{Efficient Reward Approximation}
Learning high-reward editing actions requires exploring tool and region selections, but evaluating a full trajectory is costly due to sequential diffusion calls. Enumerating and scoring all candidate sequences offline is also infeasible, as the number of tool–region combinations grows exponentially with the number of sub-tasks. To make training tractable, we introduce two structured approximations that exploit the compositional nature of high-level image edits.

%Directly learning which editing actions lead to high rewards requires exploration over tool and region selections. However, such exploration is expensive, since evaluating a single trajectory requires sequentially invoking diffusion-based editors. A natural alternative is to estimate rewards offline by enumerating candidate actions and scoring the resulting edits. In practice, this is infeasible: even for a fixed instruction and plan, the number of possible tool–region sequences grows exponentially with the number of sub-tasks.  Therefore, to make training tractable, we introduce two structured approximations that exploit the compositional nature of many high-level image edits.

% Directly optimizing over full editing trajectories is computationally expensive, as each rollout requires sequentially invoking large diffusion-based editors. To make training tractable, we introduce two structured approximations that exploit the compositional nature of many high-level image edits.

\vspace{-5pt}
\paragraph{Additive Reward Approximation.}
Many edits correspond to semantically distinct operations (e.g., object replacement, slogan modification, background recoloring) that are largely independent. Moreover, achieving a high-quality final result requires each sub-task to be executed correctly. We therefore approximate the trajectory-level reward as the sum of sub-task contributions: $R(\hat{x}, x, I) \approx \sum_{t=1}^{T} R_t,$ where $R_t$ reflects whether sub-task $s_t$ has been successfully completed.

%We observe that many edits correspond to semantically distinct operations (e.g., object replacement, slogan modification, or background recoloring), meaning they often do not depend strongly on one another. Furthermore, producing a high-quality final edit requires that each step be executed correctly, making all steps equally important to the final outcome.
% Based on the independence of steps in the plan and the equal importance of each step, we approximate the trajectory-level reward as the sum of contributions from each sub-task:
%$R(\hat{x}, x, I) \approx \sum_{t=1}^{T} R_t$.
%where $R_t$ measures whether the edit corresponding to sub-task $s_t$ has been correctly executed.

% \paragraph{Additive Reward Approximation.}
% We approximate the trajectory-level reward as a sum of sub-task contributions: $R(\hat{x}, x, I) \approx \sum_{t=1}^{T} R_t,$ where $R_t$ represents the reward attributable to sub-task $s_t$. 
% This approximation is motivated by the observation that many editing operations (e.g., object replacement, slogan modification, background recoloring) correspond to distinct semantic criteria in the judge evaluation.

\vspace{-5pt}
\paragraph{Original-Image Independence Approximation.}
Many edits correspond to largely independent operations, so the effect of a tool is often weakly dependent on prior edits (e.g., product replacement typically does not depend strongly on an earlier background object change). We therefore estimate the contribution of a tool by evaluating it directly on the original image rather than on intermediate edits. Formally, let $x_{t-1}$ denote the intermediate image before applying $(a_t, r_t)$. We approximate $R_t\big(f_{a_t, r_t}(x_{t-1}),\, x,\, I\big) \approx R_t\big(f_{a_t, r_t}(x),\, x,\, I\big).$

\vspace{5pt}
Together, these approximations allow us to precompute all tool–region candidates and their rewards, $\{(a, r, R_{a,r})\}$. For each sub-task, we identify the highest-reward tools and train the orchestrator to predict these selections.

%Together, these approximations enable precomputation of all tool–region candidates and their associated rewards: $\{(a, r, R_{a,r})\}$. Since we have the rewards for each candidate tool, we can determine which tool performs best for each sub-task. We then train the orchestrator to predict these selections.

% During training, the orchestrator can retrieve the corresponding reward without repeatedly executing expensive generative models, substantially reducing memory and compute requirements while preserving meaningful optimization signals.

\subsection{Closing the Loop: Plan Refinement}\label{sec:closedloop}

To ensure coordination between planner and orchestrator, we refine the initial plan by removing sub-tasks whose maximum achievable reward across tools and regions falls below a threshold $\tau$: $\max_{a,r} R_{a,r}(s_t) < \tau$. Such sub-tasks correspond to operations unsupported by the available toolset. Pruning them prevents systematically infeasible decompositions and improves consistency between planning and execution. Thus, before training the orchestrator, we retrain the planner on the revised plans to better reflect the feasible action space. We then train the orchestrator only on the subtasks which achieve a reward greater than the threshold. This closed-loop refinement grounds high-level reasoning in executable actions, enabling scalable and robust long-horizon image editing without handcrafted pipelines. 

% To ensure coordination between planner and orchestrator, we refine the initial plan by removing sub-tasks whose maximum achievable reward across all tools and regions falls below a threshold $\tau$: $\max_{a,r} R_{a,r}(s_t) < \tau$.  Such sub-tasks correspond to operations unsupported by available tools. Removing them improves consistency between planning and execution which prevents the planner from proposing systematically infeasible operations. Thus, before training the orchestrator, we first retrain the planner using the revised plans to better reflect the feasible action space.  We then train the orchestrator with the revised planner, closing the loop between planning and execution and promoting consistency across the two components. 

% This closed-loop refinement aligns high-level reasoning with low-level affordances, analogous to grounding symbolic plans in executable action spaces. By combining checklist-guided planning trained on its own output distribution, reward-driven orchestration, and computationally efficient approximations, our framework enables scalable and robust long-horizon image editing without handcrafted pipelines.

\subsection{Inference via Verifier-Guided Selection}

%During inference, to improve robustness during sequential editing, we augment the orchestrator with a lightweight verifier-guided selection step.
%in $[0,5]$ 

To improve robustness during sequential editing, we augment the orchestrator with a lightweight verifier-guided selection step. Specifically, we train a verifier to score intermediate edits. Given the original image $x$, a sub-task $s_t$, and the edited image $\tilde{x}_t = f_{a_t,r_t}(x_{t-1})$, the verifier predicts a score reflecting sub-task correctness, identity preservation, and visual quality. Teacher scores from the same VLM judge used during training~\cite{chatgpt} are distilled into a smaller VLM~\cite{bai2025qwen3}, enabling efficient inference. For each sub-task, the orchestrator proposes a distribution over tool–region pairs. We select the top-$k$ candidates by policy likelihood, execute these edits, and re-rank them using the verifier: $(a_t^*, r_t^*) = \arg\max_{i \in \{1,\dots,k\}} \mathrm{Verifier}\big(f_{a_t^{(i)}, r_t^{(i)}}(x_{t-1}), x, s_t\big).$ The highest-scoring edit is used for the next step. This proposal–re-ranking strategy reduces error accumulation while remaining tractable; in practice, $k=3$ or $k=5$ works well. After completing all sub-tasks, we apply a lightweight refinement on the final result to improve coherence while preserving the intended edits.

\subsection{Tools}\label{sec:tools}

%Our framework operates over three categories of tools: analysis tools for region discovery, whole-image editing tools for global transformations, and region-level editing tools for localized modifications.

Our framework uses  analysis tools for region discovery, whole-image editors for global changes, and region-level editors for localized edits.

\vspace{-5pt}
\paragraph{Analysis Tools.}
These identify editable regions: 
(i) \textbf{SAM-2 + Qwen-3VL}~\cite{ravi2024sam2} for semantic segmentation with masks and descriptions; 
(ii) \textbf{DeepSeek-OCR}~\cite{wei2025deepseek} for layout and text detection; 
(iii) \textbf{Qwen-Layered}~\cite{yin2025qwenimagelayered} for foreground-to-background layer decomposition, capturing larger structural regions that may not be detected by object-level segmentation; 
(iv) \textbf{Qwen-BBox}~\cite{bai2025qwen3} for instruction-guided bounding boxes, useful for edits involving adding or modifying objects not easily captured by image-only analysis.

\vspace{-5pt}
\paragraph{Whole-Image Editing Tools.}
(v) \textbf{Qwen-Image-Edit}~\cite{wu2025qwenimagetechnicalreport} and 
(vi) \textbf{Flux-Kontext-Edit}~\cite{labs2025flux1kontextflowmatching} apply instruction-guided edits to the entire image.

%These apply instruction-guided edits to the entire image:
%(v) \textbf{Qwen-Image-Edit}~\cite{wu2025qwenimagetechnicalreport} and (vi) \textbf{Flux-Kontext-Edit}~\cite{labs2025flux1kontextflowmatching}.

\vspace{-5pt}
\paragraph{Region-Level Editing Tools.}
% These operate on regions identified by an analysis tool and require a \texttt{region\_number} argument:
% (vii) \textbf{Flux-Inpaint}~\cite{labs2025flux1kontextflowmatching}: Diffusion-based masked editing.
(vii) \textbf{Flux-Inpaint}~\cite{labs2025flux1kontextflowmatching} performs masked diffusion editing on regions specified by an analysis tool.

%(viii) \textbf{Qwen-Textrend}~\cite{}: Programmatic text generation and rendering within a region.

%\paragraph{Workflow Constraints.}
Whole-image tools operate directly, while region-level tools require a prior analysis step and a valid region index. Allowed compositions are: 
(1) Layered/BBox/SAM-2/OCR $\rightarrow$ Flux-Inpaint; 
(2) Qwen-Image-Edit (standalone); 
(3) Flux-Kontext-Edit (standalone). 
All tools return structured JSON outputs for consistent orchestration. A comprehensive description of our tools is provided in Appendix~\ref{app:tool_desc}.

% Whole-image tools can be used directly. Region-level tools must follow an analysis step and reference a valid region. Allowed compositions are:
% (1) Layered/BBox/SAM-2/OCR $\rightarrow$ Flux-Inpaint; (2) Qwen-Image-Edit (standalone); (3) Flux-Kontext-Edit (standalone).  All tools return structured JSON outputs to ensure consistent orchestration.
%(2) BBox/OCR $\rightarrow$ Qwen-Textrend; 

\section{Experiments}

\paragraph{Implementation Details}
The planner and orchestrator are initialized from Qwen3-VL-8B~\cite{bai2025qwen3} and fine-tuned with LoRA~\cite{hu2022lora}. The planner uses a lightweight LoRA setup ($r=1$) applied to \texttt{q\_proj} and \texttt{v\_proj}, while the orchestrator uses higher capacity ($r=64$) to enable flexible tool selection. Both are trained with learning rate $2\times10^{-4}$ and scaling factor $\alpha=2r$. Training is performed on a single node with 8 A100 80GB GPUs using batch size 16.

% \paragraph{Implementation Details}
% Both the planner and orchestrator are initialized from Qwen3-VL 8B \cite{bai2025qwen3} and fine-tuned using low-rank adaptation (LoRA) \cite{hu2022lora}. The planner uses a lightweight configuration with rank $r=1$ and updates only the \texttt{q\_proj} and \texttt{v\_proj} layers. 
% The orchestrator uses a higher-capacity LoRA configuration with rank $r=64$ to allow greater flexibility in tool-selection behavior. Both models are trained with a learning rate of $2\times10^{-4}$. For both models, we follow standard practice and set the LoRA scaling factor $\alpha$ to twice the rank ($\alpha = 2r$). Training is performed on a single node with 8 A100 80GB GPUs using a batch size of 16.

\vspace{-5pt}
\paragraph{Dataset Details}
We use images from MadVerse~\cite{sagar2024madverse}, a large-scale multilingual advertisement dataset. For each image, we generate three abstract, high-level editing tasks using GPT-5, designed to require multi-step transformations such as cultural adaptation, audience retargeting, promotional shifts, product substitution, or stylistic changes. For training the orchestrator, we use a training dataset with 7,598 instances. For testing, we use a dataset comprising 200 advertisement editing requests. In addition, we also evaluate our approach on standard image editing benchmarks such as GEdit-Bench~\cite{liu2025step1x} and MagicBrush~\cite{zhang2023magicbrush}. We report these results in Appendix~\ref{app:comm-bmarks}.

\subsection{Main Results: Comparison to End-to-End Editing Baselines}\label{sec:mainresults}

We first compare to recent state-of-the-art open-source image editing models to evaluate their ability to perform complex edits directly from high-level instructions. In particular, we test whether these models can reason about multi-step modifications and execute them correctly in a single editing pass.

\vspace{-5pt}
\paragraph{Baselines}
We compare against FLUX.1-Kontext-dev~\cite{labs2025flux1kontextflowmatching} and Qwen-Image-Edit-2511~\cite{wu2025qwenimagetechnicalreport}. We evaluate these models in two settings. In the first, the high-level instruction is provided directly to the model, testing its ability to reason and perform the edit in a single step. In the second, we use our base Qwen3-VL-8B model to decompose the task into a sequence of simpler steps, which are then provided to the editing model at once. This setting evaluates whether a plan generated by a general MLLM can be executed effectively in a single-shot edit.

\vspace{-5pt}
\paragraph{Evaluation}
A successful edit should satisfy three key criteria: correct execution of the instruction, preservation of important elements from the input image, and high visual quality. To evaluate these aspects, we use a strong MLLM as a judge, specifically Gemini-3-Pro~\cite{gemini3pro}. Determining whether an edit follows the instruction often requires world knowledge and reasoning about the intended changes. Therefore, we ask Gemini to score each category on a scale of 1–5 based on the input image, edited image, and the high-level task. Visual quality is instruction-agnostic and is evaluated using only the edited image. To reduce potential bias from any single judge model, we use a different judge during evaluation from training, and final comparisons are corroborated with human A/B studies. This ensures that improvements reflect perceptual and instruction-level gains beyond judge-specific artifacts. Further details of the evaluation setup are provided in Appendix~\ref{subsubsec:wholeediteval}.

\vspace{-5pt}
\paragraph{Results}
Our method achieves the highest instruction-following score, highlighting the benefit of explicit planning and step-by-step execution for complex edits; see Table~\ref{tab:end_to_end_baselines_abstract}. While FLUX.1-Kontext-dev (High-Level Instruction) attains higher identity preservation and visual quality scores, this is mainly because it often leaves the image nearly unchanged, as reflected by its low instruction-following score (see Fig.~\ref{fig:qual_results} for examples). In contrast, our method performs the requested edits while maintaining strong input preservation and visual quality. 
%Since it is not immediately clear which trade-off would be preferred by humans, we additionally conduct a user study to evaluate the perceived quality of the edits.

%\subsection{User Studies}
%We conduct a user study to assess how effectively our method executes high-level, open-ended editing instructions. 

\vspace{-5pt}
\paragraph{User Studies}

\begin{wrapfigure}{r}{0.4\linewidth}
\vspace{-15pt}
\centering
\includegraphics[width=\linewidth]{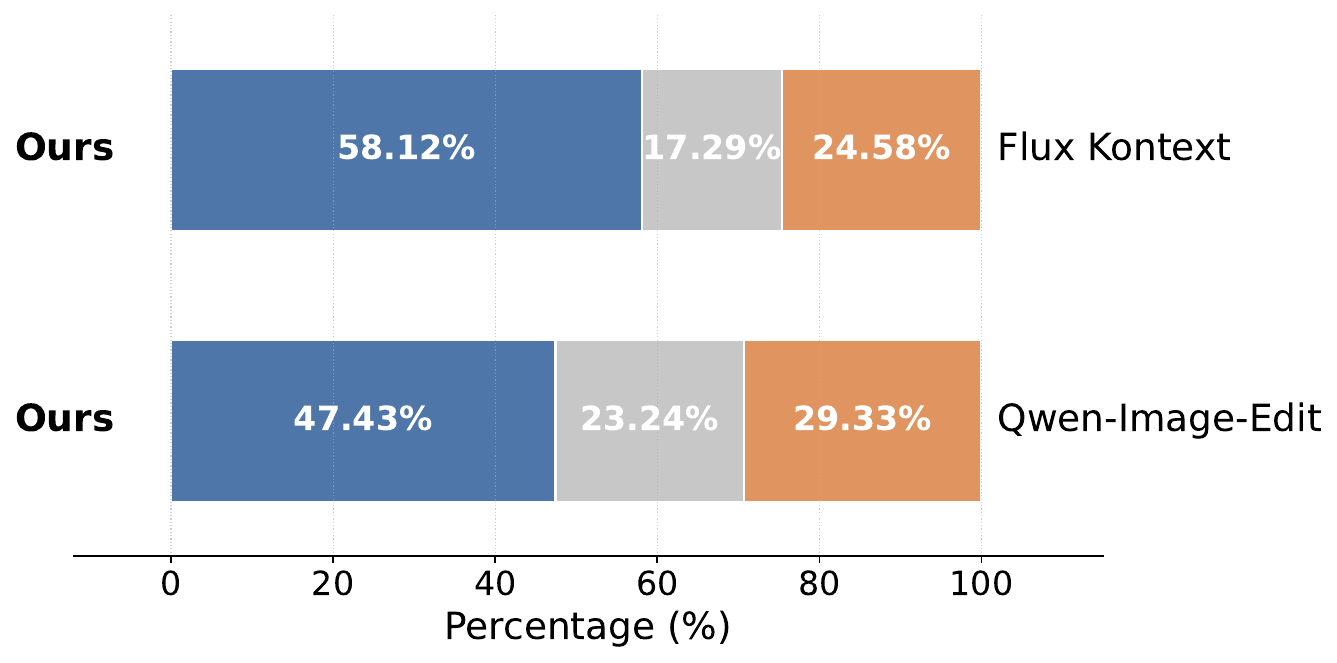}
\vspace{-15pt}
\caption{\textbf{User study (randomized A/B testing).}
Results show a consistent human preference for our approach.}
\label{fig:user_study}
\vspace{-15pt}
\end{wrapfigure}

To corroborate the MLLM judge results, we conduct a user study using randomized A/B testing. Participants are shown paired results in random order and asked to select their preferred edit or indicate a tie, while accounting for instruction following, identity preservation, and visual quality. Each pair is evaluated by three unique users. As shown in Fig.~\ref{fig:user_study}, highlighting the advantages of long-horizon planning and reward-driven orchestration for producing coherent, instruction-faithful edits.

\begin{table}[t]
\centering
\footnotesize
\setlength{\tabcolsep}{4pt}
\renewcommand{\arraystretch}{1.05}
\resizebox{\linewidth}{!}{
\begin{tabular}{l c c c}
\toprule
\textbf{Method} & \textbf{Instruction Following} & \textbf{Identity Preservation} & \textbf{Visual Quality} \\
\midrule
FLUX.1-Kontext-dev (High-Level Instruction) & 2.32 & \textbf{4.32} & \textbf{3.525} \\
FLUX.1-Kontext-dev (Qwen3-VL-8B Plan) & 3.33 & 3.005 & 2.405 \\
Qwen-Image-Edit-2511 (High-Level Instruction) & 3.355 & 2.769 & 2.26 \\
Qwen-Image-Edit-2511 (Qwen3-VL-8B Plan) & 3.807 & 3.005 & 2.16 \\
\textbf{Ours} & \textbf{4.196} & 3.155 & 2.525 \\
\bottomrule
\end{tabular}
}
\caption{Comparison with end-to-end editing baselines. Scores are averaged Gemini-3-Pro evaluations. Baselines either preserve the original image but fail to execute the instruction, or attempt the edit at the cost of identity preservation. Our method achieves the strongest instruction execution while maintaining good identity preservation.}
\label{tab:end_to_end_baselines_abstract}
\end{table}

\subsection{Ablation Study: Orchestrator Design Choices}

We next isolate the key design choices underlying the Stage 2 orchestrator. Specifically, we demonstrate that (i) leveraging multiple tools outperforms relying on a single tool, (ii) training the orchestrator to learn an explicit tool-calling policy yields better performance than prompting the base model to do so in a training-free manner, and (iii) we study the effect of using different $k$ values (number of candidates) during inference. 
%Untrained Orchestrator vs Our Orchestrator (inference. eval)

\vspace{-5pt}
\paragraph{Experiment Details}
To isolate the design choices of the orchestrator, we fix the multi-step plan generated by our planner model across all variants. We then compare the following configurations:
(i) \textit{FLUX.1-Kontext-dev (sequential)}: FLUX-Kontext is applied sequentially, once per instruction in the plan.
(ii) \textit{Qwen-Image-Edit-2511 (sequential)}: Qwen-Image-Edit is invoked sequentially for each instruction.
(iii) \textit{Qwen-BBox + FLUX Inpaint}: For each step, we first generate task-relevant regions using Qwen-BBox. A base Qwen3-VL model then selects the appropriate region, and the edit is applied locally using FLUX Inpaint. This process is repeated for every instruction in the plan.
(iv) \textit{Base-Orchestrator}: Tool selection is performed by an untrained base Qwen3-VL model, without learning an explicit tool-calling policy. We provide the same system prompt used for our model, which describes the strengths and capabilities of the available tools. Finally, to study the effect of different values of $k$ during inference, we evaluate the performance with $k = 1, 3,$ and $5$.

\vspace{-5pt}
\paragraph{Evaluation}
We evaluate visual quality using the same metric as the previous section. For instruction following and identity preservation, since all models in this experiment receive the same instructions, we adopt a more detailed evaluation. This is because models are now required to follow a specific plan that is provided uniformly to all of them. To do this, we prompt GPT-5 with the input image, high-level task, and the multi-step plan to generate a set of constraints. These constraints specify elements that should be preserved, modified, or added, as well as conditions on their placement, orientation, color, and other attributes. We then ask Gemini-3-Pro \cite{gemini3pro} to evaluate the edited image against these constraints, given the input image and task. For each constraint, the judge determines whether it is satisfied. We report the percentage of satisfied constraints as the instruction satisfaction/identity preservation score. Further details of the evaluation setup are provided in Appendix~\ref{app:plan-cond-eval}.

\begin{table}[t]
\centering
\footnotesize
\setlength{\tabcolsep}{4pt}
\renewcommand{\arraystretch}{0.95}
\begin{tabular}{l c c}
\toprule
\textbf{Method} & \textbf{Instr (\%)} & \textbf{VQ} \\
\midrule
FLUX.1-Kontext-dev (sequential) & 53.3 & 2.05 \\
Qwen-BBox + FLUX Inpaint & 60.0 & 2.105 \\
Qwen-Image-Edit-2511 (sequential) & 61.9 & 2.29 \\
Qwen3VL-Base  (k=5) & 61.6 & 2.445 \\
Ours (k=1) & 63.9 & 2.245 \\
Ours (k=3) & 71.8 & 2.38\\
Ours (k=5) & \textbf{74.0} & \textbf{2.525} \\
\bottomrule
\end{tabular}
\caption{Instruction satisfaction (Instr) measured as the percentage of checklist constraints satisfied and Visual Quality (VQ). Increasing inference branches improves both metrics. Training the orchestrator shows significant improvements (12.4\%) over the base model.}
\label{tab:orchestrator_ablations}
\end{table}

As shown in Table~\ref{tab:orchestrator_ablations}, our trained orchestrator significantly outperforms all single-tool baselines as well as the untrained orchestrator. This highlights both the benefits of tool use and the importance of learning an effective tool-selection policy through experience.

Furthermore, our multi-branch variants achieve the highest instruction satisfaction and visual quality, demonstrating that structured planning with reward-driven orchestration more effectively satisfies detailed constraints than strong end-to-end baselines.  Moreover, increasing the number of candidates (k) consistently improves performance: instruction satisfaction rises from 63.9\% (1 branch) to 71.8\% (3 branches) and 74.0\% (5 branches), with corresponding gains in visual quality. This trend indicates that broader search during orchestration enables more constraint-compliant and higher-quality edits.

\subsection{Ablation Study: Plan Dataset: Self- vs.~External Supervision}

We next evaluate our checklist-guided planner to assess whether on-policy self-distillation improves plan quality and distributional alignment compared to direct teacher imitation.  Specifically, in Sec.~\ref{subsec:planner}, we use a checklist to encourage comprehensive plan generation. An alternative to checklist-guided self-training is to directly fine-tune the base model on plans generated by a stronger teacher LLM. 
However, such supervision may introduce distribution shift if teacher-generated plans deviate from the base model’s native generation patterns. 
To quantify this effect, we measure the base model’s perplexity under teacher forcing on two sets of plans: (1) checklist-conditioned plans generated by the base model itself, and (2) plans generated by a teacher LLM (GPT-5~\cite{chatgpt}).

As shown in Table~\ref{tab:combined} (left), teacher-generated plans yield substantially higher perplexity than self-generated checklist plans. 
This gap indicates that teacher plans lie far outside the base model’s intrinsic distribution, providing empirical evidence of potential instability under off-policy imitation. 
In contrast, our checklist-guided self-distillation maintains on-policy supervision and better aligns training with the model’s natural generation behavior.

% \begin{table}[t]
% \centering
% \begin{tabular}{l c}
% \toprule
% \textbf{Plan Source} & \textbf{Perplexity} \\
% \midrule
% Checklist (self-generated) & 4.89 \\
% GPT-5 (teacher-generated) & 61.25 \\
% \bottomrule
% \end{tabular}
% \caption{Perplexity of the base model on different plan sources (lower is better).}
% \label{tab:ppl}
% \end{table}

\subsection{Ablation Study: Effect of Plan Refinement}

%\subsection{Effect of removing the steps which had a score <= 3}
%Before average max reward was 4.1708 after filtration 4.3095.

To quantify the impact of our closed-loop plan refinement (Sec.~\ref{sec:closedloop}), we compute, for each sub-task, the maximum achievable reward across all tools and regions, $\max_{a,r} R_{a,r}(s_t)$, and average this value over the samples. Rewards are assigned by a VLM-based judge. 

Table~\ref{tab:combined} (right) reports the average maximum reward before and after filtering infeasible sub-tasks. Filtering infeasible sub-tasks increases the average reward, quantifying improved compatibility between plans and executable tools.

\begin{table}[t]
\centering
\begin{subtable}[t]{0.48\linewidth}
\centering
\begin{tabular}{l c}
\toprule
\textbf{Plan Source} & \textbf{Perplexity} \\
\midrule
Checklist (self-generated) & 4.89 \\
GPT-5 (teacher-generated) & 61.25 \\
\bottomrule
\end{tabular}
\end{subtable}
\hfill
\begin{subtable}[t]{0.48\linewidth}
\centering
\begin{tabular}{l c}
\toprule
\textbf{Setting} & \textbf{Avg. Max Reward} \\
\midrule
Before filtering & 4.1708 \\
After filtering  & 4.3095 \\
\bottomrule
\end{tabular}
\end{subtable}
\caption{
\textbf{Left:} Perplexity of the base model on different plan sources (lower is better).
\textbf{Right:} Average maximum achievable sub-task reward (Gemini 3 Pro judge) before and after plan refinement.
}
\label{tab:combined}
\end{table}

\begin{figure}[htbp]
\centering
\small
\setlength{\tabcolsep}{2pt} 

\begin{tabular}{cccc}
% ---- Header Row ----
\textbf{Input Image} & \textbf{FLUX Kontext} & \textbf{Qwen-Edit} & \textbf{Ours} \\ [2pt]

% \includegraphics[width=0.24\linewidth]{figs/supplementary/iftar/input.jpg} &
% \includegraphics[width=0.24\linewidth]{figs/supplementary/iftar/flux.jpg} &
% \includegraphics[width=0.24\linewidth]{figs/supplementary/iftar/qwen.jpg} &
% \includegraphics[width=0.24\linewidth]{figs/supplementary/iftar/ours.jpg} \\
% \multicolumn{4}{l}{\hspace{0.5em} \fontsize{6.5pt}{7.5pt}\selectfont \textcolor{gray}{\textbf{Task:} adapt for a Ramadan audience with Iftar-focused items}} \\ [8pt]

\includegraphics[width=0.24\linewidth]{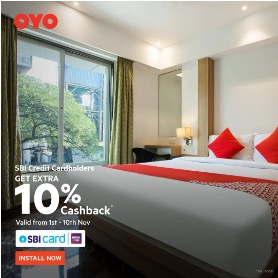} &
\includegraphics[width=0.24\linewidth]{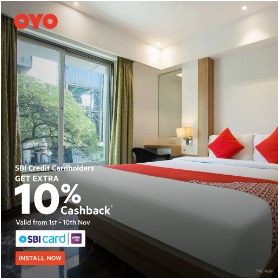} &
\includegraphics[width=0.24\linewidth]{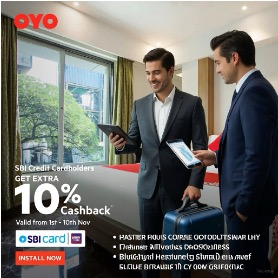} &
\includegraphics[width=0.24\linewidth]{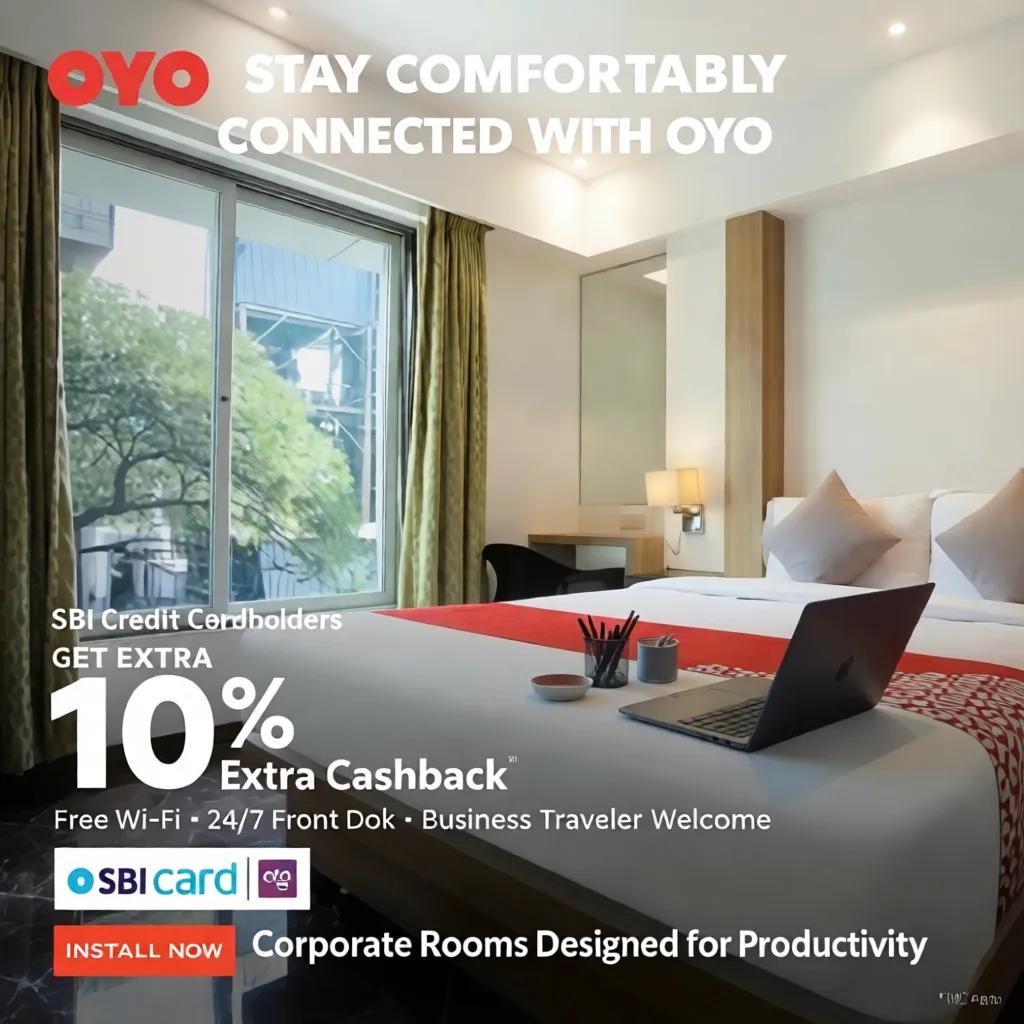} \\
\multicolumn{4}{l}{\hspace{0.5em} \fontsize{6.5pt}{7.5pt}\selectfont \textcolor{gray}{\textbf{Task:} Target business travelers with added corporate benefits}} \\ [8pt]
\includegraphics[height=5.2cm]{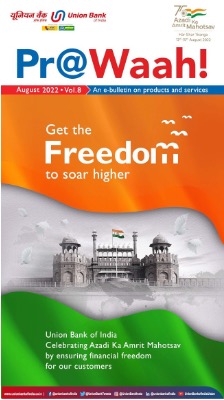} &
\includegraphics[height=5.2cm]{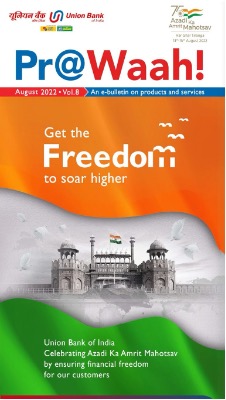} &
\includegraphics[height=5.2cm]{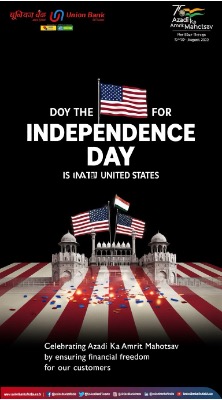} &
\includegraphics[height=5.2cm]{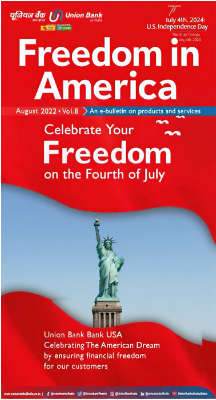} \\
\multicolumn{4}{l}{\hspace{0.5em} \fontsize{6.5pt}{7.5pt}\selectfont \textcolor{gray}{\textbf{Task:} Adapt for Independence Day in the United States.}} \\

\end{tabular}

% \caption{\textbf{Qualitative results on diverse long-horizon advertisement editing tasks.} 
% These examples show three challenging instructions—adapting for a rural audience, students, and for eco-conscious customers. Our method consistently produces edits that are more instruction-faithful and globally coherent, jointly updating visual themes, textual content, layout elements, and brand messaging. Note that we blur the faces in the images to preserve the anonymity of individuals.}
\caption{\textbf{Qualitative results on diverse long-horizon advertisement editing tasks.} 
These examples show two challenging instructions—adapting for Business travelers, and for American Independence Day. Single-step editors (Flux Kontext~\cite{labs2025flux1kontextflowmatching} and Qwen-Image-Edit~\cite{wu2025qwenimagetechnicalreport}) often perform partial stylistic changes or introduce minimal or shallow modifications in text, layout, or branding. In contrast, our method consistently produces edits that are more instruction-faithful and globally coherent, jointly updating visual themes, textual content, layout elements, and brand messaging. These results demonstrate the effectiveness of checklist-guided planning and reward-driven orchestration in handling abstract, multi-step transformations beyond localized edits.}
\label{fig:qual_results}
\end{figure}

\begin{figure}[htbp]
\centering
\small
\setlength{\tabcolsep}{2pt} 

\begin{tabular}{cccc}
% ---- Header Row ----
\textbf{Input Image} & \textbf{FLUX Kontext} & \textbf{Qwen-Edit} & \textbf{Ours} \\ [2pt]

% ---- Row 1: Lunar New Year ----
\includegraphics[width=0.24\linewidth,height=3.5cm]{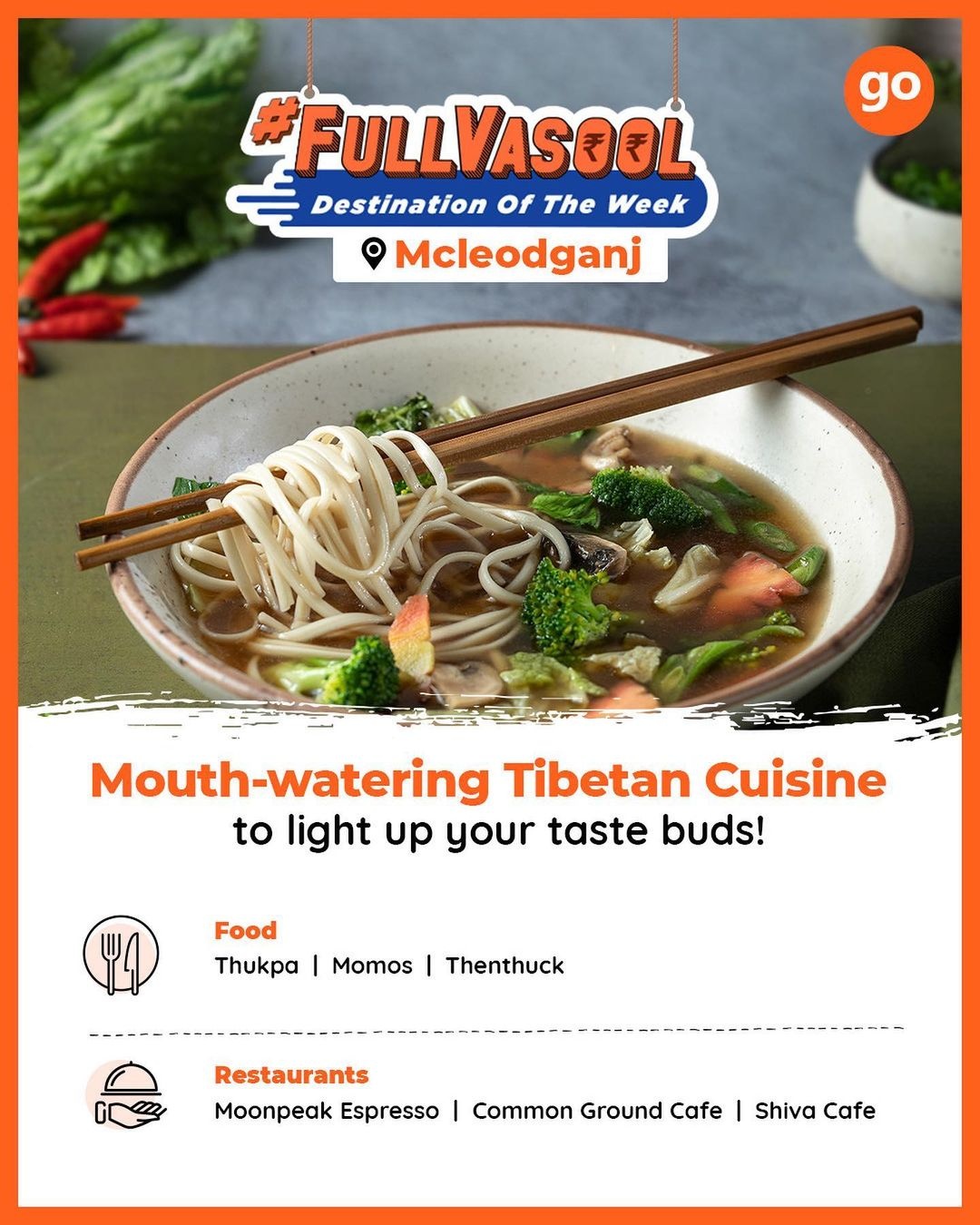} &
\includegraphics[width=0.24\linewidth,height=3.5cm]{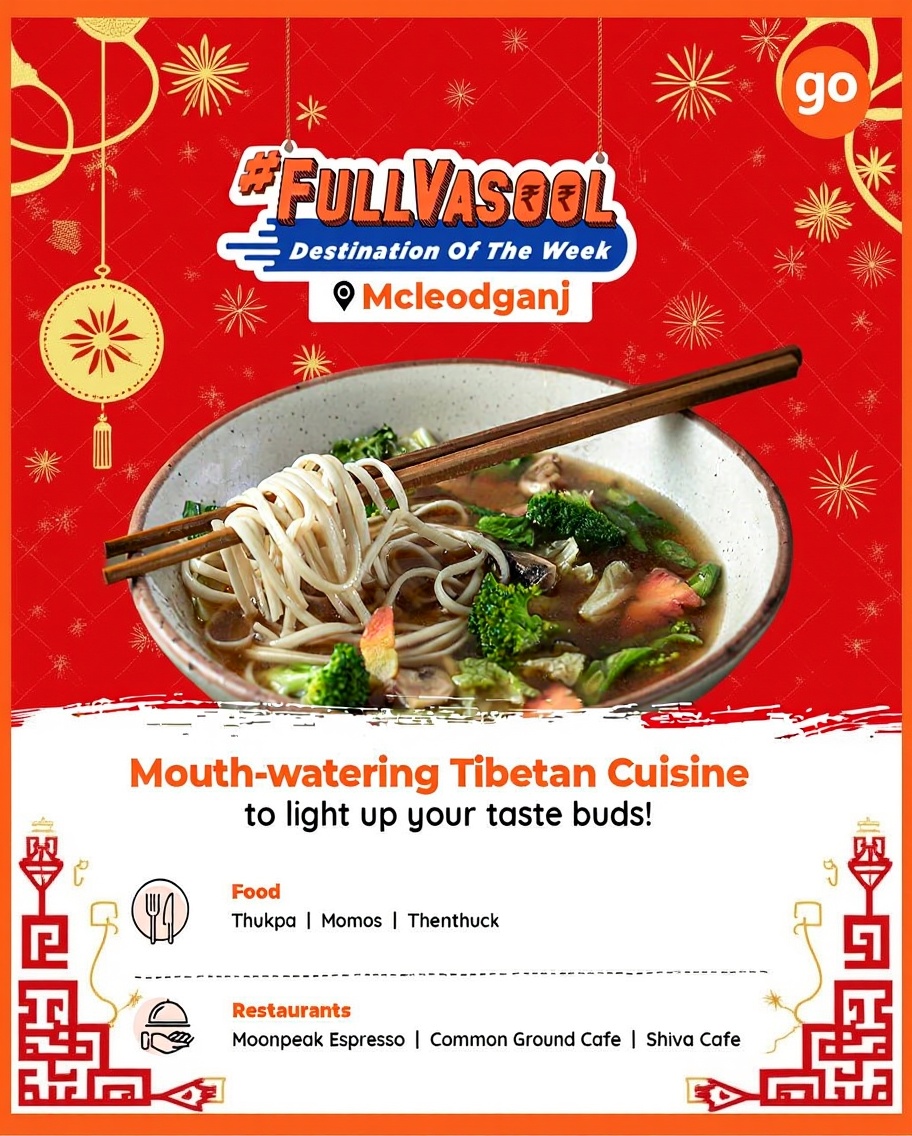} &
\includegraphics[width=0.24\linewidth,height=3.5cm]{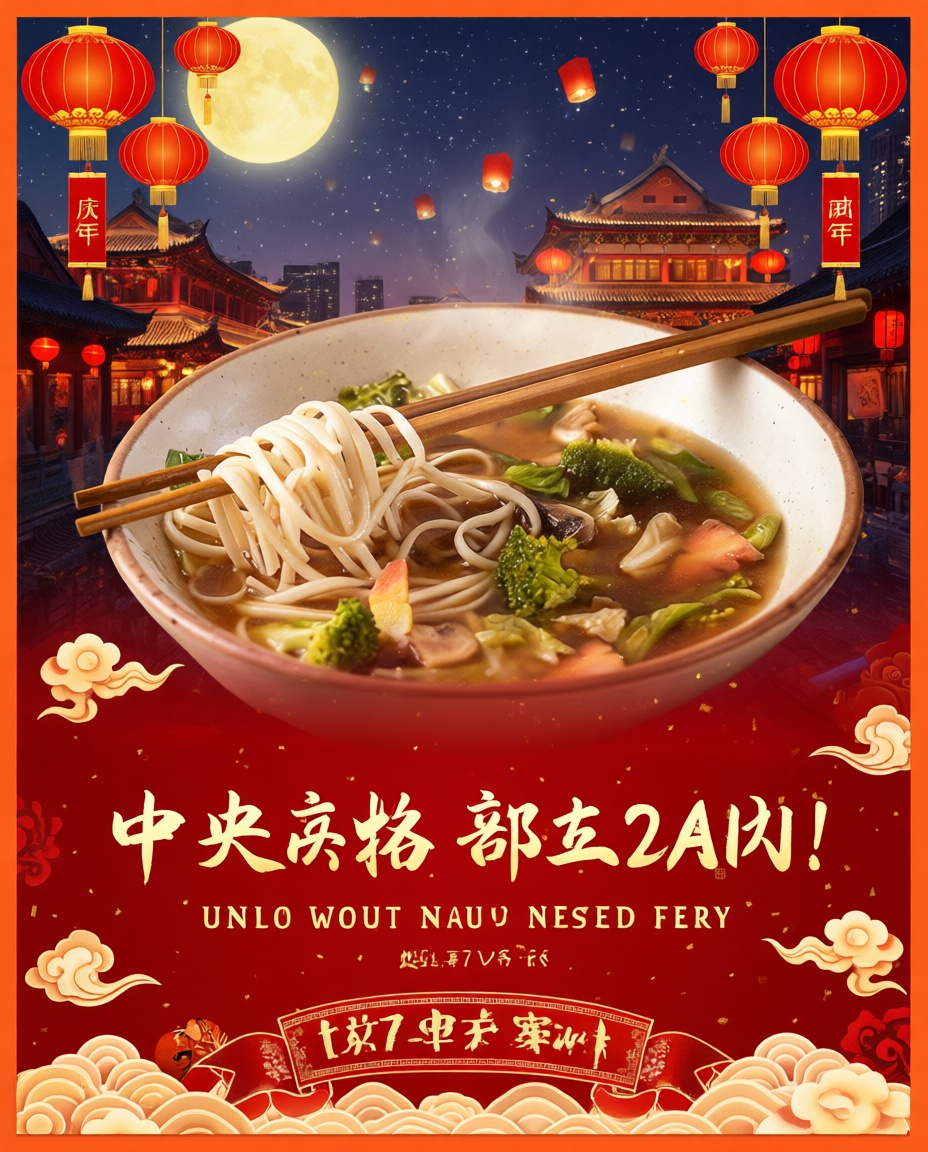} &
\includegraphics[width=0.24\linewidth,height=3.5cm]{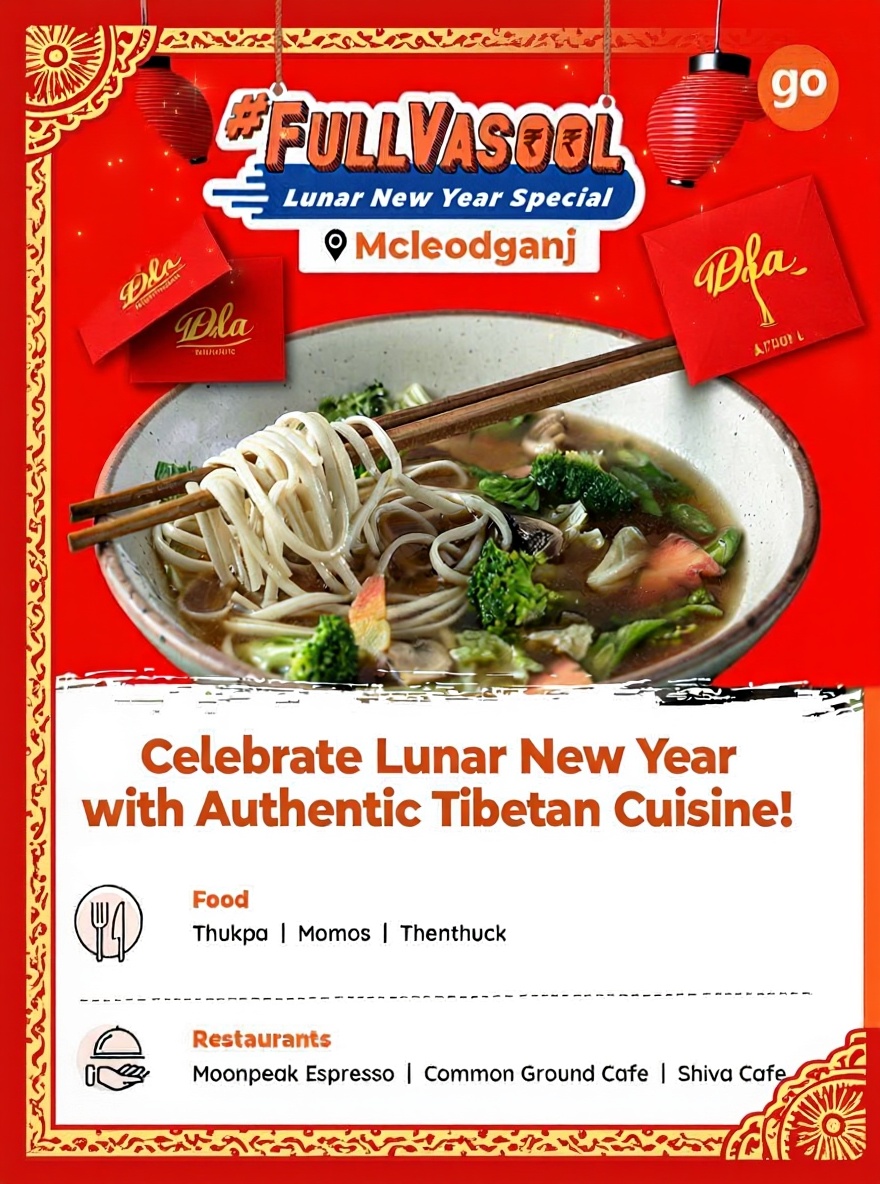} \\
\multicolumn{4}{l}{\hspace{0.5em} \fontsize{6.5pt}{7.5pt}\selectfont \textcolor{gray}{\textbf{Task:} Create a festive edition for Lunar New Year celebrations}} \\ [8pt]

% ---- Row 2: Tropical Fruit ----
\includegraphics[width=0.24\linewidth,height=3.2cm]{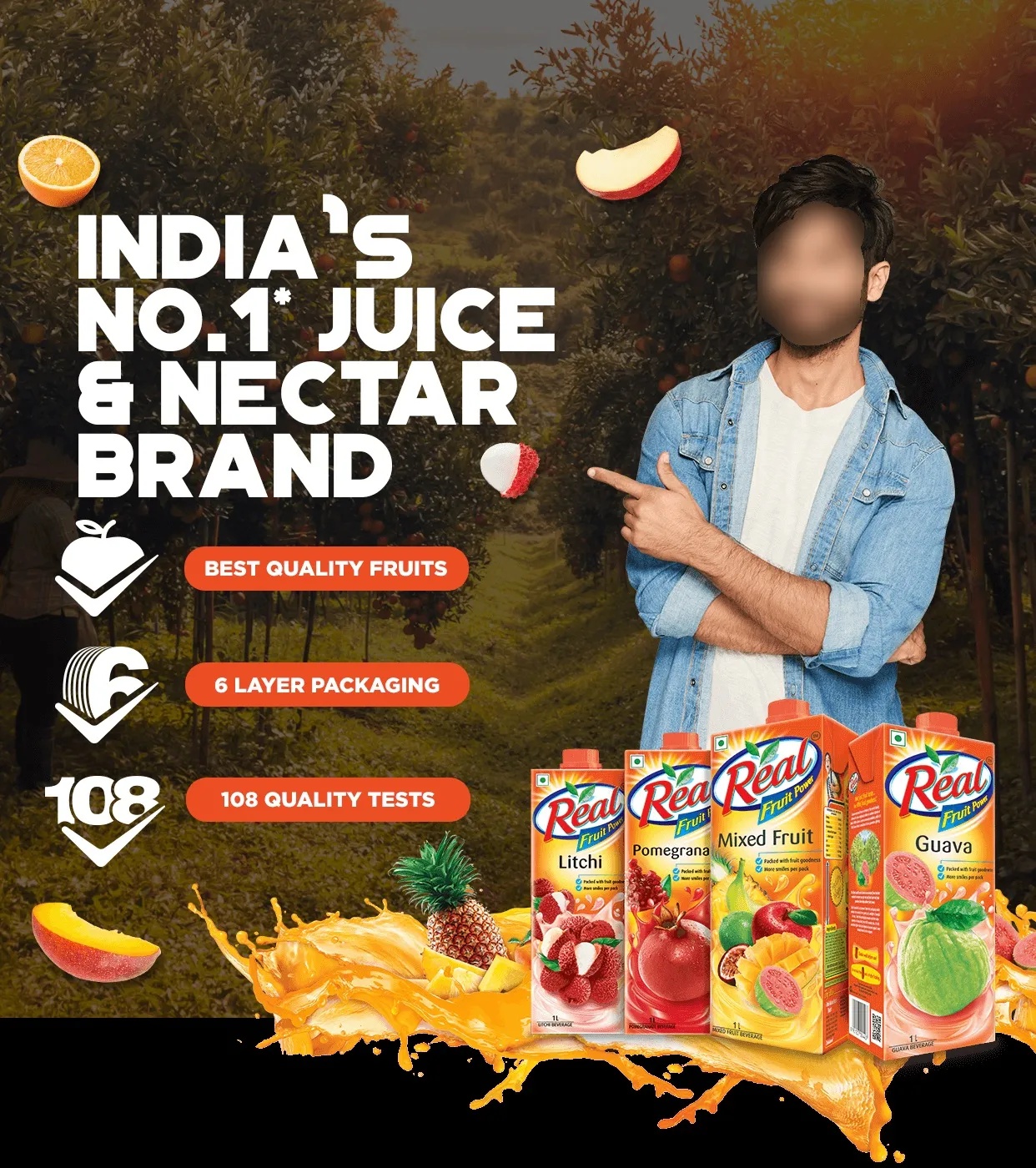} &
\includegraphics[width=0.24\linewidth,height=3.2cm]{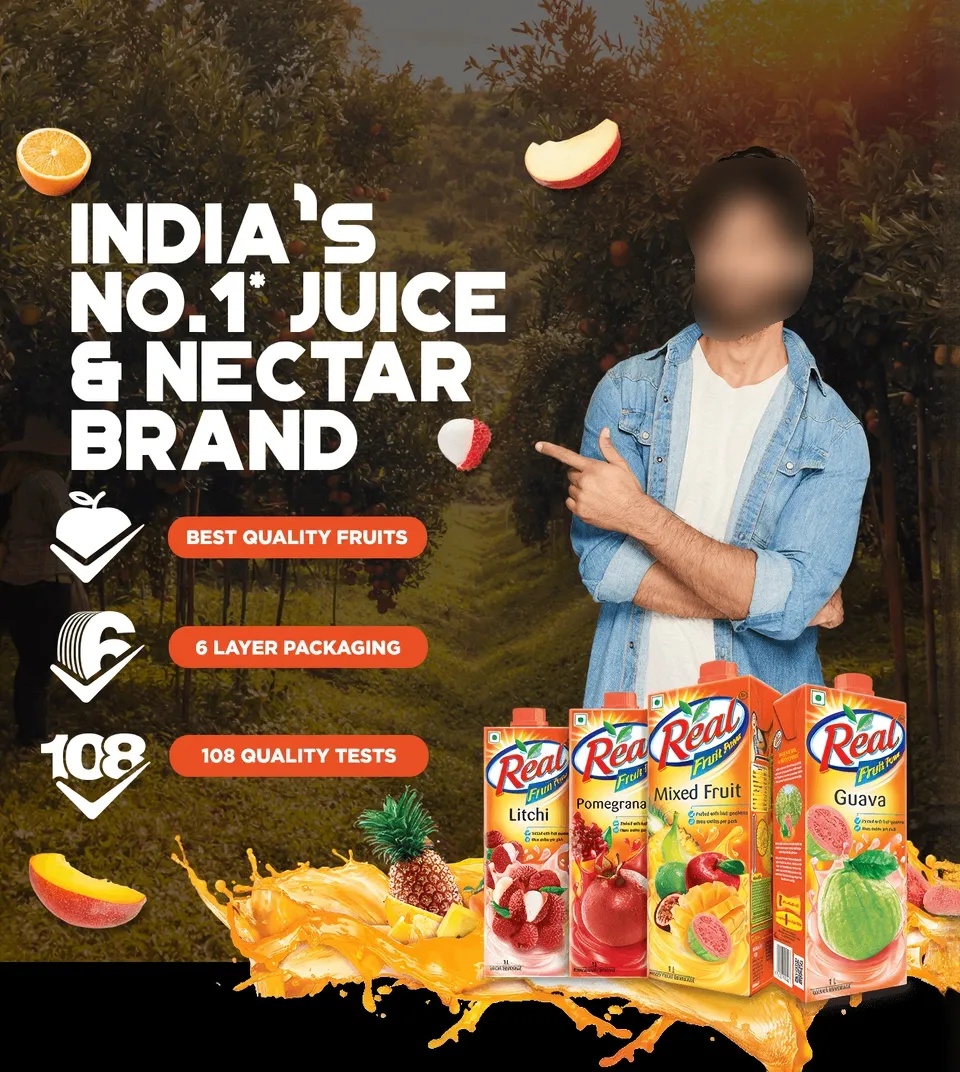} &
\includegraphics[width=0.24\linewidth,height=3.2cm]{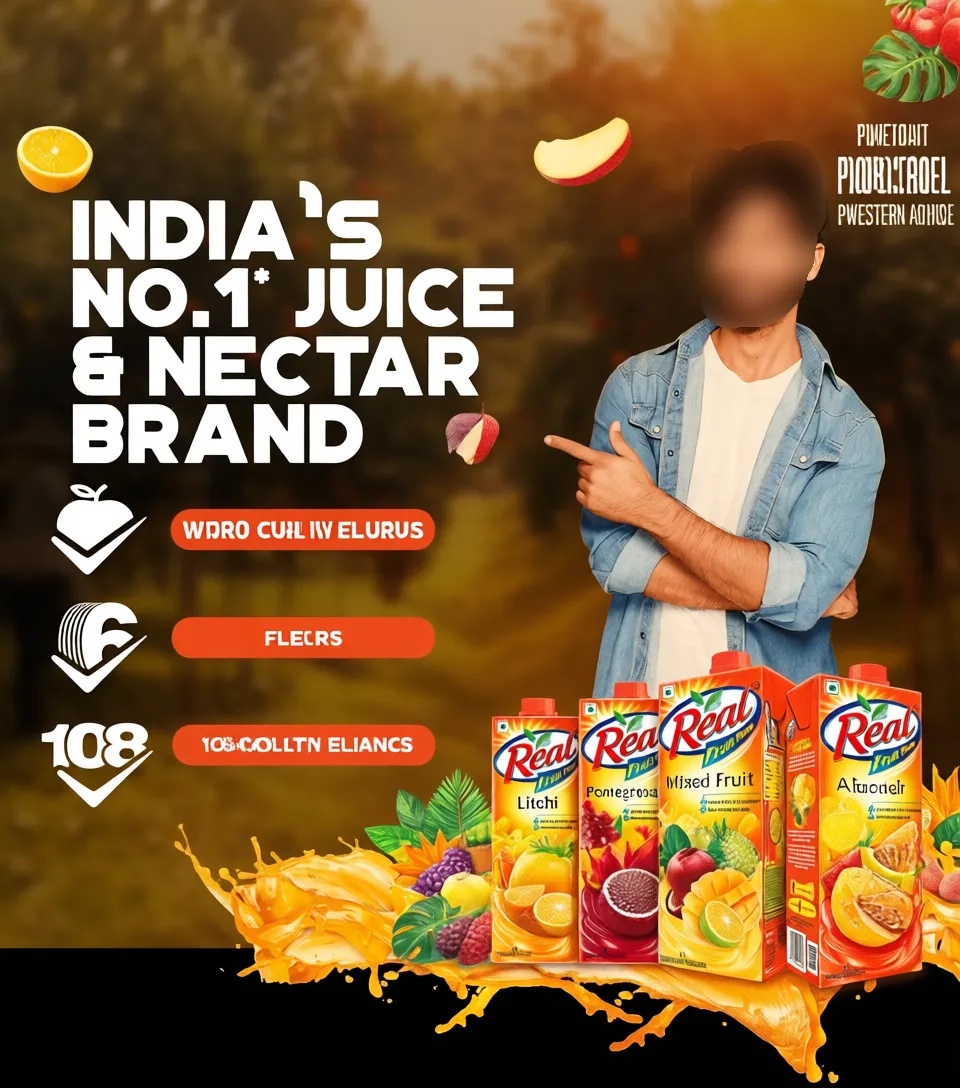} &
\includegraphics[width=0.24\linewidth,height=3.2cm]{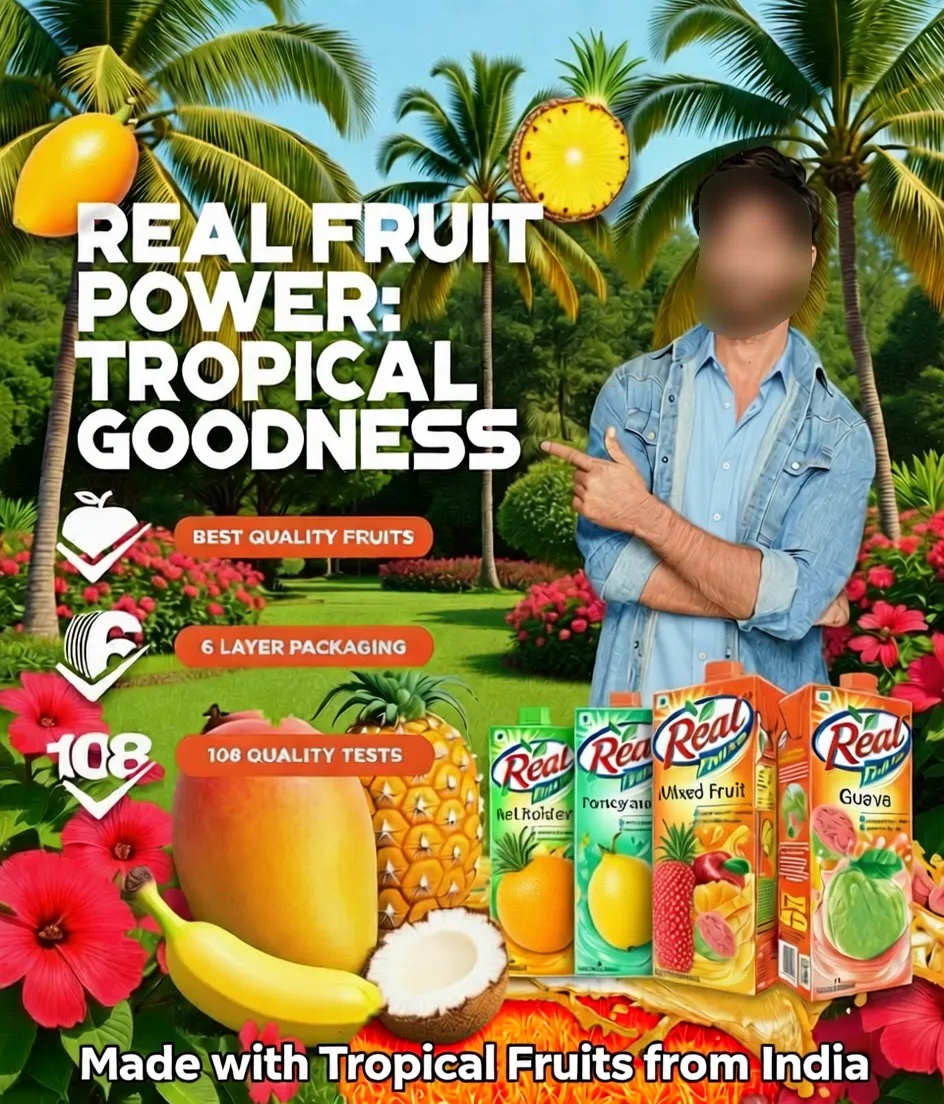} \\
\multicolumn{4}{l}{\hspace{0.5em} \fontsize{6.5pt}{7.5pt}\selectfont \textcolor{gray}{\textbf{Task:} Adapt for a Western audience emphasizing tropical fruit flavors}} \\ [8pt]

% ---- Row 3: Fitness ----
\includegraphics[width=0.24\linewidth]{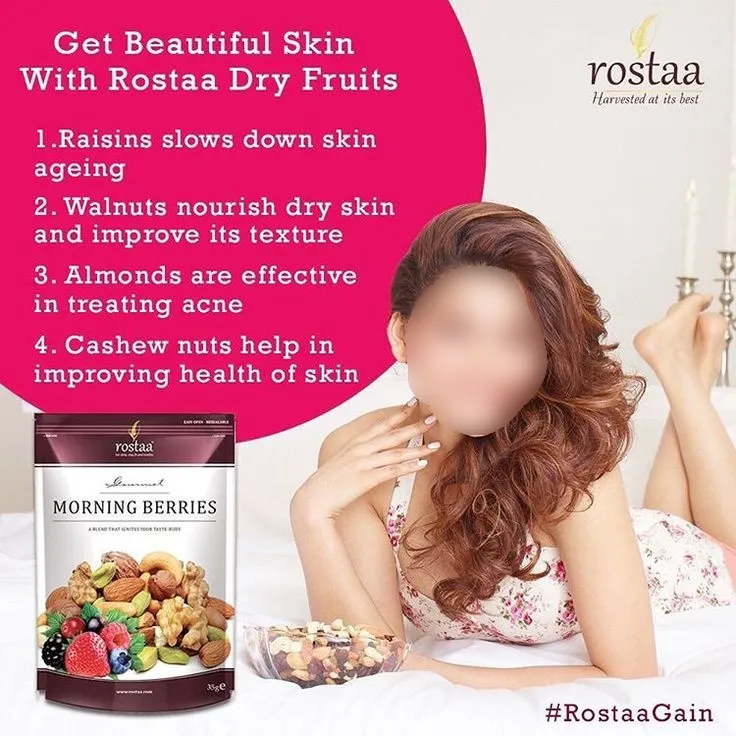} &
\includegraphics[width=0.24\linewidth]{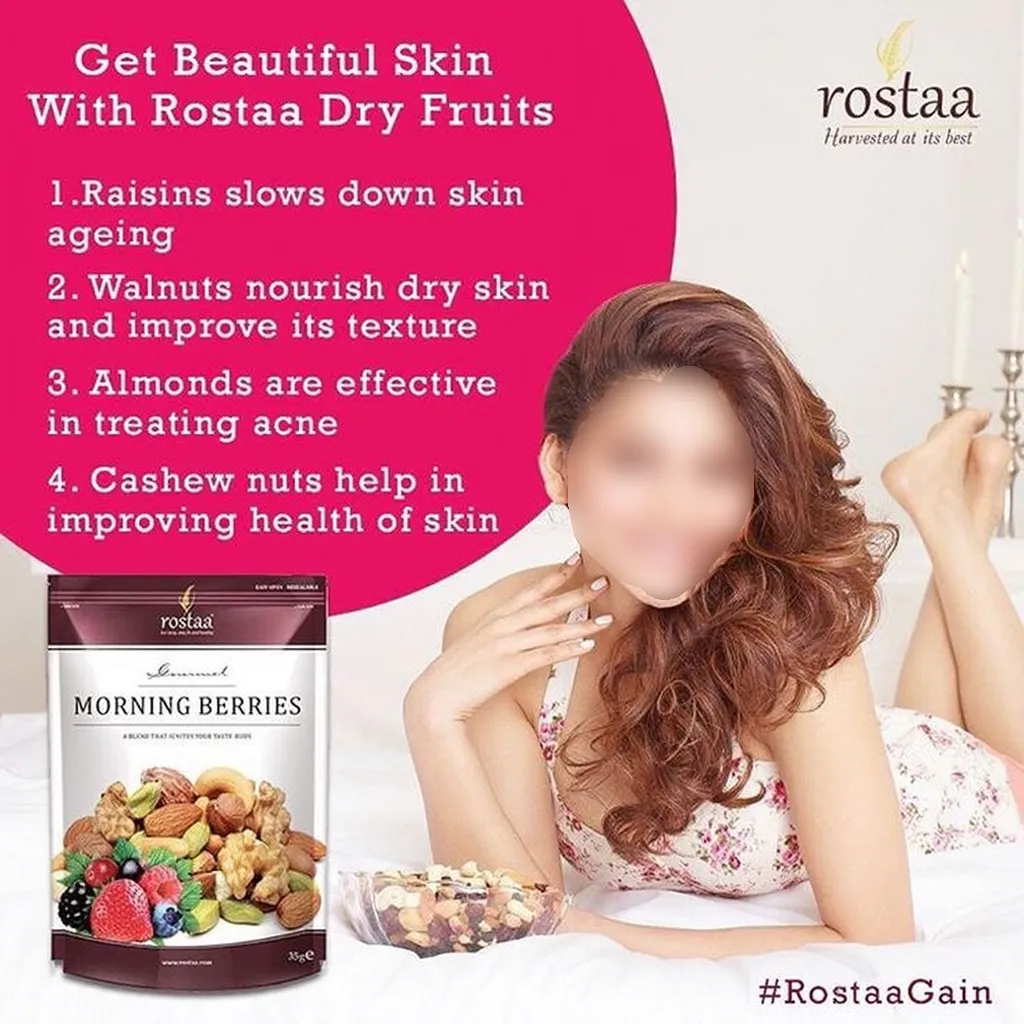} &
\includegraphics[width=0.24\linewidth,]{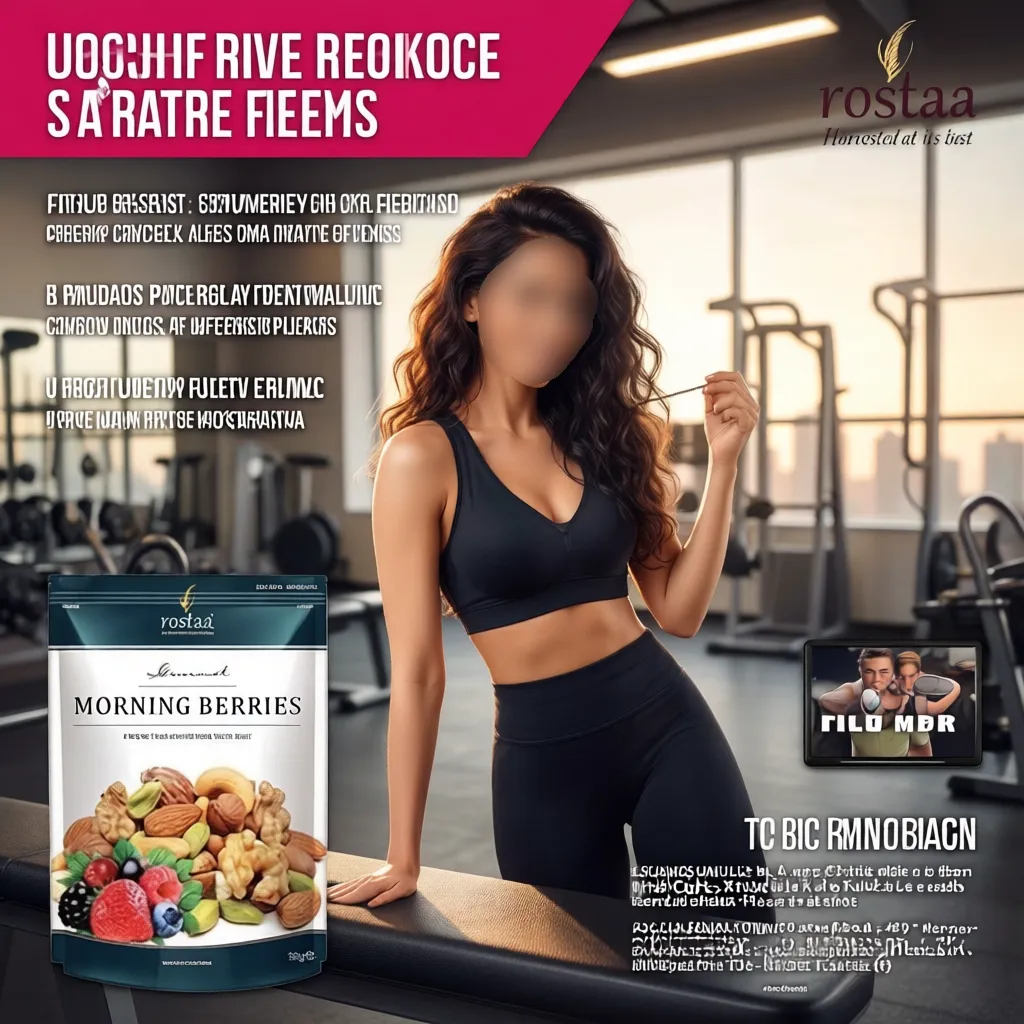} &
\includegraphics[width=0.24\linewidth]{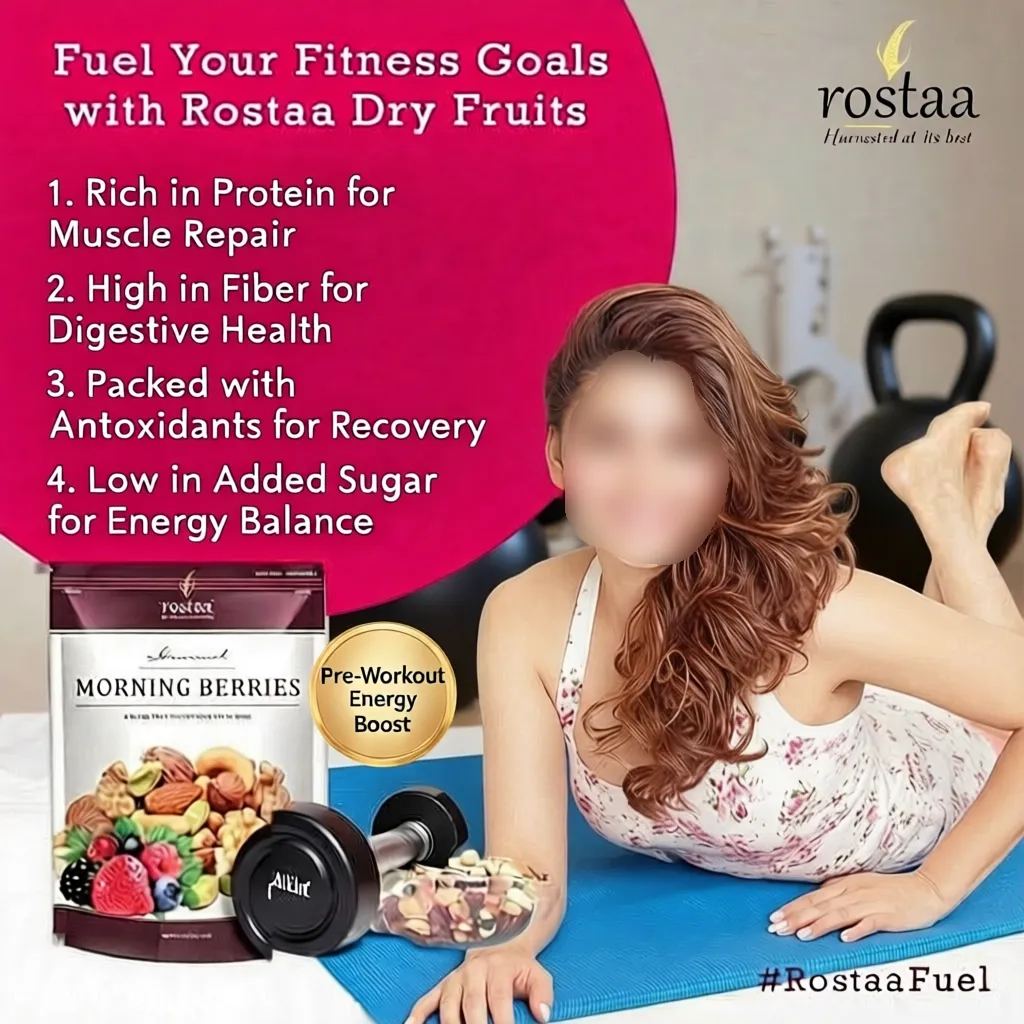} \\
\multicolumn{4}{l}{\hspace{0.5em} \fontsize{6.5pt}{7.5pt}\selectfont \textcolor{gray}{\textbf{Task:} Adapt for a fitness-conscious audience}} \\

\end{tabular}

\caption{\textbf{Qualitative results on diverse long-horizon advertisement editing tasks.} 
These examples show three challenging instructions—adapting for Lunar New Year, western audience, and for fitness-conscious audience. Our method consistently produces edits that are faithful to the instruction and globally coherent, jointly updating visual themes, textual content, layout elements, and brand messaging. Note that we blur the faces in the images to preserve the anonymity of individuals.}
\label{fig:qualitative_comparison}
\end{figure}

% \begin{figure}[!t]
% \centering
% \includegraphics[width=0.9\linewidth]{figs/comp_blur.png}
% \caption{\textbf{Qualitative results on diverse long-horizon advertisement editing tasks.} 
% These examples show four challenging instructions—adapting for eco-conscious consumers, \textcolor{purple}{rural audience, and for students}. Single-step editors (Flux Kontext~\cite{labs2025flux1kontextflowmatching} and Qwen-Image-Edit~\cite{wu2025qwenimagetechnicalreport}) often perform partial stylistic changes or introduce minimal or excessive modifications in text, layout, or branding. In contrast, our method consistently produces edits that are more instruction-faithful and globally coherent, jointly updating visual themes, textual content, layout elements, and brand messaging. 
% These results demonstrate the effectiveness of checklist-guided planning and reward-driven orchestration in handling abstract, multi-step transformations beyond localized edits.}
% \label{fig:qual_results}
% \end{figure}

\begin{figure}[t]
\centering
\includegraphics[width=\linewidth, height=0.45\textheight]{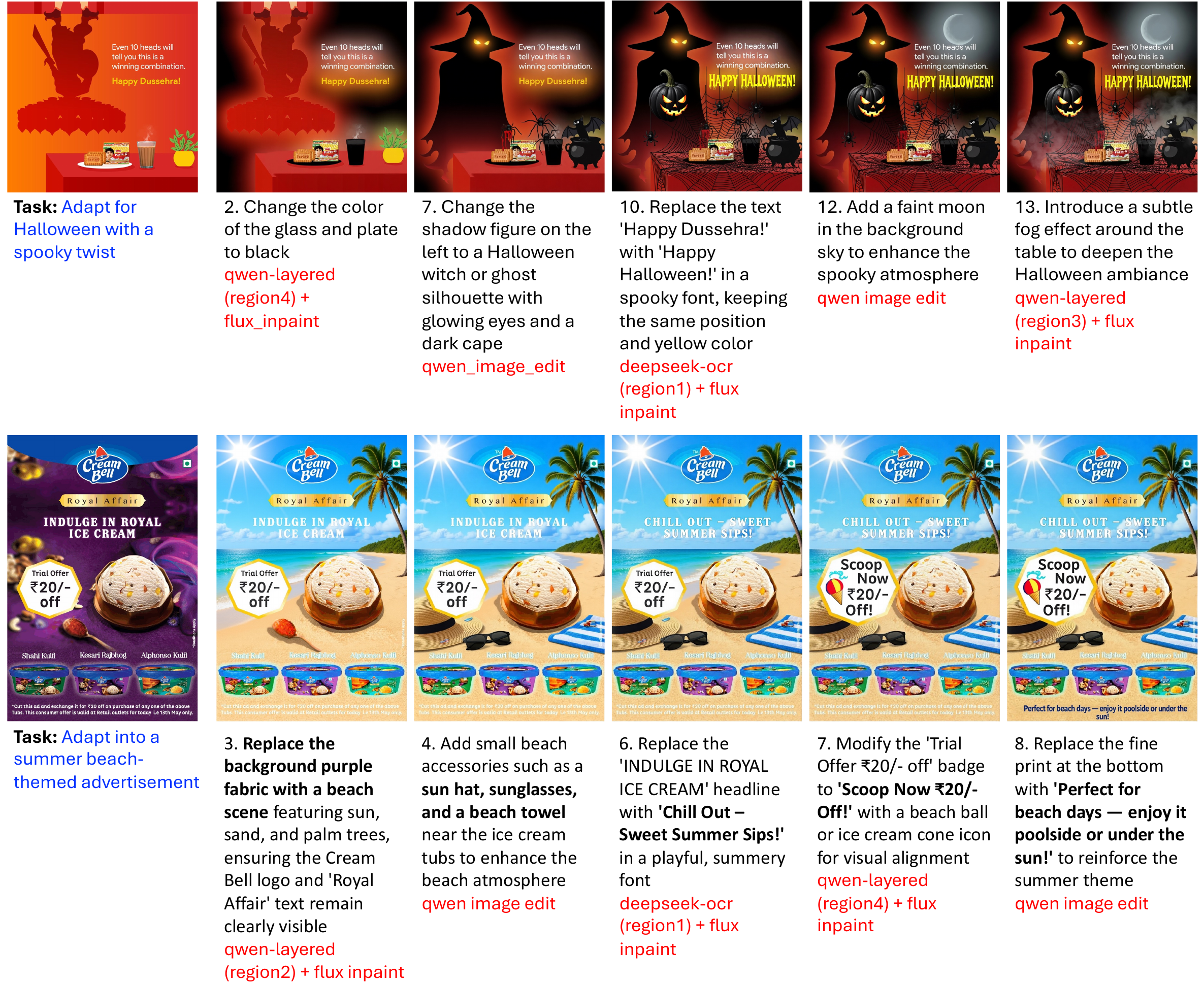}
\caption{\textbf{Planner–orchestrator subtasks and their outputs.} 
Each row begins with the input advertisement (first column) and a high-level instruction. The subsequent columns show the sequence of subtask plans and edits produced by our system. These results illustrate how our checklist-guided planning decomposes an abstract instruction into concrete atomic edits and how our orchestrator selects tool-regions to execute those edits to produce a coherent, instruction-faithful final design. Due to space constraints, we only show samples from the full sequence.} 
\label{fig:subtask-edits}
\end{figure}

\subsection{Qualitative Results}

Figures~\ref{fig:qual_results}, \ref{fig:qualitative_comparison} and~\ref{fig:subtask-edits} showcase our method on diverse long-horizon advertisement editing tasks that require coordinated updates to visuals, text, layout, and branding. Note that we blur the faces in the images to preserve the anonymity of individuals.

\vspace{-5pt}
\paragraph{Long-Horizon Adaptation.}
 Figures~ \ref{fig:qual_results} and \ref{fig:qualitative_comparison} presents challenging transformations, including adapting for business travelers, American Independence Day etc. These tasks involve coupled changes to background, color palette, slogans, badges, and overall branding. Single-step editors often produce partial or excessive modifications that disrupt layout consistency or identity preservation. In contrast, our method generates more globally coherent, instruction-faithful edits while maintaining better brand consistency. 

\vspace{-5pt}
\paragraph{Planner–Orchestrator Decomposition.}
Figure~\ref{fig:subtask-edits} illustrates how checklist-guided planning decomposes abstract instructions into atomic subtasks that are sequentially executed via reward-driven orchestration. This structured decomposition enables comprehensive coverage and coherent multi-step transformations, rather than isolated local edits.

Overall, these results demonstrate that coupling on-policy planning with outcome-driven orchestration enables robust handling of abstract, open-ended image editing tasks beyond the capabilities of single-step or rule-based multi-tool pipelines.

Appendix~\ref{app:qualres} presents additional qualitative step-by-step visualizations.

% \begin{table}[t]
% \centering
% \begin{tabular}{l c}
% \toprule
% \textbf{Setting} & \textbf{Avg. Max Reward} \\
% \midrule
% Before filtering & 4.1708 \\
% After filtering  & 4.3095 \\
% \bottomrule
% \end{tabular}
% \caption{Average maximum achievable sub-task reward (Gemini 3 Pro judge) before and after plan refinement.}
% \label{tab:refinement}
% \end{table}

%\kr{MAYBE IN SUPP WE CAN SHOW FULL SEQUENCE?} Yep}

% \subsection{Additional Results}

% Please see supplementary for additional quantitative and qualitative results.
%\kr{WE DON'T NEED SEPARATE SECTION, WE CAN MENTION IT AT START OF EXPERIMENT SECTION?}

%\vspace{-15pt}

\section{Conclusion}
%\vspace{-15pt}

We presented an experiential framework for long-horizon, open-ended image editing. By combining checklist-guided, on-policy planning with reward-driven orchestration, our approach moves beyond handcrafted pipelines and single-step generation. The planner learns structured decompositions, while the orchestrator selects tools and regions from outcome-based feedback, with closed-loop refinement aligning plans with executable actions.

Extensive experiments and user studies show that our method produces more coherent, instruction-faithful edits than strong single-step and rule-based multi-step baselines, highlighting the value of coupling planning with experiential learning for abstract, multi-step editing tasks.

\paragraph{\textbf{Acknowledgements.}} We thank Scott Cohen for his technical feedback and support throughout the project. We thank Eslam Abdelrahman for valuable discussions and suggestions on evaluating image editing systems and designing the evaluation. We also thank Sicheng Mo for his help with evaluation design, and Zhaowen Wang for discussions regarding image editing tools.

\paragraph{\textbf{Disclaimer.}} All trademarks and copyrighted images are the property of their respective owners and are used here for identification and descriptive purposes only. No affiliation, sponsorship, or endorsement is implied.

\bibliographystyle{splncs04}
\bibliography{main}

\clearpage
\appendix
% \nolinenumbers

\section*{Appendix}

We provide additional qualitative results, experimental comparisons, and implementation details below.

\section{Qualitative Results}\label{app:qualres}
% In Figure~\ref{fig:qualitative_comparison}, we present additional qualitative results for editing images from the MadVerse dataset. We compare our method with Qwen-Image-Edit and Flux Kontext. When provided with high-level editing plans, Qwen-Image-Edit often exhibits noticeable degradation in visual quality or layout consistency, making excessive edits. In contrast, Flux Kontext tends to apply only minimal modifications, frequently failing to implement the full set of requested edits. As a result, both approaches struggle to faithfully execute complex multi-step instructions. Our method, however, produces more globally coherent and instruction-faithful edits while maintaining stronger brand consistency.

We provide step-wise editing visualizations in Tables~\ref{tab:millennial_savings_pipeline}, \ref{tab:mexican_spice_shortened}, and \ref{tab:sustainable_car}. These examples illustrate our method’s ability to execute long sequences of diverse edits, including text changes, background modifications, and object-level alterations. Together, they demonstrate that the system can reliably compose multiple heterogeneous edits while maintaining visual coherence and consistency with the original content.

% --- Inside your document ---

\begin{table}[htbp] 
\centering
\small
\renewcommand{\arraystretch}{2.2} % Vertical breathing
\setlength{\tabcolsep}{12pt}

\begin{tabular}{>{\centering\arraybackslash}m{0.32\linewidth}  m{0.62\linewidth}}
\toprule
\textbf{Execution Result} & \textbf{Plan} \\ \midrule

% --- Input Block (The only one followed by a line) ---
\includegraphics[width=\linewidth, height=3.5cm, keepaspectratio]{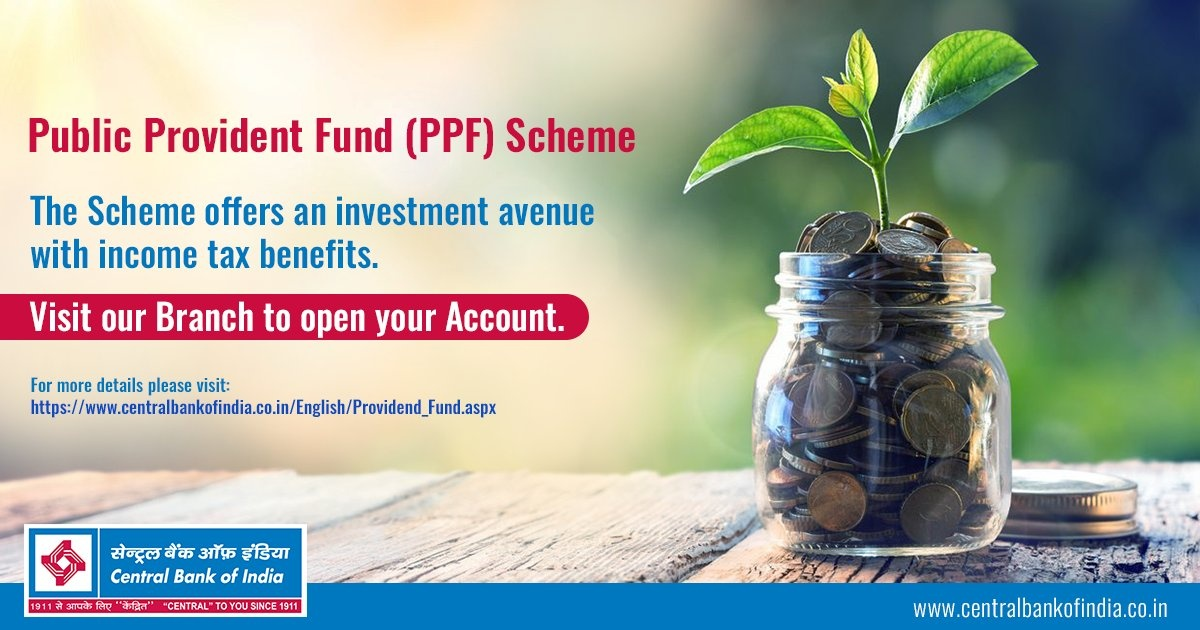} & 
\textbf{Goal:} Adapt the advertisement into a vibrant millennial-focused savings campaign while maintaining brand recognition. \\ \midrule

% --- Step 1 ---
\includegraphics[width=\linewidth, height=3.5cm, keepaspectratio]{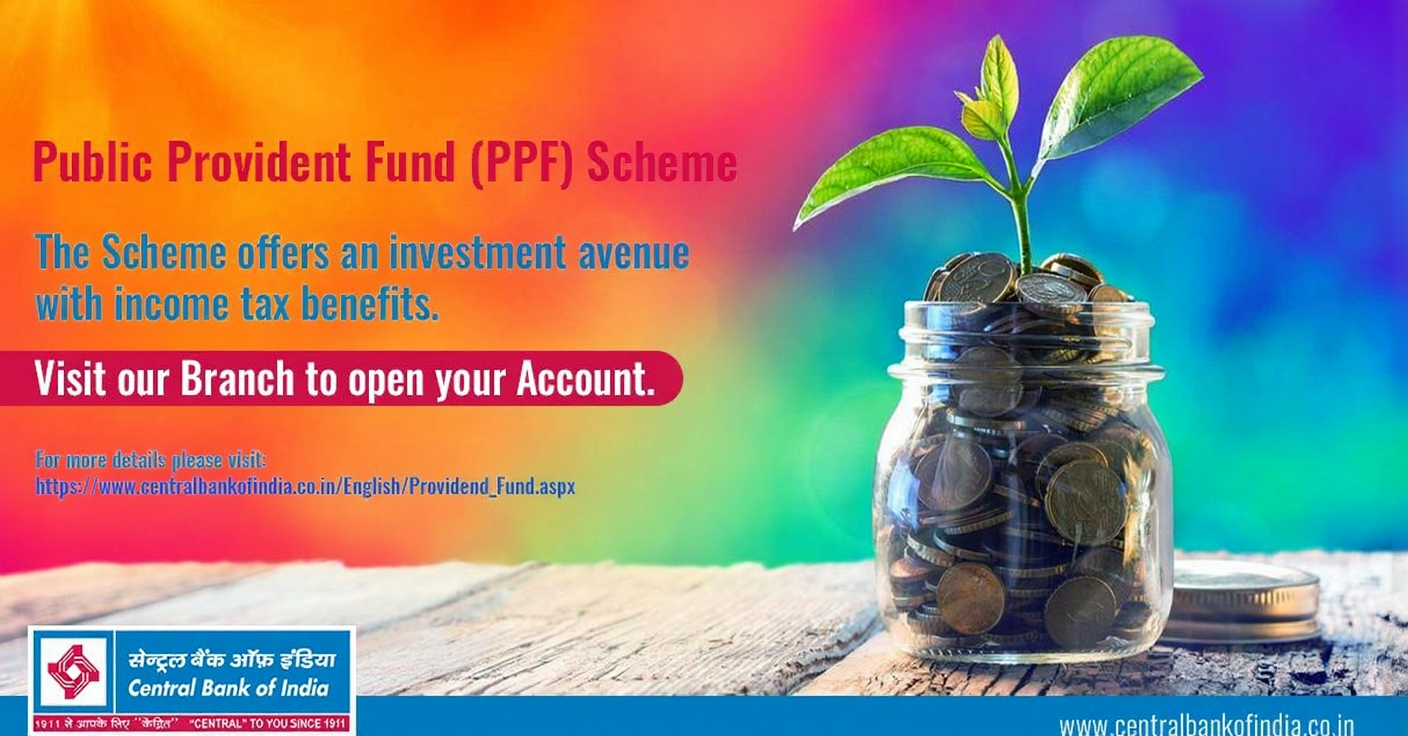} & 
\textbf{1:} Replace the background with a vibrant gradient of orange, teal, and purple, keeping the jar of coins and plant unchanged. \newline
\textcolor{red}{\fontsize{7pt}{8pt}\selectfont flux kontext edit} \\

% --- Step 2 ---
\includegraphics[width=\linewidth, height=3.5cm, keepaspectratio]{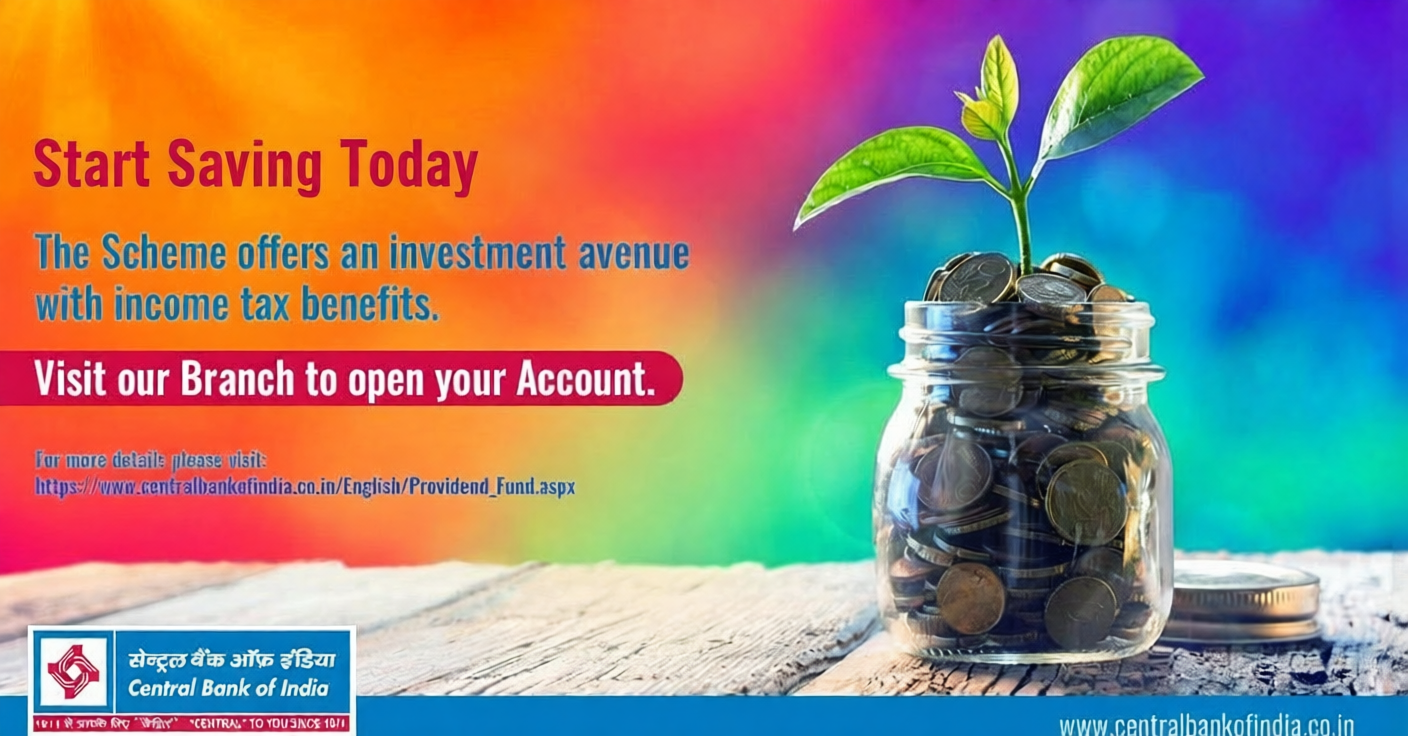} & 
\textbf{2:} Replace the text `Public Provident Fund (PPF) Scheme' with `Start Saving Today' in bold, modern sans-serif font, centered in the top-left area. \newline
\textcolor{red}{\fontsize{7pt}{8pt}\selectfont qwen image edit} \\

% --- Step 3 ---
\includegraphics[width=\linewidth, height=3.5cm, keepaspectratio]{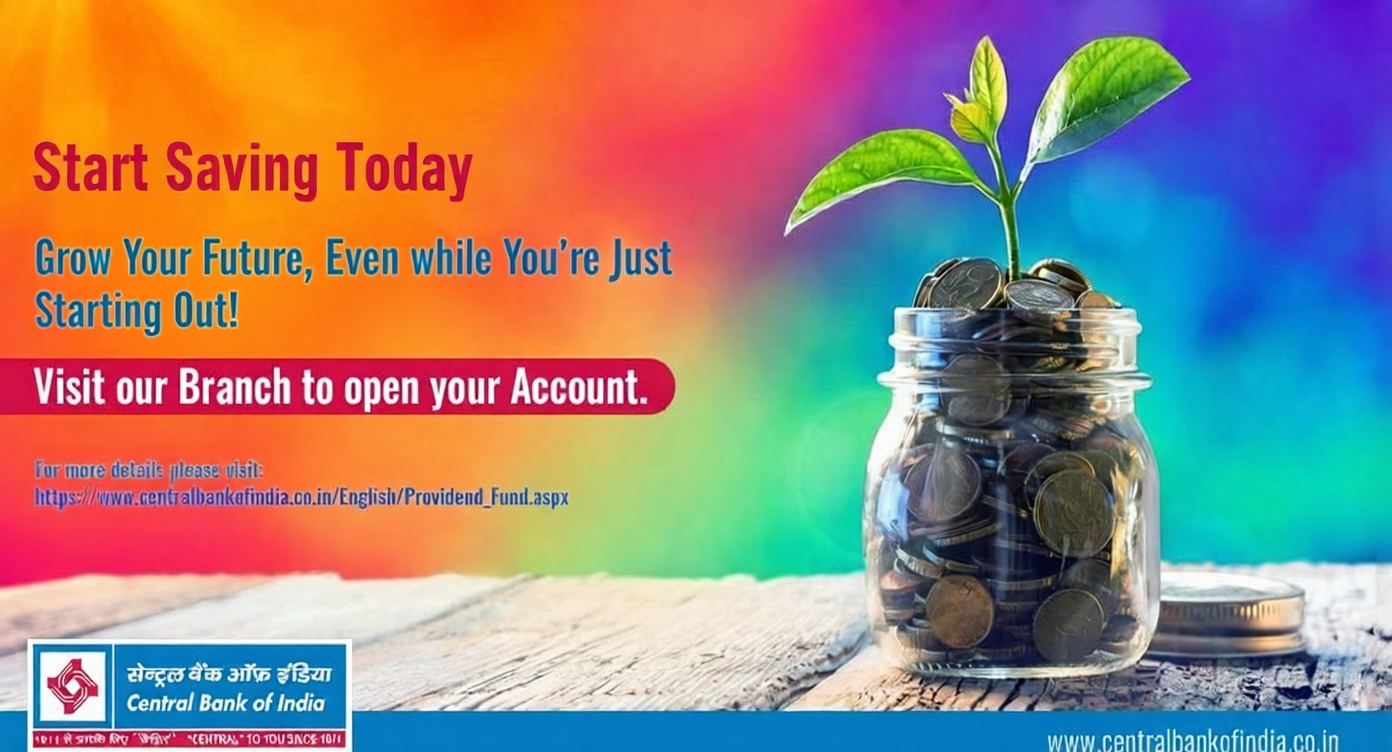} & 
\textbf{3:} Replace the text `The Scheme offers an investment avenue...' with `Grow Your Future, Even While You're Just Starting Out!' in bold, modern sans-serif font. \newline
\textcolor{red}{\fontsize{7pt}{8pt}\selectfont qwen-layered (region3) + flux inpaint} \\

% --- Step 8 ---
\includegraphics[width=\linewidth, height=3.5cm, keepaspectratio]{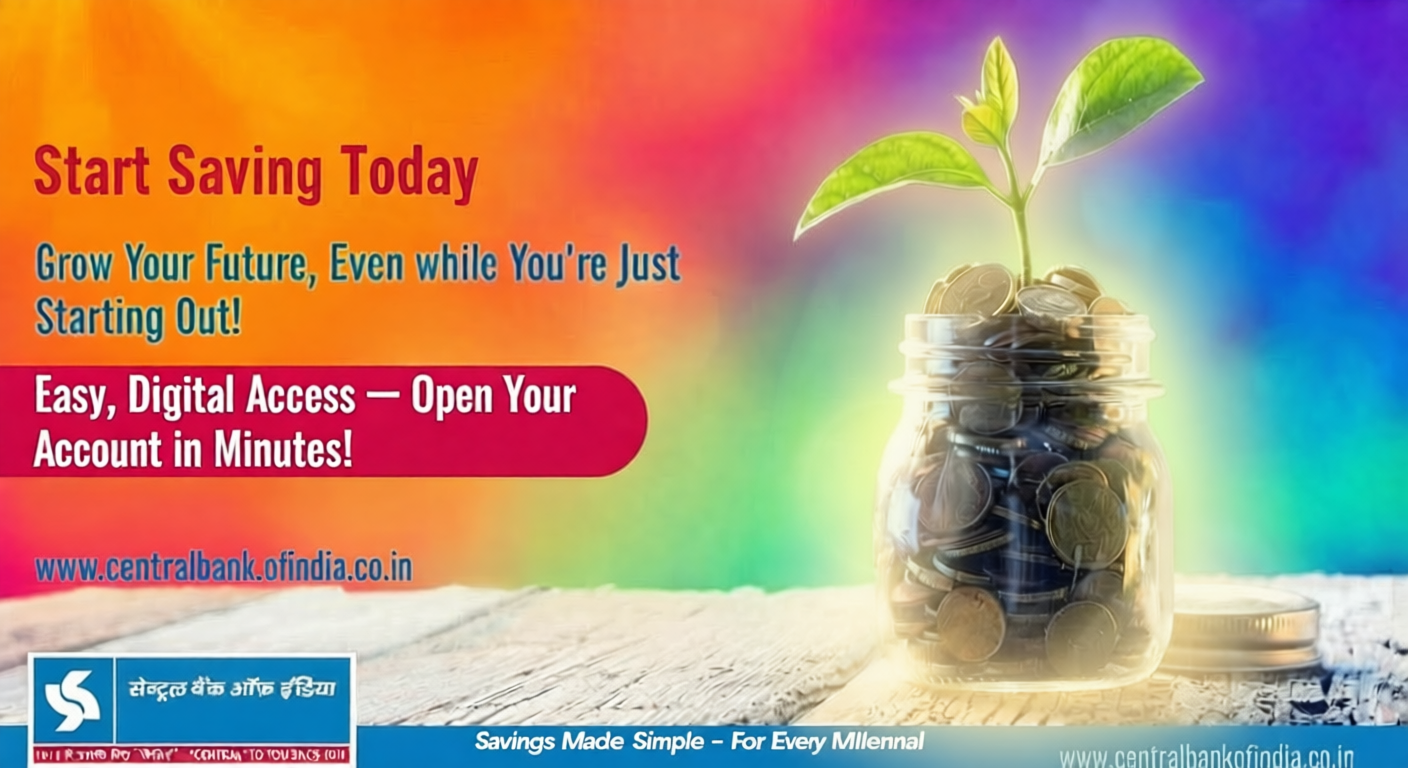} & 
\textbf{8:} Add a subtle glow effect around the jar of coins and plant to draw attention to them while preserving all other elements. \newline
\textcolor{red}{\fontsize{7pt}{8pt}\selectfont qwen image edit} \\

% --- Step 9 ---
\includegraphics[width=\linewidth, height=3.5cm, keepaspectratio]{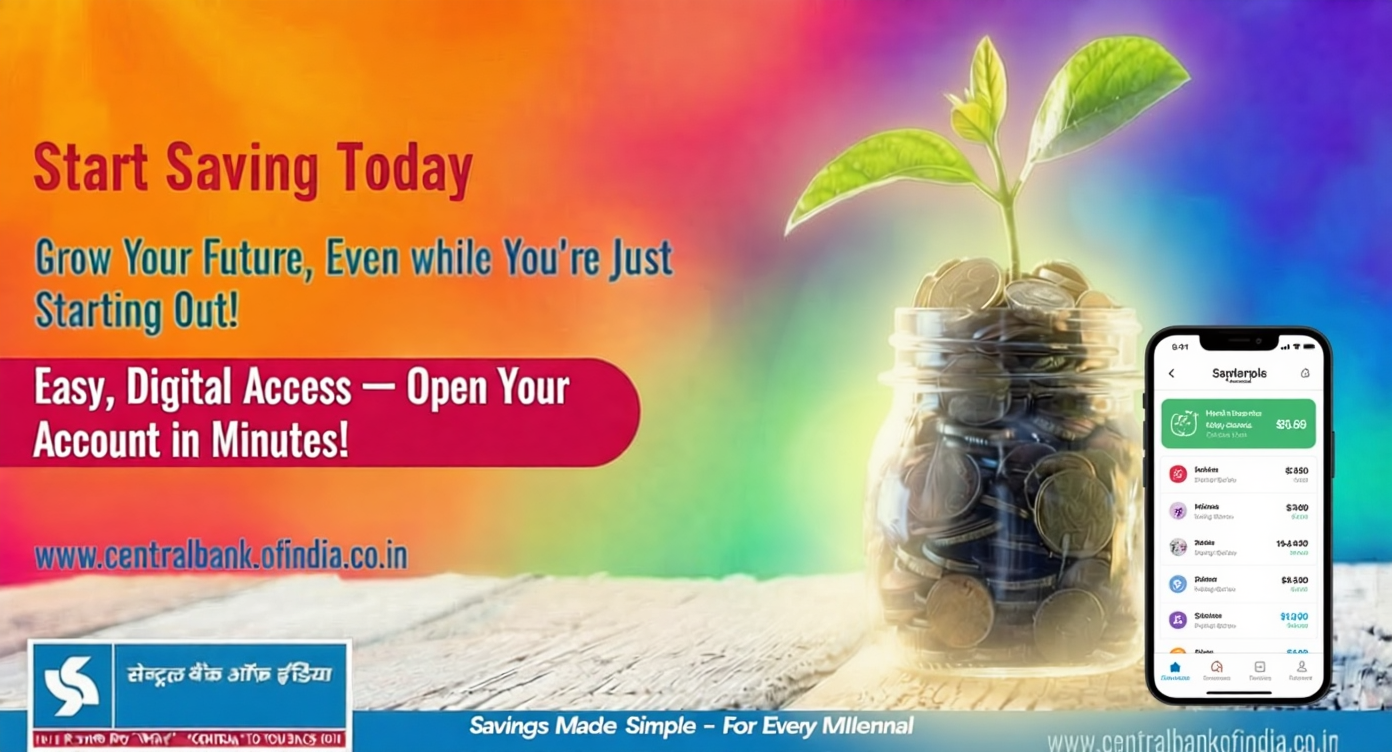} & 
\textbf{9:} Include a small, clean smartphone screen graphic in the bottom-right corner showing a simple savings app interface to reinforce digital accessibility. \newline
\textcolor{red}{\fontsize{7pt}{8pt}\selectfont qwen-layered (region1) + flux inpaint} \\ 

\bottomrule
\end{tabular}

\caption{Example editing sequence showcasing the transition from the original advertisement to the final millennial-focused design. Our planner decomposes the task into steps, and the orchestrator carries out each step by selecting the appropriate editing tools.}
\label{tab:millennial_savings_pipeline}
\end{table}
\begin{table}[htbp] 
\centering
\small
\renewcommand{\arraystretch}{1.4} % Reduced from 2.2 to further tighten vertical space
\setlength{\tabcolsep}{12pt}

\begin{tabular}{>{\centering\arraybackslash}m{0.30\linewidth}  m{0.64\linewidth}}
\toprule
\textbf{Execution Result} & \textbf{Plan} \\ \midrule

% --- Input Block ---
\includegraphics[width=\linewidth, height=2.5cm, keepaspectratio]{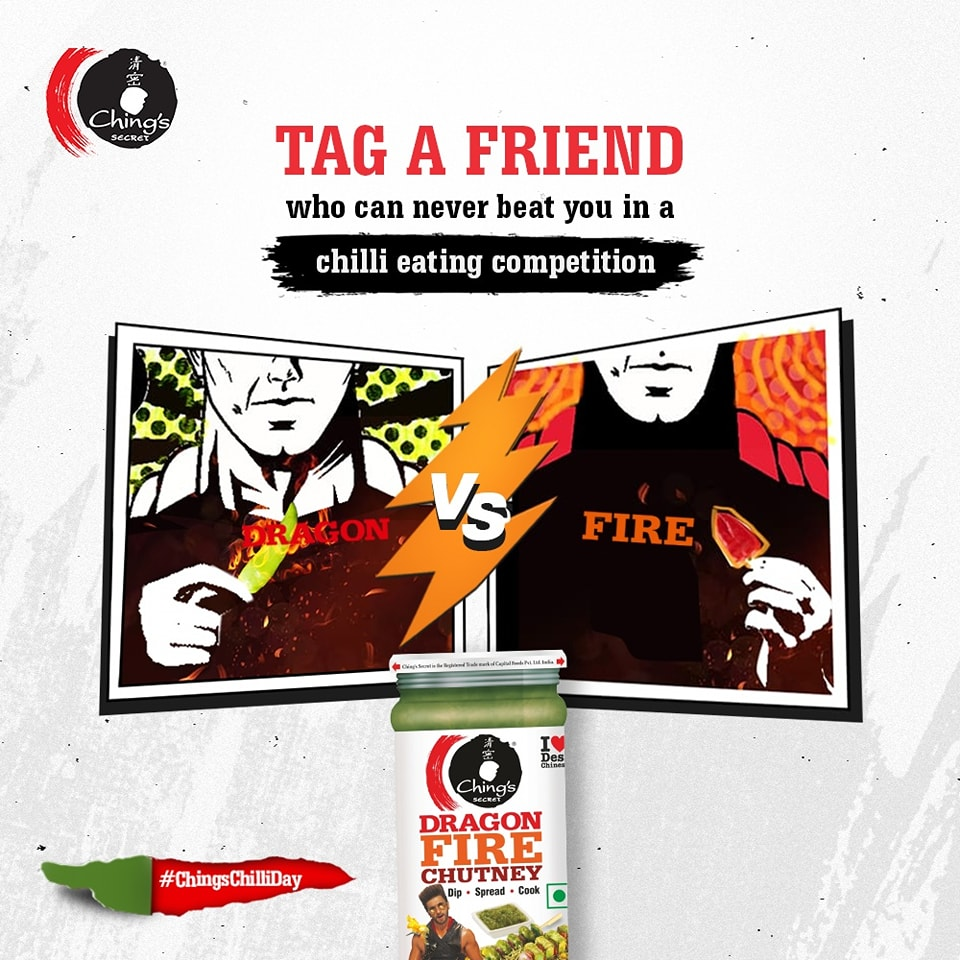} & 
\textbf{Goal:} Adapt the advertisement into a localized version targeted at spice lovers in Mexico. \\ \midrule

% --- Step 1 ---
\includegraphics[width=\linewidth, height=2.5cm, keepaspectratio]{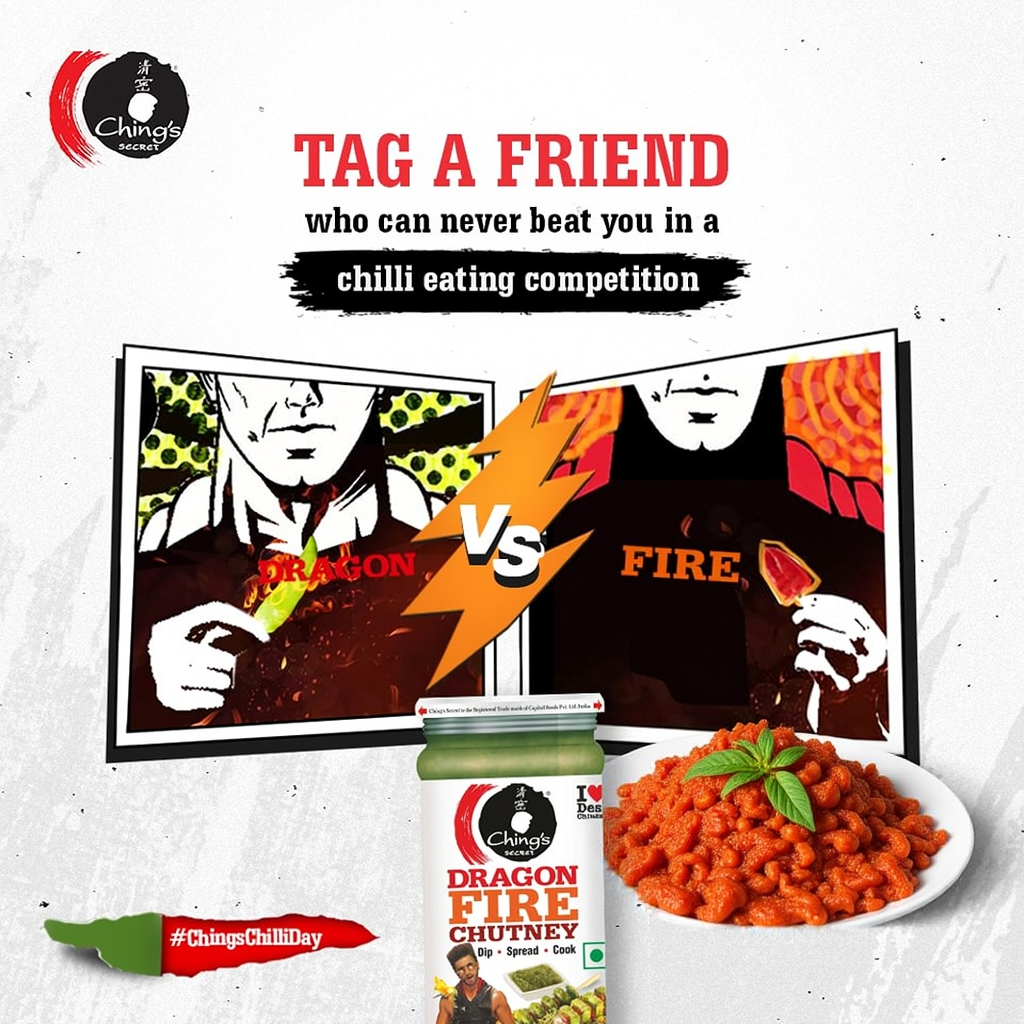} & 
\textbf{1:} Add a spicy dish on the right side of the image, placing it next to the product, without altering the product or existing text. \newline
\textcolor{red}{\fontsize{7pt}{8pt}\selectfont qwen-bbox (region3) + flux inpaint} \\

% --- Step 3 ---
\includegraphics[width=\linewidth, height=2.5cm, keepaspectratio]{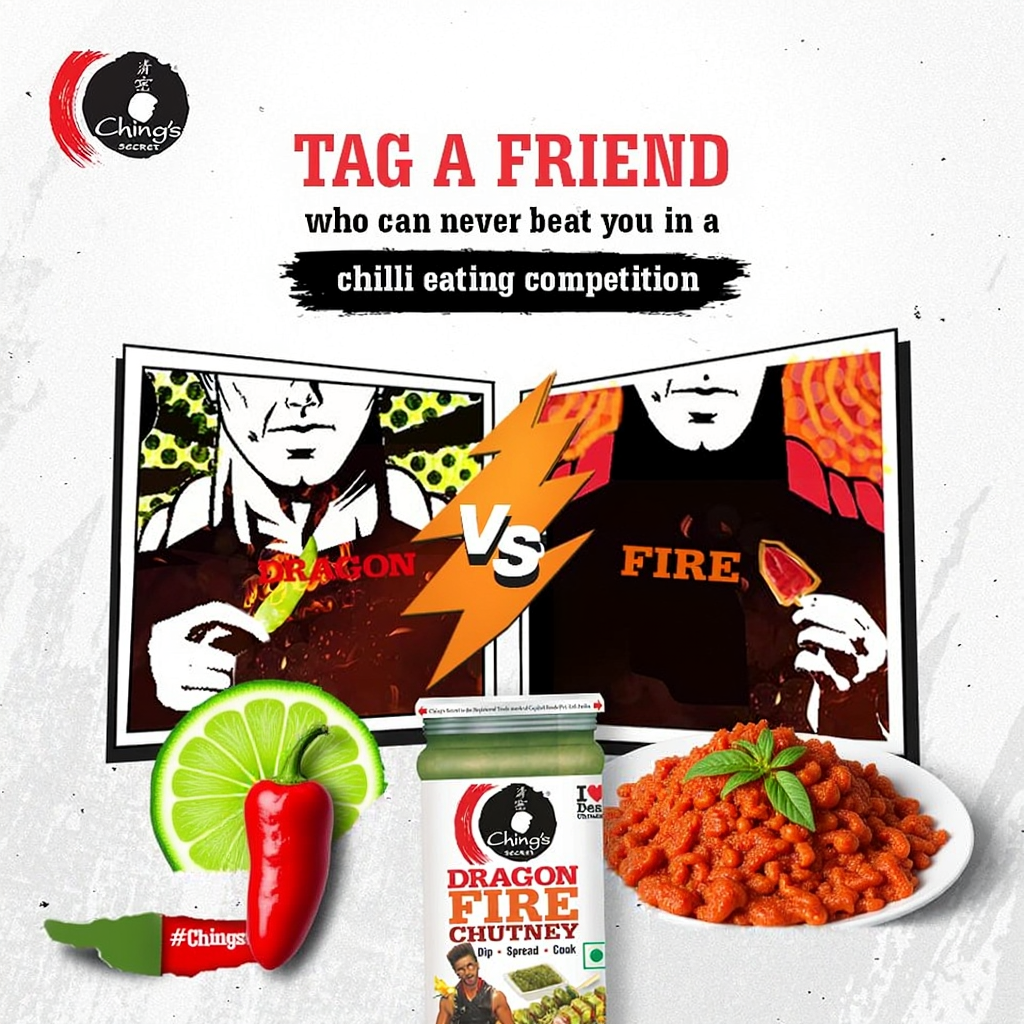} & 
\textbf{3:} Add a small image of a traditional Mexican chili pepper, such as a jalapeño or habanero, near the bottom center of the image, below the product\newline
\textcolor{red}{\fontsize{7pt}{8pt}\selectfont qwen-bbox (region2) + flux inpaint} \\

% --- Step 4 ---
\includegraphics[width=\linewidth, height=2.5cm, keepaspectratio]{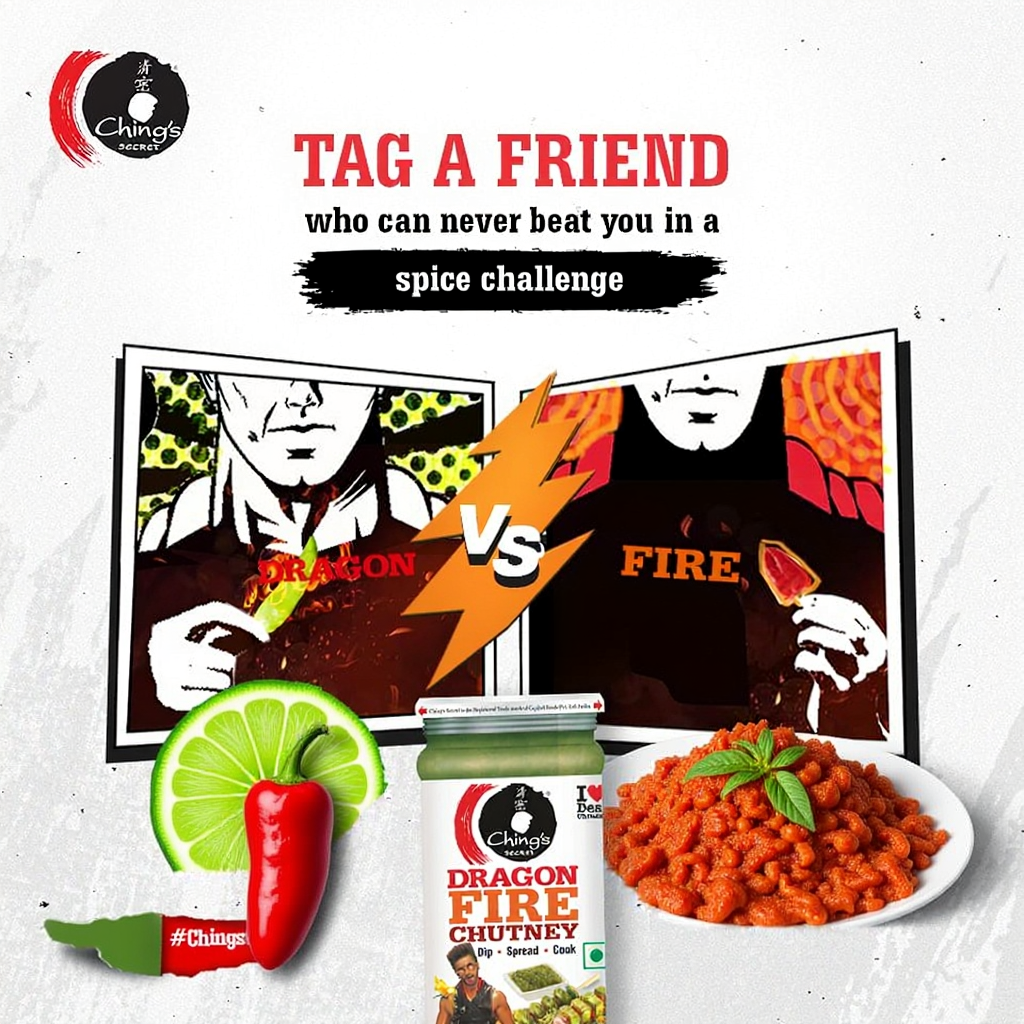} & 
\textbf{4:} Replace `chilli eating competition' with `spice challenge' in the black banner while preserving design elements. \newline
\textcolor{red}{\fontsize{7pt}{8pt}\selectfont deepseek-ocr (region4) + flux inpaint} \\

% --- Step 6 ---
\includegraphics[width=\linewidth, height=2.5cm, keepaspectratio]{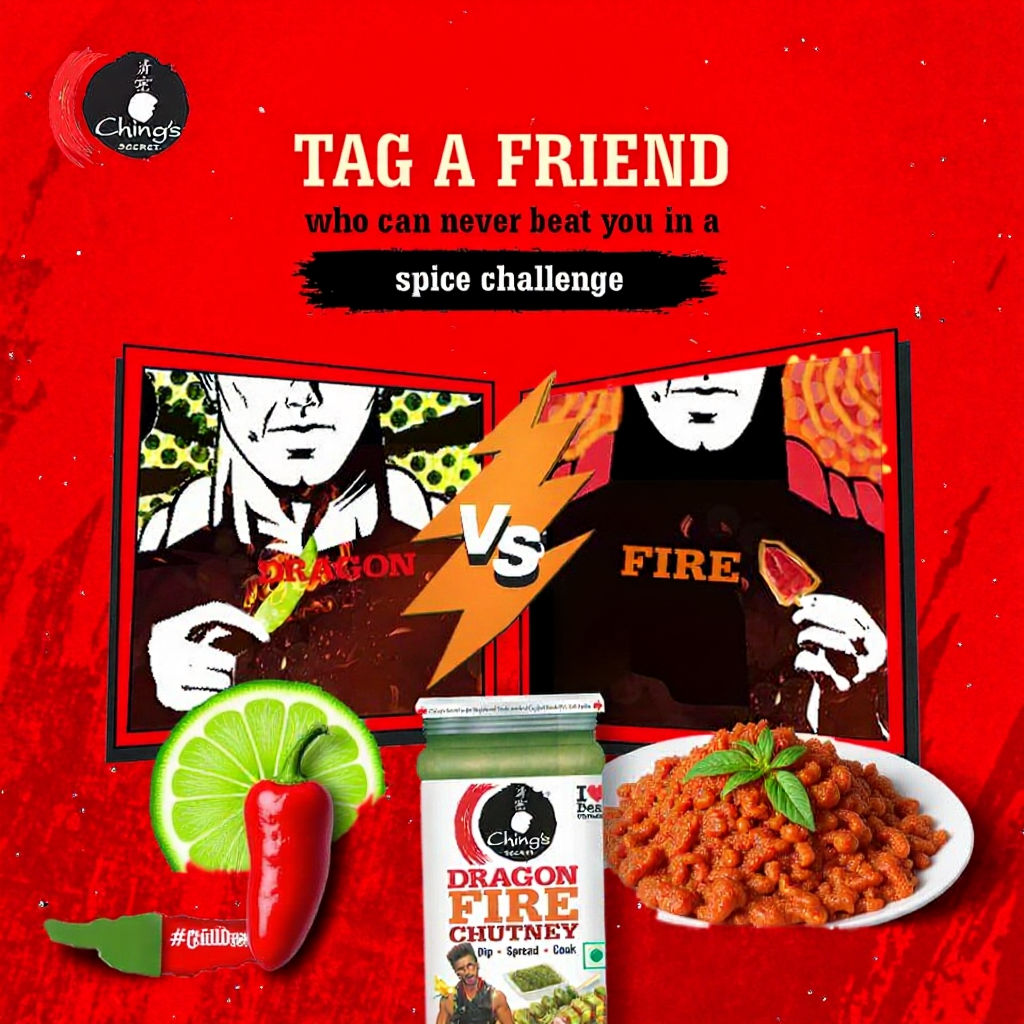} & 
\textbf{6:} Change the background to a vibrant red with subtle spice-themed patterns (chili outlines/smoke), keeping product and text in place. \newline
\textcolor{red}{\fontsize{7pt}{8pt}\selectfont qwen-layered (region1) + flux inpaint} \\

% --- Step 7 ---
\includegraphics[width=\linewidth, height=2.5cm, keepaspectratio]{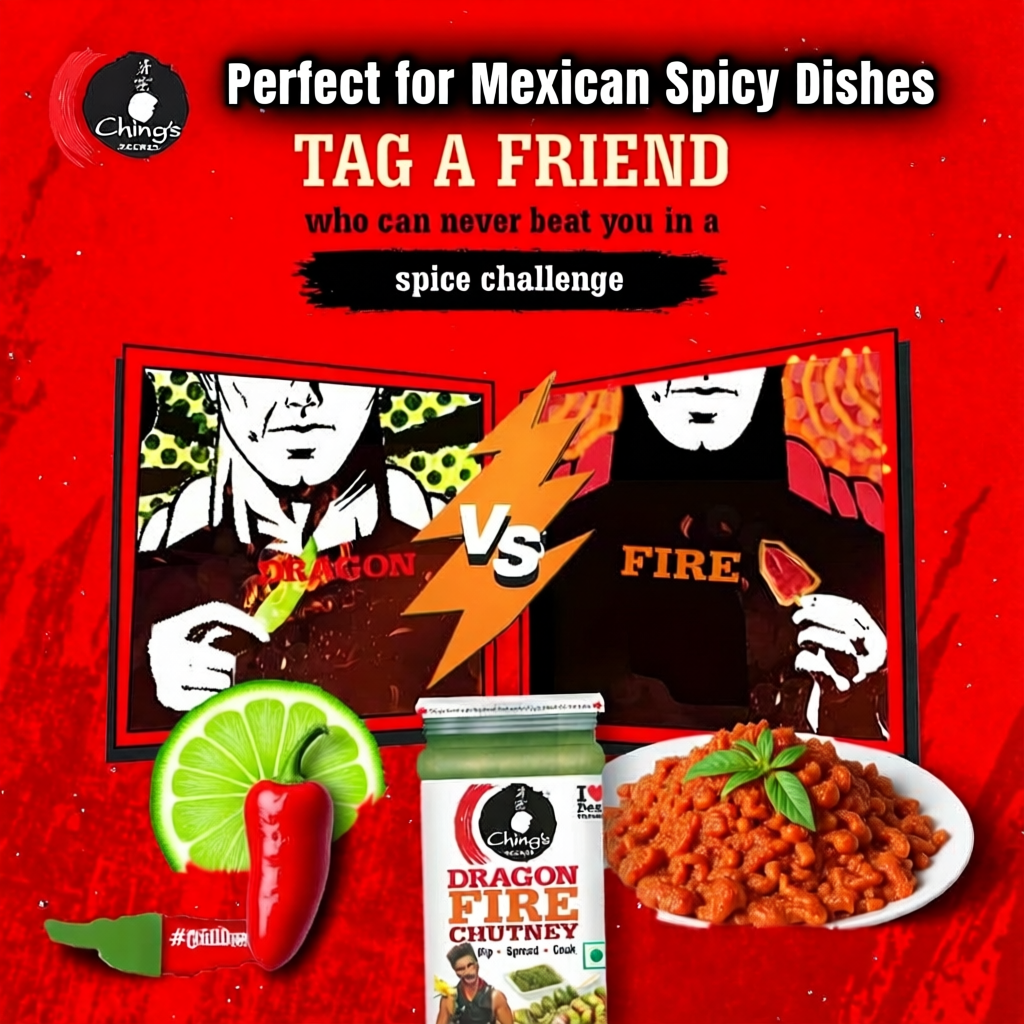} & 
\textbf{7:} Add bold white text `Perfect for Mexican Spicy Dishes' at the top of the image, positioned above the `TAG A FRIEND' headline. \newline
\textcolor{red}{\fontsize{7pt}{8pt}\selectfont qwen image edit} \\

% % --- Step 9 ---
% \includegraphics[width=\linewidth, height=2.5cm, keepaspectratio]{figs/supplementary/task_046/step_9_qwenedit.png} & 
% \textbf{9:} Replace comic panel faces with Mexican-themed figures (e.g., wearing sombreros) while keeping `Dragon' and `Fire' labels intact. \newline
% \textcolor{red}{\fontsize{7pt}{8pt}\selectfont qwen image edit} \\ 

\bottomrule
\end{tabular}

\caption{Example editing sequence showcasing the transition from the original advertisement to a localized version targeted at spice lovers in Mexico. Our planner decomposes the task into steps, and the orchestrator carries out each step by selecting the appropriate editing tools.}
\label{tab:mexican_spice_shortened}
\end{table}

\begin{table}[htbp] 
\centering
\small
\renewcommand{\arraystretch}{1} 
\setlength{\tabcolsep}{12pt}

\begin{tabular}{>{\centering\arraybackslash}m{0.32\linewidth}  m{0.62\linewidth}}
\toprule
\textbf{Execution Result} & \textbf{Plan} \\ \midrule

% --- Input Block ---
\includegraphics[width=\linewidth, height=2.7cm, keepaspectratio]{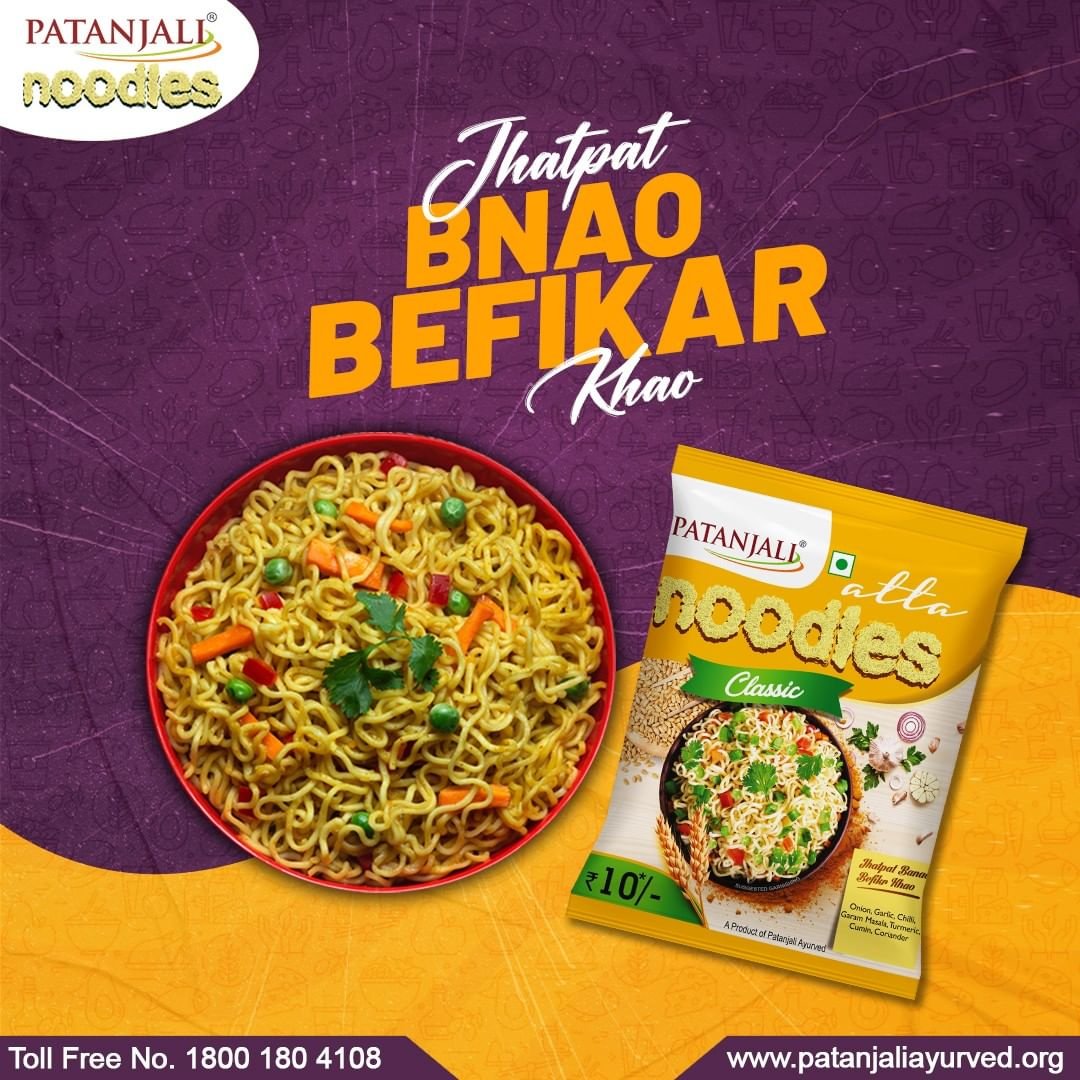} & 
\textbf{Goal:} Create a healthy lifestyle version targeting fitness-conscious buyers. \\ \midrule

% --- Step 1 ---
\includegraphics[width=\linewidth, height=2.7cm, keepaspectratio]{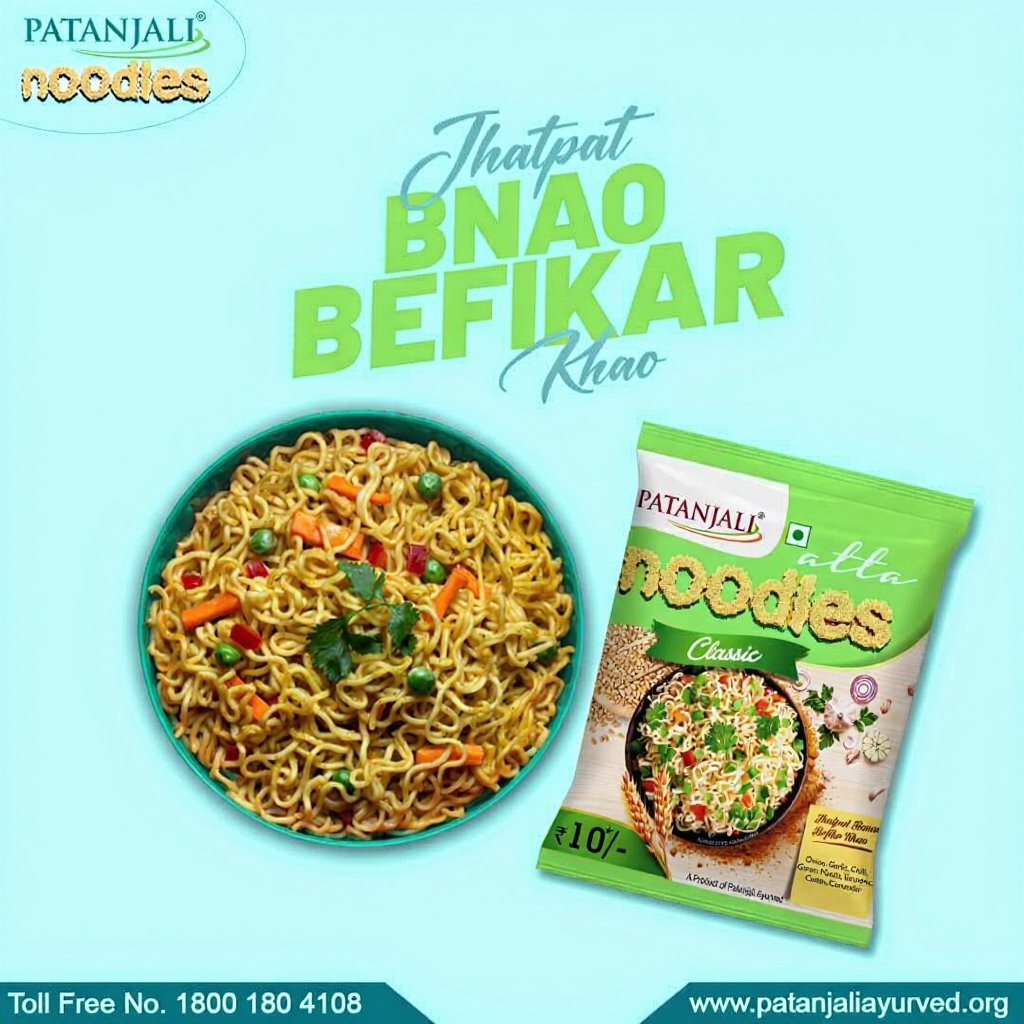} & 
\textbf{2:} Change the color palette to light blue, green, and earthy tones across the entire image, including the noodles bowl, packaging, and background. \newline
\textcolor{red}{\fontsize{7pt}{8pt}\selectfont qwen-bbox (region1) + flux inpaint} \\

\includegraphics[width=\linewidth, height=2.7cm, keepaspectratio]{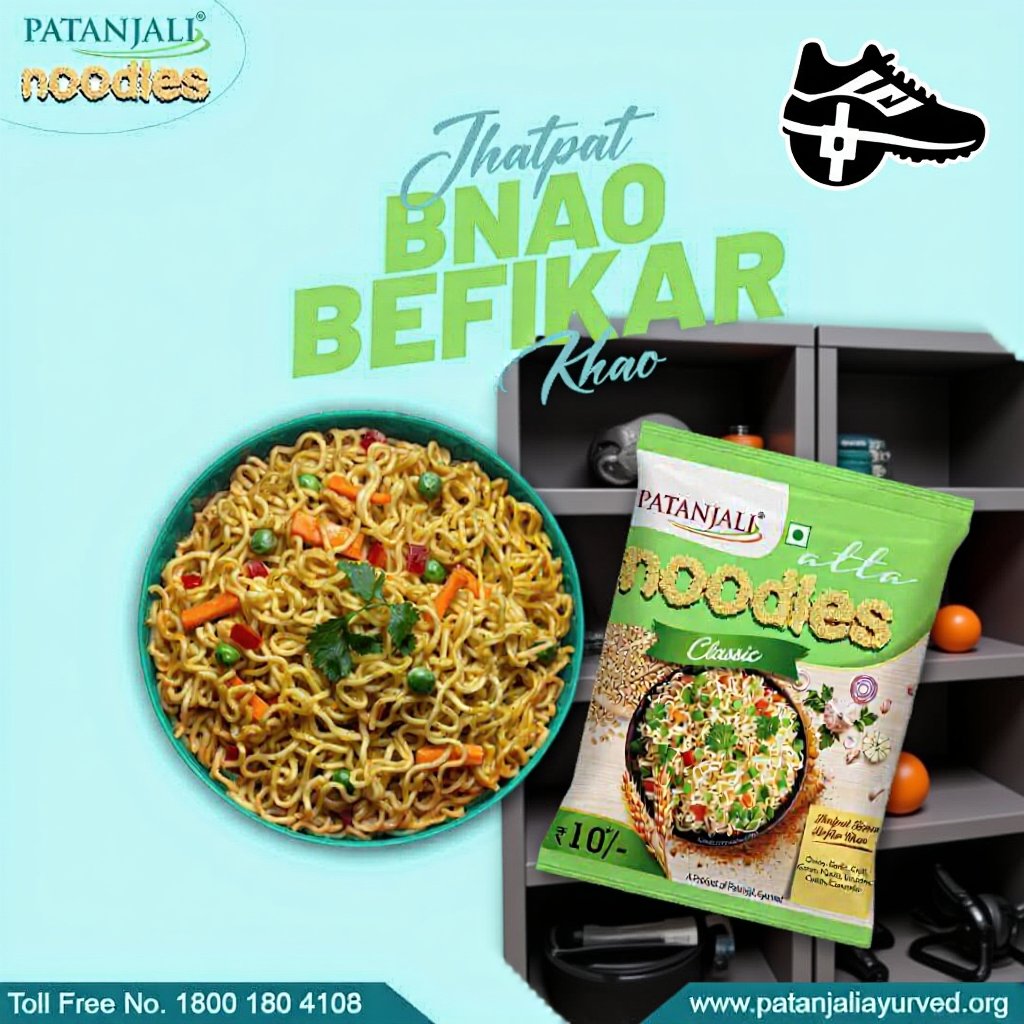} & 
\textbf{4:} Add a gym locker room background or gym setting behind the noodles bowl, without obscuring the food or packaging. \newline
\textcolor{red}{\fontsize{7pt}{8pt}\selectfont qwen-bbox (region2) + flux inpaint} \\

% --- Step 4 ---
\includegraphics[width=\linewidth, height=2.7cm, keepaspectratio]{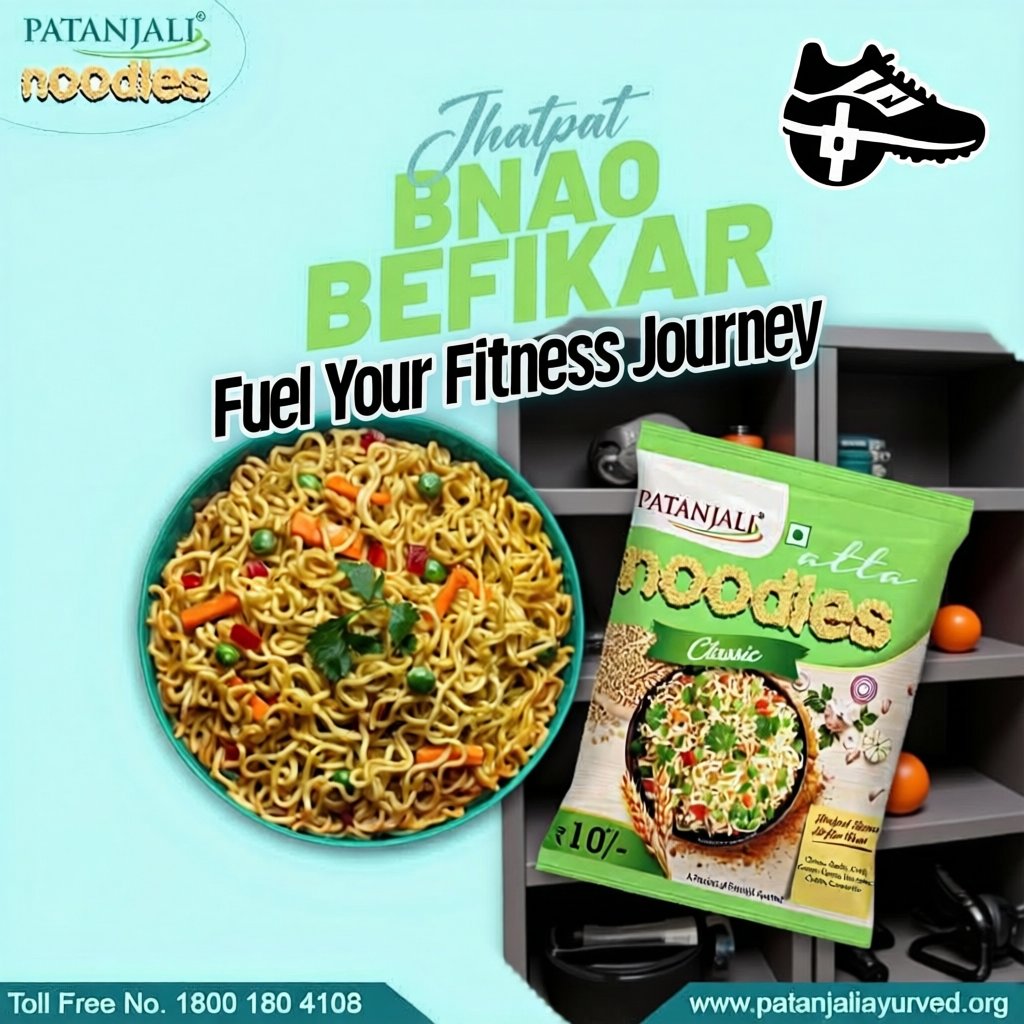} & 
\textbf{5:} Add the tagline 'Fuel Your Fitness Journey' in bold, modern font directly below the main slogan 'Jhatpat BNAO BEFIKAR Khao', keeping the existing font style and placement intact. \newline
\textcolor{red}{\fontsize{7pt}{8pt}\selectfont qwen image edit} \\

% --- Step 5 ---
\includegraphics[width=\linewidth, height=2.7cm, keepaspectratio]{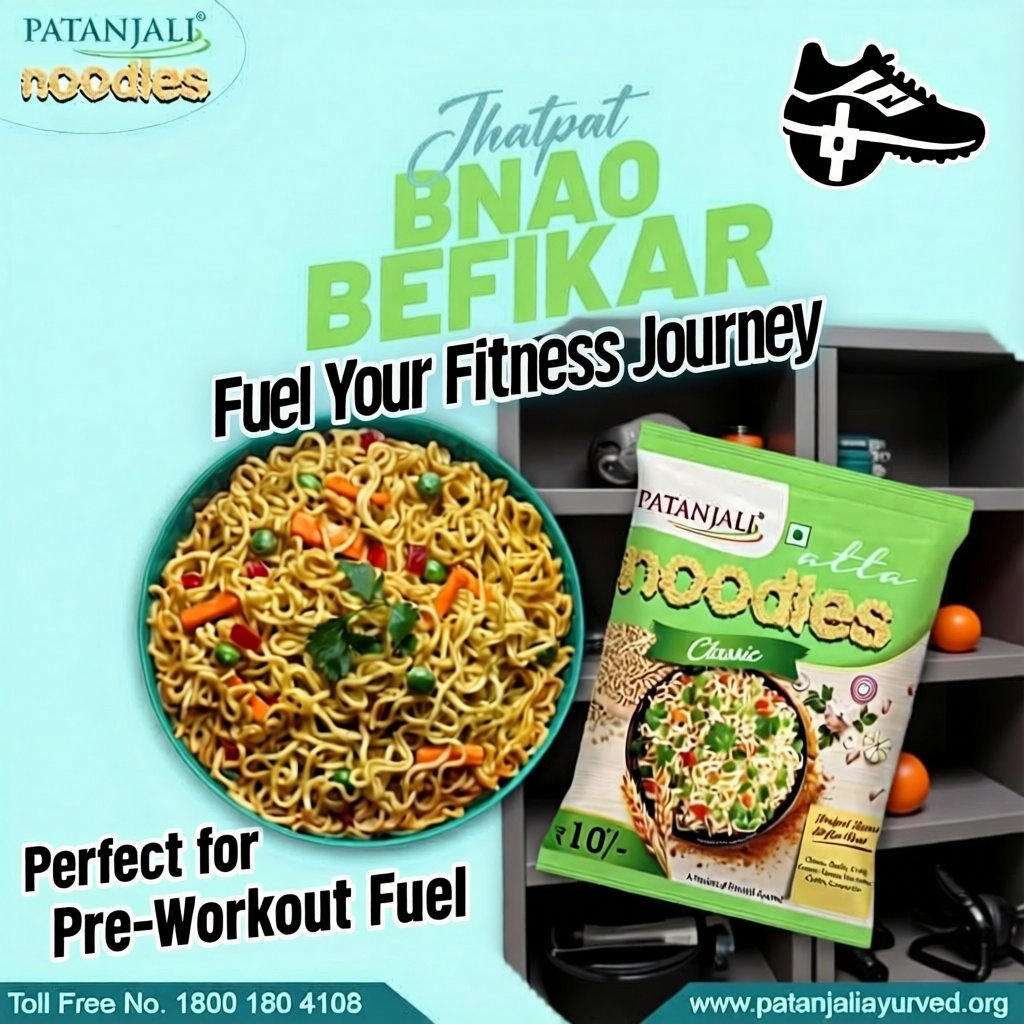} & 
\textbf{6:} Add the phrase 'Perfect for Pre-Workout Fuel' next to the noodle bowl on the left, using the same font as the existing slogan. \newline
\textcolor{red}{\fontsize{7pt}{8pt}\selectfont qwen image edit} \\

% --- Step 9 ---
\includegraphics[width=\linewidth, height=2.7cm, keepaspectratio]{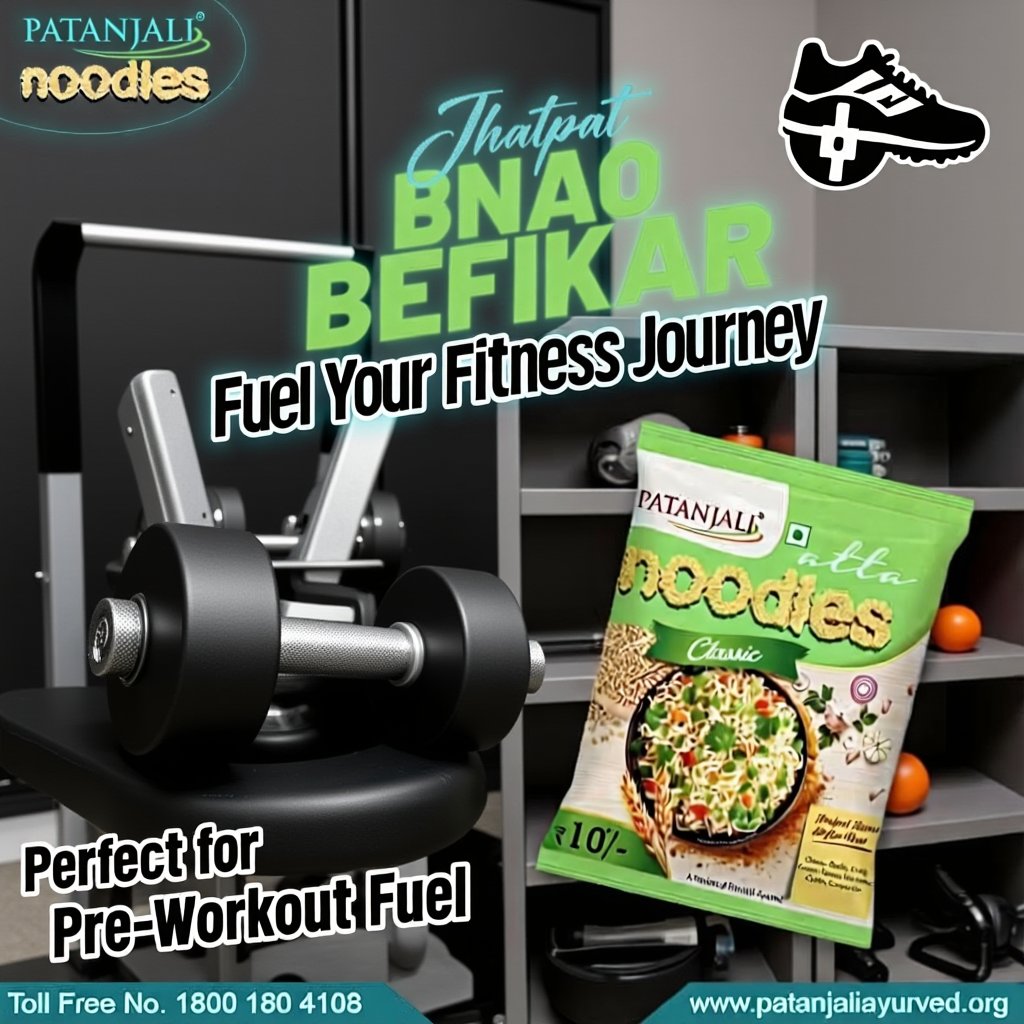} & 
\textbf{7:} Replace the existing decorative elements with a sports-themed background (e.g., crossed dumbbells or a 'Grab a Pair' prompt in front of dumbbells) to reinforce the fitness theme.' \newline
\textcolor{red}{\fontsize{7pt}{8pt}\selectfont qwen-layered (region1) + flux inpaint} \\ 

\bottomrule
\end{tabular}

\caption{Example editing sequence showcasing the transition from the original advertisement to a healthy lifestyle focused version. Our planner decomposes the task into steps, and the orchestrator carries out each step by selecting the appropriate editing tools.}
\label{tab:sustainable_car}
\end{table}

\section{Additional Experiments}

\subsection{Results on Image Editing Benchmarks}\label{app:comm-bmarks}
%\kr{Maybe instead of Common say Editing as common is little subjective}

In the main paper, we primarily reported results on a dataset consisting of complex multi-step advertisement editing tasks. In this section, we further evaluate our method on several widely used image editing benchmarks. Specifically, we consider the multi-turn editing setting of MagicBrush~\cite{zhang2024magicbrush} as well as the single-turn editing setting introduced in GEdit-Bench~\cite{liu2025step1x}.

\subsubsection{Performance on MagicBrush}

The multi-turn editing setting in MagicBrush consists of a sequence of edits (typically ranging from one to three) applied to a single image. The instructions in this benchmark specify direct edits, rather than requiring high-level planning. Therefore, in this experiment we employ only our learned orchestrator and do not use the planner. Through this evaluation, we examine whether the policy learned by our orchestrator—trained on advertisement editing tasks—generalizes to other commonly studied image editing scenarios.

\paragraph{Baselines}
We compare our method against several existing agentic image editing systems, including GenArtist \cite{wang2024genartist}, LayerCraft \cite{zhang2025layercraft}, and Talk2Image \cite{ma2025talk2image}. In addition to these agent-based pipelines, we also evaluate against widely used instruction-based editing models such as MagicBrush \cite{zhang2024magicbrush}, HIVE \cite{zhang2024hive}, and InstructPix2Pix \cite{brooks2023instructpix2pix}. 

Unlike our approach, these agentic baselines do not learn a tool-selection policy from experience; instead, they rely on prompt-engineered orchestration to select the appropriate actions. 

\begin{figure*}[t]
\centering
\includegraphics[width=\linewidth]{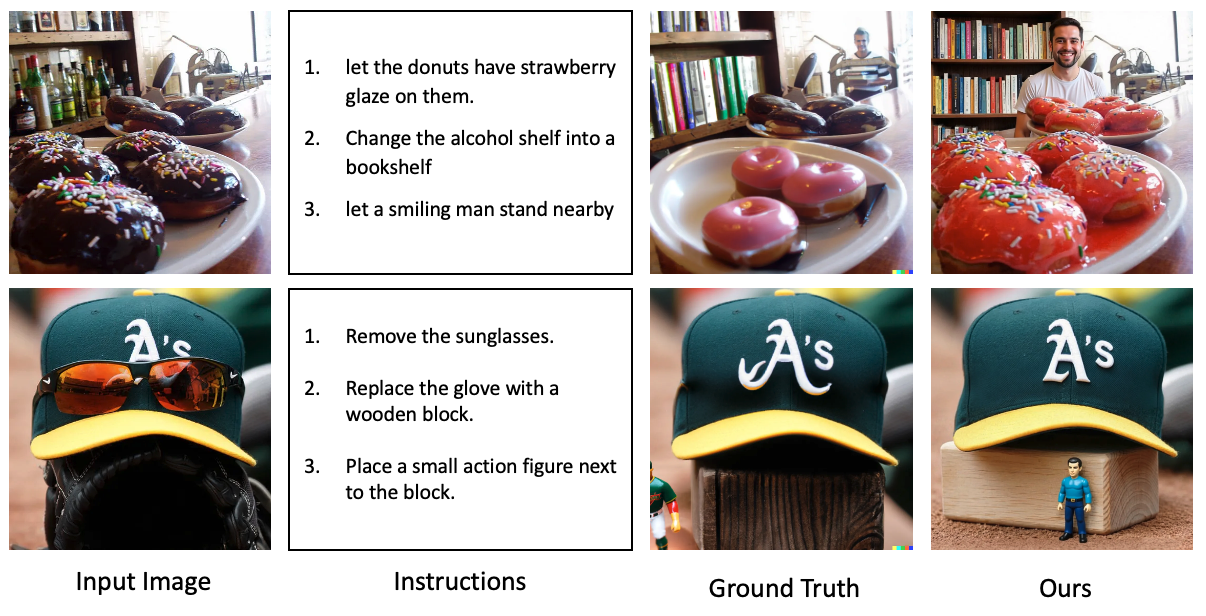}
\caption{
Examples from the MagicBrush benchmark. Each example shows the input image, the sequence of editing instructions, the provided ground-truth result, and the output produced by our method. While the benchmark provides a single ground-truth image for evaluation, many editing instructions can be satisfied in multiple visually valid ways. As illustrated here, our method produces edits that follow the instructions while maintaining visual coherence, but may differ from the specific appearance of the provided ground truth. This highlights a limitation of evaluation metrics that rely on similarity to a single reference image when assessing open-ended image editing tasks.
}
\label{fig:magicbrush_examples}
\end{figure*}

\begin{table}[t]
\centering
\small
\setlength{\tabcolsep}{6pt}
\renewcommand{\arraystretch}{1.15}
\begin{tabular}{lc}
\toprule
\textbf{Baseline} & \textbf{CLIP-T} \\
\midrule
\multicolumn{2}{l}{\textit{Instruction-based editing models}} \\
InstructPix2Pix & 0.2726 \\
\quad w/ MagicBrush & 0.2754 \\
HIVE & 0.2673 \\
\quad w/ MagicBrush & 0.2796 \\
\midrule
\multicolumn{2}{l}{\textit{Agentic pipelines}} \\
GenArtist & 0.3067 \\
LayerCraft & 0.3157 \\
Talk2Image & 0.3157 \\
\midrule
%Oracle (human-annotated images) & 0.3088 \\
\textbf{Ours (k=5)} & \textbf{0.3256} \\
\bottomrule
\end{tabular}
\caption{
Performance on the MagicBrush benchmark measured using CLIP-T. 
Our method outperforms prior agentic approaches (GenArtist, LayerCraft, and Talk2Image) as well as instruction-based editing approaches (InstructPix2Pix, HIVE, MagicBrush). 
%\yj{Remove this?  Interestingly, several baselines obtain higher text-alignment scores than the provided human-annotated ground-truth images. 
%This highlights a limitation of evaluation metrics that rely on similarity to a single reference image when assessing open-ended image editing tasks.}
}
\label{tab:magicbrush_results}
\end{table}

\paragraph{Evaluation}
The MagicBrush benchmark conventionally provides five evaluation metrics. CLIP-T measures text alignment by comparing the CLIP embedding of the generated image with the CLIP embedding of the target caption. The remaining metrics—L1 distance, L2 distance, CLIP-I, and DINO—measure similarity between the generated image and a ground-truth image created by human annotators, conditioned on a target mask. However, due to the open-ended nature of image editing tasks, the same instruction can often be satisfied in multiple visually valid ways. 

We attach illustrative examples in Fig.~\ref{fig:magicbrush_examples}. In both examples, the edits produced by our method correctly follow the sequence of instructions while preserving the overall scene structure. However, the resulting images differ from the single ground-truth image provided in the benchmark. 

For instance, in the first example, our method successfully applies strawberry glaze to the donuts, replaces the alcohol shelf with a bookshelf, and places a smiling person nearby. While the specific appearance of these elements differs from the ground truth (e.g., the exact pose of the person or the texture of the glaze), the requested transformations are clearly satisfied. Similarly, in the second example, our method correctly removes the sunglasses, replaces the glove with a wooden block, and adds a small action figure beside it. Although the precise visual realization differs from the reference image, the instruction is executed faithfully.

These examples highlight a limitation of metrics that rely on similarity to a single ground-truth image: multiple visually valid edits may satisfy the instruction, yet still receive a lower score due to differences in appearance. As a result, such metrics may penalize correct edits that deviate from the specific reference provided in the dataset. For this reason, we focus on measuring alignment with the textual description, as captured by the CLIP-T score.

\paragraph{Results}
We report the CLIP-T results in Table~\ref{tab:magicbrush_results}. Our method achieves the best performance, outperforming both prior agentic approaches and instruction-based editing models in terms of semantic alignment with the target instructions. The policy learned through experience enables our orchestrator to select the tools best suited for each task, resulting in minimal yet accurate edits to the original image. This highlights the advantage of a learning-based orchestration strategy, which allows the system to adapt tool usage based on experience rather than relying on heuristic, prompt-engineered orchestration strategies.

%\yj{Remove this? Furthermore, our earlier observation regarding the limitations of reference-based evaluation is supported by the fact that the human-annotated ground-truth images obtain lower CLIP-T scores than several of the evaluated methods.}

\subsubsection{Performance on GEdit}
In addition to multi-turn editing, we also test our orchestrator, on the GEdit benchmark \cite{liu2025step1x}. This benchmark contains a series of diverse edits including background changes, object-level modifications, text editing, etc. %\kr{Not sure if we should say complex as its direct edits, maybe diverse?}

\paragraph{Baselines}
We compare our method against a diverse set of instruction-guided image editing models, including InstructPix2Pix~\cite{brooks2023instructpix2pix}, MagicBrush~\cite{zhang2024magicbrush}, AnyEdit~\cite{yu2025anyedit}, OmniGen~\cite{xiao2025omnigen}, OmniGen2~\cite{wu2025omnigen2}, and Step1X-Edit~\cite{liu2025step1x} along with its improved version Step1X-Edit-v1.1. We also include several recent multimodal and image generation systems capable of performing instruction-based edits, including Qwen-Image-Edit~\cite{wu2025qwenimagetechnicalreport}, UniWorld-v1~\cite{lin2025uniworld}, Gemini~2.0~\cite{deepmind2025gemini2}, BAGEL~\cite{deng2025emerging}, FLUX.1~Kontext~\cite{labs2025flux1kontextflowmatching}, and GPT-Image-1~\cite{openai2025gptimage1}.

\paragraph{Evaluation}
We follow the evaluation protocol of GEdit, which uses VIEScore~\cite{ku2024viescore}. VIEScore employs a multimodal large language model to assess the edited images along two dimensions: Semantic Consistency (SC), which measures how well the output follows the instruction, and Perceptual Quality (PQ), which evaluates the visual fidelity of the edit. In addition, an overall score is reported as the geometric mean of these two metrics, i.e., $\sqrt{\text{SC} \times \text{PQ}}$. As a strong and reliable judge, GEdit uses GPT-4.1 for evaluation, and we adopt the same setting in our experiments.

\begin{table}[t]
\centering
\small
\setlength{\tabcolsep}{6pt}
\renewcommand{\arraystretch}{1.1}
\begin{tabular}{lccc}
\toprule
\textbf{Method} & \textbf{SC} & \textbf{PQ} & \textbf{Overall} \\
\midrule
AnyEdit & 3.053 & 5.882 & 2.854 \\
Instruct-Pix2Pix & 3.296 & 6.189 & 3.219 \\
MagicBrush & 4.517 & 6.371 & 4.185 \\
UniWorld-v1 & 4.93 & 7.43 & 4.85 \\
OmniGen & 5.879 & 5.871 & 5.005 \\
Gemini 2.0 (DeepMind) & 6.73 & 6.61 & 6.32 \\
OmniGen2 & 7.16 & 6.77 & 6.41 \\
Step1X-Edit & 7.131 & 6.998 & 6.444 \\
BAGEL & 7.36 & 6.83 & 6.52 \\
FLUX.1 Kontext [Pro] & 7.02 & 7.60 & 6.56 \\
Step1X-Edit-v1.1 & 7.658 & 7.354 & 6.969 \\
GPT Image 1 [High] & 7.85 & 7.62 & 7.53 \\
Qwen-Image Edit & 8.000 & 7.860 & 7.560 \\
\midrule
\textbf{Ours} & \textbf{8.153} & \textbf{8.030} & \textbf{7.604} \\
\bottomrule
\end{tabular}
\caption{Comprehensive results on the GEdit benchmark. SC denotes semantic consistency, PQ measures perceptual quality, and the overall score is the geometric mean ($\sqrt{\text{SC} \times \text{PQ}}$). Our method maintains the top position even when compared against the recent proprietary and open-source models.}
\label{tab:gedit_results}
\end{table}

\paragraph{Results}
We report the GEdit results in Table~\ref{tab:gedit_results}. Our method achieves the best performance across all metrics, outperforming existing baselines in terms of semantic consistency (SC), perceptual quality (PQ), and the overall score. These results demonstrate that the policy learned by our orchestrator generalizes effectively beyond the advertisement editing tasks considered in the main paper and performs well on more general image editing benchmarks.

\subsection{Are Checklist-Based Plans Better?}\label{supp:clist_comp}

In this section, we investigate whether checklist-based supervision leads to higher-quality plans compared to directly generating plans with a base model. While the final edited images produced by executing these plans provide one signal of performance, they do not directly indicate which plan reflects a deeper understanding of the task itself.

To study this, we compare two types of plans: (1) plans generated by our planner trained with checklist-based supervision, and (2) plans generated by a base Qwen3-VL model that is simply prompted to produce a plan for adapting the image, without being required to satisfy any explicit constraints.
\paragraph{Evaluation.}
Evaluating the quality of plans is inherently subjective. Therefore, we rely on a strong MLLM judge, Gemini-3-Pro~\cite{gemini3pro}, to compare pairs of plans and select the better one. To mitigate positional bias (e.g., a tendency to favor either the first or second option in pairwise comparisons), we evaluate each pair of plans twice: once with the original ordering and once with the order reversed. The final preference score is computed by averaging the outcomes across both evaluations.

\paragraph{Results.}
Table~\ref{tab:plan_pref} reports the pairwise preference results. Gemini prefers the checklist-based plan over the base model plan in \textbf{70.25\%} of comparisons. This result suggests that checklist supervision encourages the planner to produce more coherent and task-aware editing strategies.

\begin{table}[t]
\centering
\footnotesize
\setlength{\tabcolsep}{6pt}
\begin{tabular}{l c}
\toprule
\textbf{Comparison} & \textbf{Gemini Preference} \\
\midrule
Checklist Planner vs Base Qwen3-VL & \textbf{70.25\%} \\
\bottomrule
\end{tabular}
\caption{Pairwise plan preference measured using Gemini-3-Pro. Each pair of plans is evaluated twice with reversed ordering to mitigate positional bias.}
\label{tab:plan_pref}
\end{table}

\subsection{GenArtist Performance on Advertisement Editing}

In the main paper, we compare our method with several agentic baselines, including Qwen Image Edit and FLUX Kontext, with an external MLLM for planning. In this section, we additionally evaluate a widely used open-source agentic editing system, GenArtist~\cite{wang2024genartist} on our MadVerse image-based benchmark. Other recent agentic approaches such as X-Planner~\cite{yeh2025beyond} and MIRA~\cite{zeng2025mira} do not provide publicly available implementations, making direct comparisons difficult. We therefore focus on GenArtist as the strongest reproducible baseline and evaluate it under the same experimental setting described in Sec.~4.1. We report our results in Table \ref{tab:end_to_end_baselines}. We observe that GenArtist struggles to effectively execute the requested edits compared to our method, often resulting in significant degradation. We hypothesize that this may stem from certain tools being poorly suited for these tasks, as well as limitations in the orchestrator's policy.

\begin{table}[t]
\centering
\footnotesize
\setlength{\tabcolsep}{4pt}
\renewcommand{\arraystretch}{1.05}
\resizebox{\linewidth}{!}{
\begin{tabular}{l c c c}
\toprule
\textbf{Method} & \textbf{Instruction Following} & \textbf{Identity Preservation} & \textbf{Visual Quality} \\
\midrule
GenArtist & 1.252 & 1.007 & 1.660 \\
\textbf{Ours} & \textbf{4.196} & \textbf{3.155} & \textbf{2.525} \\
\bottomrule
\end{tabular}
}
\caption{\textbf{Comparison with the agentic editing baseline GenArtist.}
Scores are averaged Gemini-3-Pro evaluations measuring instruction following, identity preservation, and visual quality. GenArtist often introduces excessive degradation to the original image while attempting edits, resulting in low scores across all metrics. In contrast, our method performs the requested edits while maintaining substantially stronger identity preservation and overall visual quality.}
\label{tab:end_to_end_baselines}
\end{table}

\section{Implementation Details}
Our framework consists of three main components: (i) a \textbf{Planner} that decomposes high-level editing requests into a sequence of atomic operations, (ii) a set of \textbf{editing tools} that perform the underlying image transformations, and (iii) an \textbf{Orchestrator} that selects the appropriate tool and/or spatial region to execute each operation. In this section, we primarily focus on the editing tools and the orchestrator used during execution.

For each component, we describe how it is constructed, trained, and used during inference. Finally, since the editing tasks we consider are open-ended, we also describe the evaluation framework used to assess instruction following, identity preservation, and visual quality. Finally, we also provide more details on our inference algorithm.

\subsection{Editing Tools}\label{app:tool_desc}

Image editing tasks involve a wide range of transformations, from global changes such as modifying the background or color palette to localized edits such as replacing objects or modifying specific regions. No single model reliably supports all of these operations, motivating the use of multiple specialized editing tools.

We therefore employ a modular toolset consisting of three categories: \textit{analysis tools}, \textit{whole-image editing tools}, and \textit{region-level editing tools}. Analysis tools identify regions of interest (e.g., objects, layout elements, or semantic layers). Whole-image editing tools apply instruction-guided transformations to the entire image. Region-level editing tools instead operate on specific regions identified by the analysis tools, enabling more precise localized edits.

\subsubsection{Whole-Image Editing Models}

For global edits, we use two instruction-guided editing models: \textit{FLUX.1-Kontext-dev} and \textit{Qwen-Image-Edit-2511}. Both take an input image and a textual instruction and generate a modified image that reflects the requested change while preserving the overall structure of the original image.

\textbf{FLUX.1-Kontext-dev} and \textbf{Qwen-Image-Edit-2511} are image editing models that take an input image and a textual instruction and generate an edited image consistent with the requested modification. 

Both models provide strong instruction-following capabilities and produce high-quality edits, but they also exhibit certain weaknesses. In particular, since edits are conditioned on the full image, modifications are not strictly spatially constrained and may unintentionally affect regions unrelated to the intended change. To address this limitation, we additionally incorporate a region-level editing pipeline that first identifies relevant regions using analysis tools and then performs targeted modifications via diffusion-based inpainting.

\subsubsection{Analysis Tools}
These tools identify editable regions in the input image that can later be modified by region-level editing models. Because different editing tasks require different forms of spatial understanding, we employ multiple complementary region discovery mechanisms to detect relevant areas of the image.

\begin{figure}[ht!]
\centering
\includegraphics[width=0.8\linewidth]{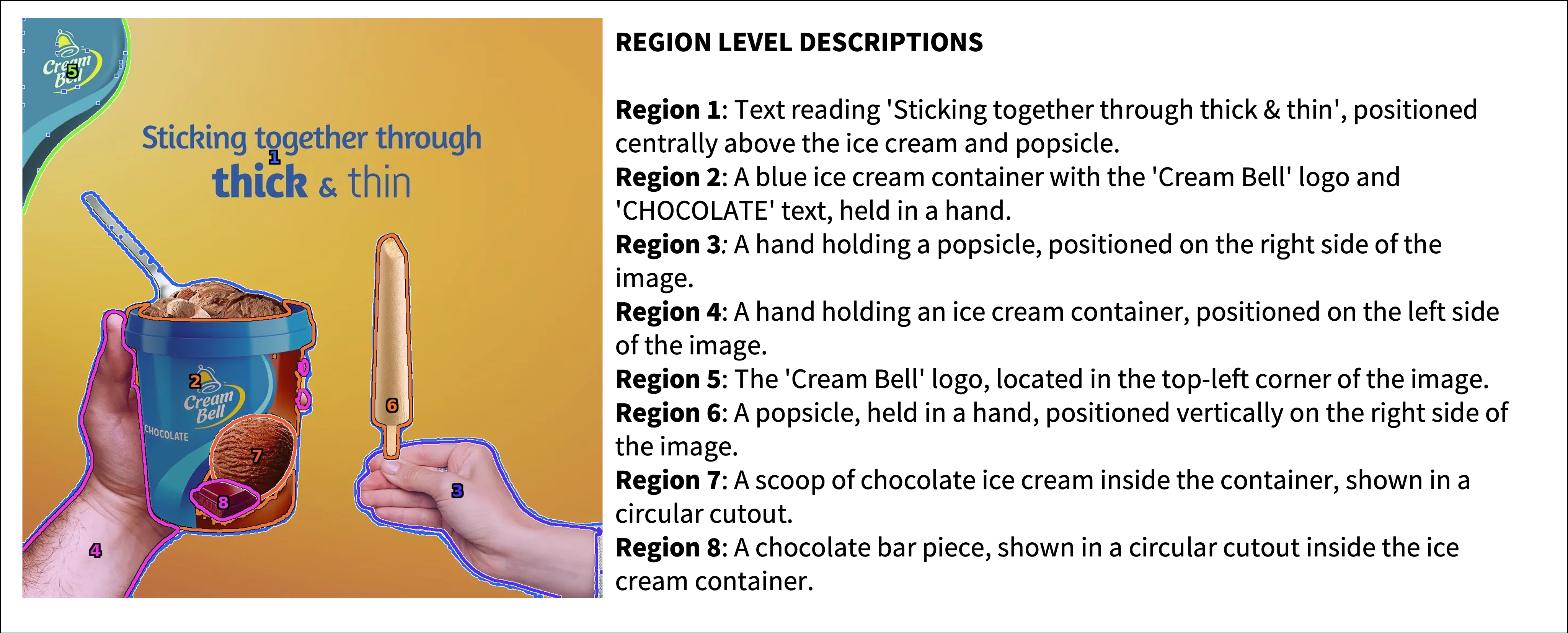}
\caption{
Region discovery using the SAM-2 + Qwen3-VL pipeline. SAM-2 first segments the image into candidate regions, which are visualized with numbered masks (left). The marked image is then provided to Qwen3-VL, which generates semantic descriptions for each region (right). These region indices and descriptions allow the system to reference specific parts of the image during subsequent editing steps.
}
\label{fig:sam_examples}
\end{figure}

(i) \textbf{SAM-2 + Qwen3-VL} performs semantic region discovery using a Set-of-Marks representation. We first apply SAM-2 to segment the image into candidate regions and overlay numbered markers on the resulting masks \cite{yang2023set}. The marked image is then provided to Qwen3-VL-8B, which generates a semantic description for each numbered region. This produces a structured mapping between region indices, masks, and textual descriptions, allowing the system to reference specific regions during editing. Fig.~\ref{fig:sam_examples} shows an example. We consider up to eight candidate regions, selected based on the largest area.

\begin{figure}[htbp]
\centering
\includegraphics[width=0.5\linewidth]{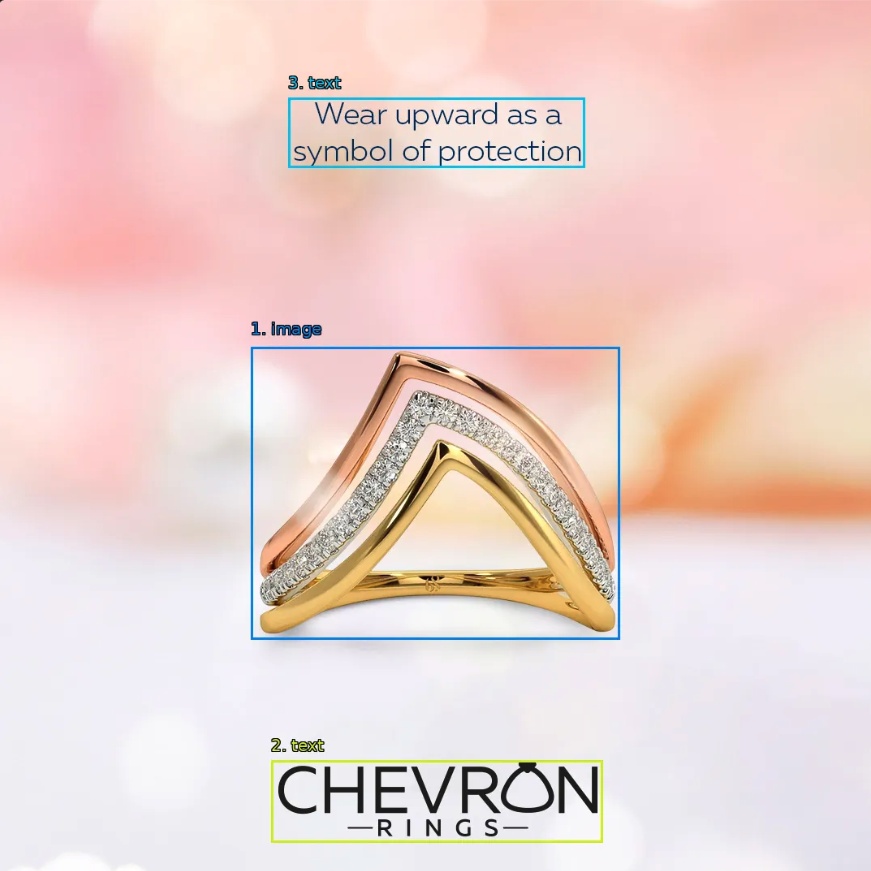}
\caption{
DeepSeek-OCR identifies textual and layout elements in the image by predicting bounding boxes around text regions and structured layout components. Each detected region is assigned an index, allowing the system to reference and modify specific textual elements during editing. 
}
\label{fig:ocr_examples}
\end{figure}

(ii) \textbf{DeepSeek-OCR} 
performs layout and text detection, identifying bounding boxes corresponding to textual elements and structured layout regions. Fig. \ref{fig:ocr_examples} shows an example. We consider up to 10 candidate regions, selected based on the largest area.

\begin{figure}[htbp]
\centering

\includegraphics[width=0.32\linewidth]{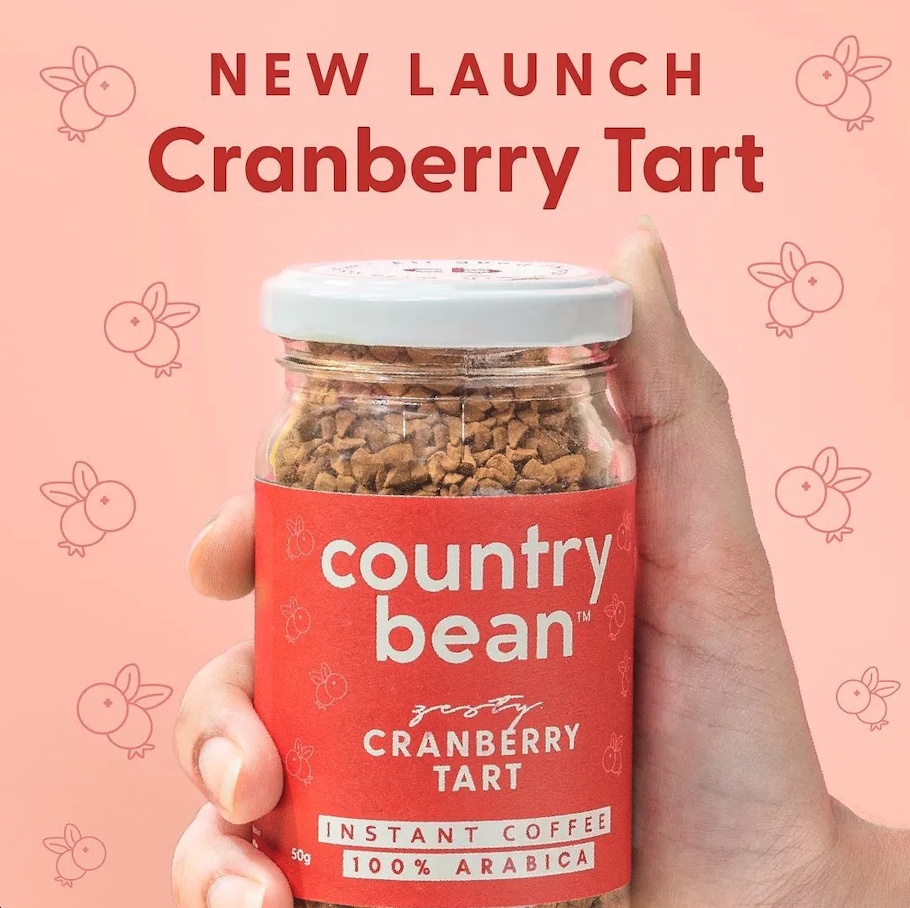}
\hfill
\includegraphics[width=0.32\linewidth]{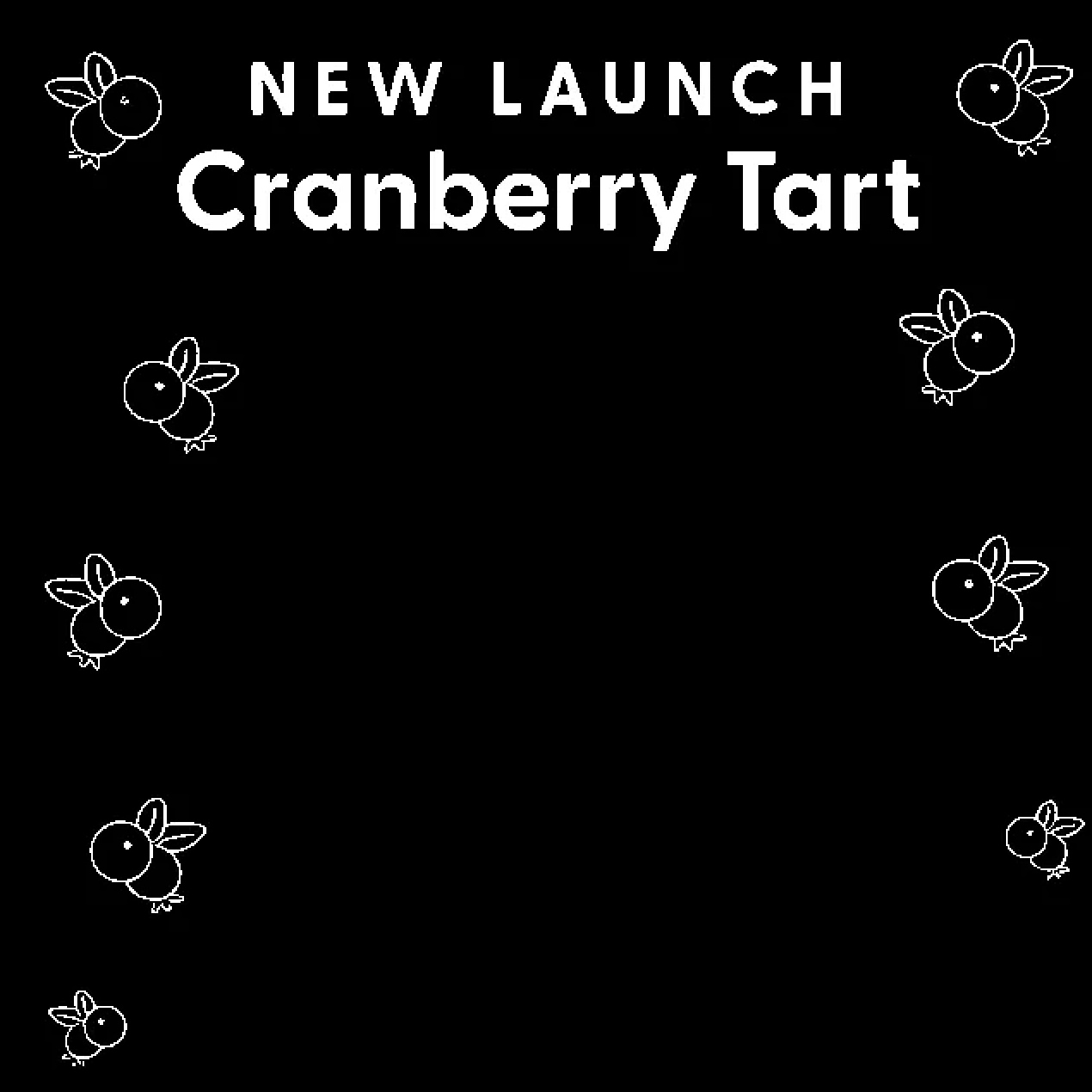}
\hfill
\includegraphics[width=0.32\linewidth]{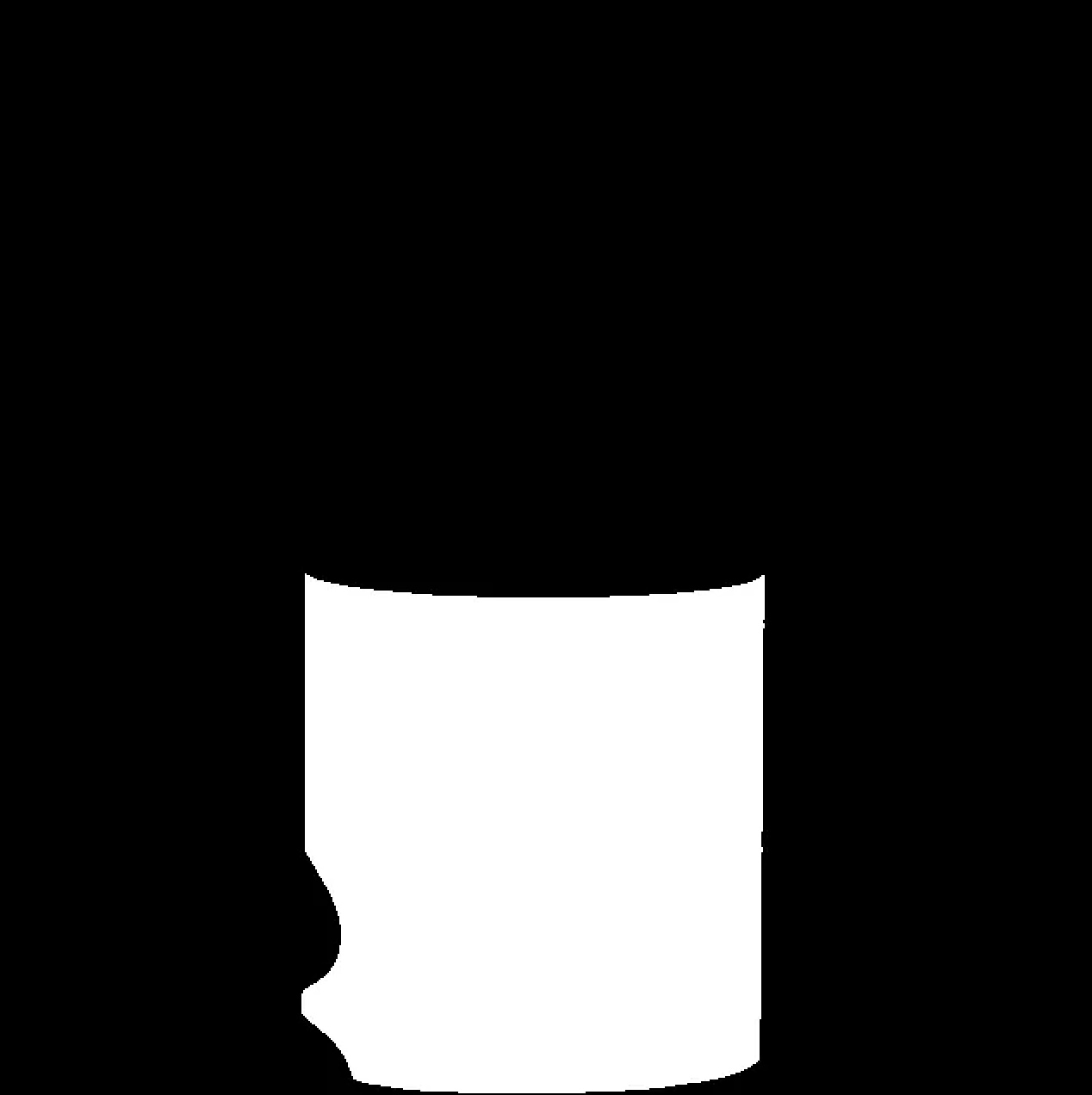}

\caption{
Layer-based region discovery using Qwen-Layered. \textbf{Left:} Original input image. \textbf{Middle and Right:} Example masks produced by Qwen-Layered that highlight different structural regions of the scene. These masks correspond to different alpha-composable layers that can be independently edited in subsequent region-level editing steps.
}

\label{fig:qwen_layered_examples}
\end{figure}

(iii) \textbf{Qwen-Layered} decomposes the image into a set of alpha-composable layers ordered from foreground to background, capturing larger structural components that may not correspond to individual objects. Each predicted alpha layer is converted into a binary mask by thresholding the alpha values at 128. These masks can then be used as candidate editable regions. Example layers are shown in Fig.~\ref{fig:qwen_layered_examples}. We consider four candidate regions.

\begin{figure}[htbp]
\centering
\includegraphics[width=\linewidth]{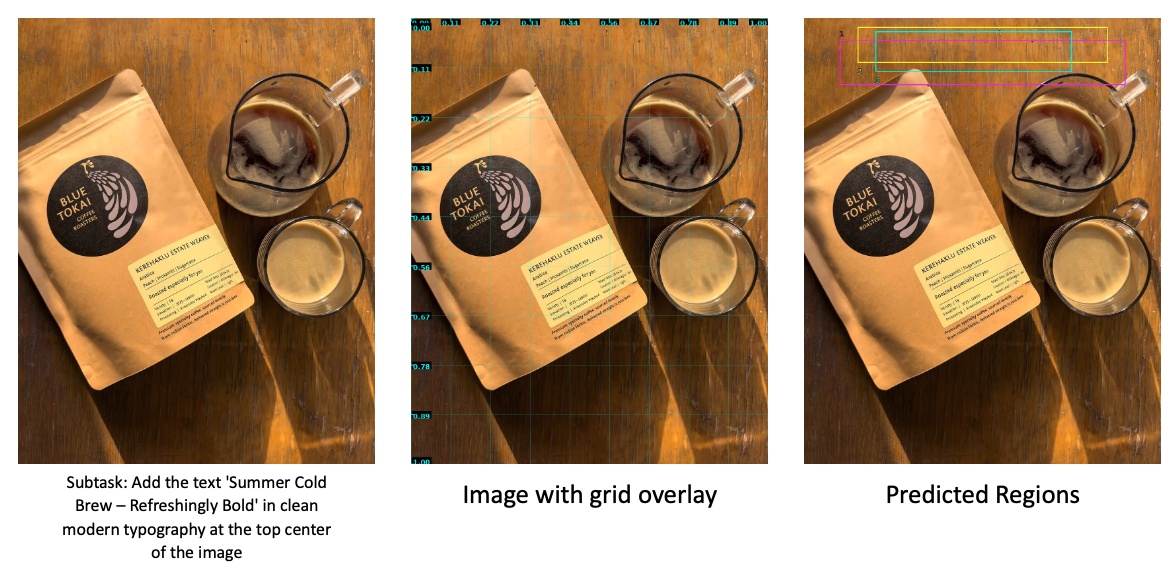}
\caption{
Instruction-guided region discovery using Qwen-BBox. \textbf{Left:} Original input image with the editing instruction. \textbf{Middle:} A grid overlay is added to the image, providing spatial reference cues that help the model reason about locations when predicting regions. \textbf{Right:} Candidate bounding boxes predicted by Qwen3-VL conditioned on the editing instruction. The model predicts three candidate regions that can be selected individually or combined during region-level editing.
}
\label{fig:qwen_bbox_examples}
\end{figure}

(iv) \textbf{Qwen-BBox} addresses a limitation of the previous analysis tools. While SAM-2, DeepSeek-OCR, and Qwen-Layered are effective at identifying existing objects, text, or structural components, they are not task-specific. As a result, they can fail to identify regions required for edits that involve adding new objects or modifying areas that do not correspond to clearly defined semantic entities. 

To address this, we use a Qwen3-VL-8B model to directly predict candidate regions conditioned on the editing instruction. However, we observed that predicting absolute bounding box coordinates directly is difficult for the model. Instead, we parameterize bounding boxes using normalized coordinates expressed as percentages of the image width and height. Even with normalized coordinates, the model struggles to localize regions reliably without visual references. Therefore, we overlay a grid on the image (Fig.~\ref{fig:qwen_bbox_examples}, middle), which serves as a visual prompt that helps the model reason about spatial locations.

Given the instruction and the grid-annotated image, the model predicts three candidate bounding boxes (Fig.~\ref{fig:qwen_bbox_examples}, right). During editing, the system may choose to operate on any of these individual regions or on the union of all predicted boxes.

\subsubsection{Region-Level Editing Tools}

For region-level editing, the orchestrator first selects a target region from the outputs of the analysis tools. Each analysis tool proposes candidate regions (e.g., segmentation masks or bounding boxes), from which the system chooses a single mask corresponding to the intended edit. In the case of bounding boxes we convert the box to a binary mask.

Given the selected mask, we use \textbf{FLUX-Kontext Inpaint}, a diffusion-based editor that performs instruction-guided modifications within the specified region. The model takes as input the image, the textual instruction, and the binary mask, and generates edits that are constrained to the selected area while preserving the surrounding content.

To provide the editing model with greater flexibility when modifying the target object, we dilate the predicted mask by 100 pixels before applying inpainting. Expanding the mask allows the model to adjust the size, shape, or surrounding context of the edited region, rather than being strictly constrained to the original mask boundary. The masked region is then edited according to the instruction, while pixels outside the mask remain unchanged.

This region-level editing mechanism enables precise localized modifications that are difficult to achieve with whole-image editing models alone.

\subsection{Orchestrator Details}\label{app:orchestrator-details}

Given the input image and editing instruction, the orchestrator selects the next action by producing a structured tool call. Each action is represented as a JSON object of the form
\[
\{\texttt{"tool"}: t,\ \texttt{"arguments"}: a\},
\]
where $t$ denotes the selected tool and $a$ contains any required parameters.

At each step, the orchestrator can choose between two types of actions: invoking an \textit{analysis tool} or directly applying a \textit{whole-image editing tool}. If a whole-image editing tool is selected, the model performs the requested modification across the entire image.

If an analysis tool is selected, the tool returns a set of candidate regions (e.g., segmentation masks or bounding boxes). These regions are then made available to the orchestrator, which subsequently selects one of them when invoking a region-level editing tool. In this case, the tool call includes both the editing instruction and the index of the region to be modified.

For example, a region-level edit is represented as
\begin{verbatim}
{
  "tool": "flux_inpaint",
  "arguments": {"region_number": 3}
}
\end{verbatim}

\subsubsection{Reward Model}

Since the editing tasks we consider are open-ended and do not have a single fixed ground-truth target, it is difficult to directly determine whether a generated edit successfully satisfies the instruction. Therefore, we require a signal that evaluates the quality of candidate edits and allows the system to identify which tool execution performs best. 

As discussed in the main paper, we pre-compute the outputs of candidate tool calls and use a reward model to score each resulting edit. The orchestrator is then trained to select the tool which yielded the highest reward.

Existing reward models such as EditScore \cite{luo2025editscore} and EditReward \cite{wu2025editreward} are primarily designed for natural image editing and do not fully capture the requirements of other kinds of images e.g., advertisement-style edits. Therefore, we design a custom evaluation rubric based on three criteria: \textit{instruction execution}, \textit{identity preservation}, and \textit{visual quality}. The rubric used by the evaluator is shown below.
\begin{tcolorbox}
\small

\textbf{CRITERION 1: INSTRUCTION EXECUTION }

\vspace{0.4em}

\textbf{Question:} \\
Did the edited image correctly and completely execute the requested instruction?

\vspace{0.4em}

\textbf{Scoring (0--5):}

\vspace{0.2em}

0: No attempt / wrong task: The instruction was not executed at all, or a completely unrelated edit was performed. The new advertisement does not actually advertise what the user intended to advertise.

\vspace{0.3em}

1: Attempted but ineffective: The model made a relevant attempt (e.g., tried to change the text or background as instructed), but it did not succeed in completely executing any part of the instruction.

\vspace{0.3em}

2: Partial but meaningful: At least one major part of the instruction is executed correctly, but other parts are missing or wrong.

\vspace{0.3em}

3: Mostly right, clearly flawed: Mostly correct, but some part of the instruction has been completely misinterpreted. Such as text is placed below instead of above.

\vspace{0.3em}

4: Complete with minor issues: All requested changes are present and recognizable, but with small imperfections—e.g., slightly off colors, minor typos ("Summmer" instead of "Summer"), or incorrect font/texture.

\vspace{0.3em}

5: Perfect execution: The instruction is executed exactly as requested with no errors or ambiguities.

\vspace{0.4em}

\textbf{IMPORTANT:} \\
Evaluate ONLY instruction execution here. Ignore aesthetics unless they prevent instruction execution.

\end{tcolorbox}

\begin{tcolorbox}
\small

\textbf{CRITERION 2: IDENTITY PRESERVATION}

\vspace{0.4em}

\textbf{Question:} \\
Did the edit preserve all parts of the original image that were NOT explicitly requested to change?

\vspace{0.4em}

\textbf{Scoring (0--5):}

\vspace{0.2em}

0: Catastrophic over-editing: The edit has completely destroyed the original layout and composition of the image, the new image does not preserve anything from the original image.

\vspace{0.3em}

1: Significant over-editing: The edited image retains some parts of the original image, but the new layout is still extremely different, such as unnecessary zooming into the product or removing the background.

\vspace{0.3em}

2: Moderate over-editing: The layout is mostly preserved, but few elements such as important objects, or critically important informative text boxes have been changed unnecessarily.

\vspace{0.3em}

3: Minor over-editing: Small, yet noticeable changes, such as removing less important object/label or adding a new one. Performing add instead of modify etc.

\vspace{0.3em}

4: Negligible over-editing: Subtle artifacts, such as different outline color, or minor color/texture changes.

\vspace{0.3em}

5: Perfect preservation: Nothing has been unnecessarily changed.

\end{tcolorbox}

\begin{tcolorbox}
\small

\textbf{CRITERION 3: VISUAL QUALITY}

\vspace{0.4em}

\textbf{Question:} \\
Does this image have good visual quality and aesthetics for use as an advertisement?

\vspace{0.4em}

\textbf{Scoring (0--5):}

\vspace{0.2em}

0: Aesthetically broken: All regions of the image are visually unpleasant or chaotic (e.g., severe artifacts, incoherent composition, unreadable text, clashing elements). It does not function as a usable advertisement.

\vspace{0.3em}

1: Poor aesthetics: Not all parts, but majority of the image has major quality issues—heavy blurring, over-saturation, or distortion that makes it look unprofessional.

\vspace{0.3em}

2: Weak aesthetics: Important regions suffer from noticeable quality issues—blurring, over-saturation, warping, or cluttered placement/overlapping text regions—that make the ad look amateurish. Otherwise the image looks acceptable.

\vspace{0.3em}

3: Acceptable aesthetics: The image is usable but has style issues—low-quality textures, fonts that look out of place, layout is not nice on the eye etc.

\vspace{0.3em}

4: Good aesthetics: Visually pleasing and well-composed overall, with only minor aesthetic imperfections, such as text which is slightly garbled or objects which slightly overlap.

\vspace{0.3em}

5: Excellent aesthetics: Highly polished, visually harmonious, and professionally composed. All elements work together cleanly and attractively as a high-quality advertisement. None of the components objectively detract from the overall visual appeal.

\end{tcolorbox}

To compute these scores, we use GPT-5 as a strong judge to evaluate the edited images according to the defined criteria. Finally, we aggregate the three criterion scores into a single scalar reward. Let $\text{IE}$, $\text{IP}$, and $\text{VQ}$ denote the scores for Instruction Execution, Identity Preservation, and Visual Quality respectively. Rather than summing the scores, which would allow a high score in one dimension to compensate for poor performance in another, we compute the geometric mean of the three scores:
\[
R = (\text{IE} \cdot \text{IP} \cdot \text{VQ})^{1/3}.
\]
This formulation encourages balanced performance across all criteria, since a low score in any single dimension significantly reduces the overall reward. Furthermore, many of the editing tools used in our pipeline (e.g., diffusion-based models) are inherently stochastic and may produce different outputs for the same input due to sampling noise. To reduce the variance, we generate two outputs for each tool invocation. The reward for a given tool selection is then computed as the average reward across these two outputs. Now we have a dataset where we have precomputed the tool output for every instruction given the original image on the training set and we have scored them, therefore we know which tools have the highest reward and can train on them.

\paragraph{Reward Model for Inference}
During inference, verification is critical to avoid selecting a poor editing action that could negatively affect subsequent steps in the editing process. Since the orchestrator considers multiple candidate tool executions, we require a reward model to evaluate the resulting edits and select the most desirable outcome.

Although the rubric described above can be evaluated using a strong closed-source judge, relying on such models during inference would be computationally expensive. To keep inference costs manageable, we instead distill this evaluation signal into a lightweight open-source reward model.

Specifically, we use \textit{Qwen3-VL-8B} as the backbone and train separate classification heads to predict the evaluation scores for each criterion. Each head predicts the score distribution for one of the three axes—Instruction Execution (IE), Identity Preservation (IP), and Visual Quality (VQ)—using supervision derived from the judge's scores for each tool execution. This allows the model to approximate the behavior of the original evaluator while remaining efficient enough for use during inference.

During inference, the reward model outputs logits over the possible score levels for each criterion. These logits are converted into probabilities using a softmax, and the expected score for each criterion is computed by weighting the possible score values by their predicted probabilities. Formally, if $z_{c,k}$ denotes the logit corresponding to score level $k \in \{1,\dots,5\}$ for criterion $c \in \{\text{IE}, \text{IP}, \text{VQ}\}$, the expected score is computed as
\[
\hat{s}_c = \sum_{k=0}^{5} k \cdot \mathrm{softmax}(z_{c})_k.
\]
The final reward is then obtained by aggregating the predicted scores using the same geometric mean formulation used during training:
\[
R = (\hat{s}_{\text{IE}} \cdot \hat{s}_{\text{IP}} \cdot \hat{s}_{\text{VQ}})^{1/3}.
\]

\subsection{Evaluation}\label{app:eval-setup}
Open-ended image editing, and in particular advertisement editing, has not been studied in detail by prior work. Therefore, we need to design a comprehensive measure of success. In the paper, we report results in the main section 4.1 of experiments as well as the ablations in section 4.2. In addition, we also compare the two plans in Section \ref{supp:clist_comp}. In this section, we provide the details of each of these evaluations.

\subsubsection{Whole-Edit Evaluation}\label{subsubsec:wholeediteval}
We aim to evaluate both the quality of the generated plan and the correctness of its execution by assessing the final edited image. To do so, we require a metric that captures both the reasoning and knowledge demonstrated by the planner when determining the required modifications, as well as the system's ability to faithfully execute those modifications and produce a high-quality image that remains semantically consistent with the user's request. 

To this end, we evaluate edits along three axes: \textit{Instruction Execution}, \textit{Identity Preservation}, and \textit{Visual Quality}. The judgement is conditioned on the initial image, the final edited image, and the high-level task description. Since we trained our model using GPT based rewards, in order to remove any potential bias in evaluation, we use Gemini-3-Pro \cite{gemini3pro} as a judge here. 

For \textit{Instruction Execution} we use the following rubric to score the edits:

\begin{tcolorbox}[colback=gray!5!white, colframe=black, boxrule=0.5pt,
                  left=2pt,right=2pt,top=2pt,bottom=2pt]
\scriptsize

\textbf{Instruction Execution (IE)}

\textbf{System Instruction:}
You are an expert evaluator for image editing quality with a focus on \textit{instruction execution}. Your task is to determine how accurately the requested edit has been carried out in the edited image.

\textbf{Evaluation Description:}
The model must do two things: (1) identify what should change in the original image to achieve the requested theme, and (2) correctly execute those changes. Scoring considers \textit{intent} (correct understanding of what must change), \textit{coverage} (whether sufficient changes are made to convey the theme), and \textit{execution} (whether those changes are implemented correctly). Compare the original image with the edited image and determine whether the intended transformation has been properly realized.

\textbf{Scoring Guide}

\textbf{1 — Very Poor:} Intent is incorrect. The instruction is misunderstood and the requested transformation is not meaningfully reflected (e.g., unrelated edits or changing the subject).

\textbf{2 — Below Average:} Intent is roughly correct but execution is poor, contradictory, or incorrect. Alternatively, execution may be reasonable but the edits are too minimal to convey the requested theme.

\textbf{3 — Acceptable:} Intent and coverage are reasonable, but execution contains several flaws. Multiple sub-edits may be inconsistent or visually incorrect.

\textbf{4 — Good:} Intent and coverage are good and execution is mostly correct. Minor imperfections may exist, but the overall transformation clearly reflects the instruction.

\textbf{5 — Excellent:} Intent is precise, coverage is complete, and execution is correct throughout. All necessary edits are applied and the resulting image fully conveys the requested transformation.

\end{tcolorbox}

With this rubric, we observe that the judge model evaluates the edit based on both the knowledge demonstrated as well as the success of execution.

In order to measure \textit{Identity Preservation}, we use the following rubric:
\begin{tcolorbox}[colback=gray!5!white, colframe=black, boxrule=0.5pt,
                  left=2pt,right=2pt,top=2pt,bottom=2pt]
\scriptsize

\textbf{Identity Preservation (IP)}

\textbf{System Instruction:}
You are an expert evaluator for image editing quality with a focus on \textit{identity preservation}. Your task is to evaluate how well the edited image preserves the core identity and non-target content of the input image.

\textbf{Evaluation Description:}
Identity preservation measures how well the edited image maintains the core identity and non-target content from the input image while allowing necessary attribute changes required by the editing instruction. Acceptable modifications may include changes to color, texture, background, style, or other properties when they are explicitly required to follow the instruction. Changes that go beyond what is necessary, introduce unintended drift, or alter unrelated elements of the image should reduce the score.

\textbf{Scoring Guide}

\textbf{1 — Very Poor:} Core identity is heavily altered or unrecognizable compared to the input image. Major non-target content is changed without justification. Extensive unintended modifications or over-editing are present.

\textbf{2 — Below Average:} Significant identity drift beyond what is required by the instruction. Multiple unnecessary changes to attributes, structure, or background. Edits exceed what is needed to satisfy the instruction.

\textbf{3 — Acceptable:} Core identity is mostly preserved and required attribute changes are applied, but there is noticeable drift in some structural elements or non-target regions beyond what the instruction necessitates.

\textbf{4 — Good:} Core identity and non-target content are well preserved. Attribute changes (e.g., color, texture, background) are applied only when justified by the instruction. Minor unintended drift or slight over-editing may be present.

\textbf{5 — Excellent:} Core identity is faithfully preserved relative to the input image. Only the modifications necessary to fulfill the instruction are applied, and unrelated content remains unchanged.

\end{tcolorbox}

And for visual quality, we only take in the final image as input (it is independent of either the instruction or the initial image) and evaluate the quality based on the following rubric:

\begin{tcolorbox}[colback=gray!5!white, colframe=black, boxrule=0.5pt,
                  left=2pt,right=2pt,top=2pt,bottom=2pt]
\scriptsize

\textbf{Visual Quality (VQ)}

\textbf{System Instruction:}
You are an expert evaluator for image editing quality with a focus on \textit{visual quality}. Your task is to evaluate the overall visual quality, coherence, and artifact level of the edited image.

\textbf{Evaluation Description:}
Visual quality measures how clearly and professionally the edited image presents information while remaining free from artifacts, distortions, or rendering errors. The image should appear polished and visually coherent, with edited regions integrated naturally with the rest of the image. Quality issues such as blurriness, broken structure, layout inconsistencies, or rendering artifacts should reduce the score.

\textbf{Scoring Guide}

\textbf{1 — Very Poor:} Severe artifacts, distortions, broken anatomy or structure, or major rendering failures. Strong blurriness. Text is completely garbled, blurred, or unreadable and conveys no information. Layout is unusable and information is effectively lost due to quality issues.

\textbf{2 — Below Average:} Obvious artifacts (e.g., warped regions, visible overlaps) affecting important objects or regions. Text has issues in every block and important information is not clearly conveyed. Typos may obscure meaning. Major objects may overlap incorrectly. Visual flaws significantly reduce usability.

\textbf{3 — Acceptable:} Generally usable image quality, but noticeable issues such as mild artifacts, slight blur, or weak integration between edited and original regions. Textures and text are mostly correct but may show minor distortions or typos. Color, font, or layout consistency may be poor. The image appears amateurish but information remains understandable.

\textbf{4 — Good:} Clean and coherent image with minor imperfections. Textures and edited regions blend well with the original content. Text is readable and correctly placed but may have minor stylistic issues (e.g., color inconsistencies or symmetry problems). Some imperfections in lighting, perspective, or design choices may remain.

\textbf{5 — Excellent:} High-fidelity, sharp, and visually coherent image. Lighting and shadows are natural and consistent. No visible artifacts, distortions, blur, or rendering errors. Edited content is seamlessly integrated with the original image. Text is crisp, readable, and free of overlap or distortion.

\end{tcolorbox}

This allows us to compare methods under a common high-level instruction while different techniques attempt to solve the task using different strategies.

\subsubsection{Plan-Conditioned Evaluation}\label{app:plan-cond-eval}

In the ablation studies presented in the main paper, we compared several components of our method while keeping the underlying plan fixed. The goal of this experiment is to determine whether, given the same plan, the combination of editing tools and a learned orchestration policy leads to improved performance.

Under this setting, evaluating instruction execution and identity preservation becomes more specific. In addition to assessing whether the overall task is addressed, we must also evaluate whether the resulting edits follow the plan itself. To enable this, we rely on a dense checklist derived from the plan.

Concretely, given the input image, the high-level instruction, and the multi-step plan, we use a strong MLLM to generate a dense checklist describing the criteria that the edited image should satisfy. The checklist enumerates specific concepts that should be modified, preserved, added, or removed, as well as important relationships between objects in the scene that should remain consistent with the plan.

During evaluation, we again use Gemini-3-Pro~\cite{gemini3pro}. The judge receives the original image, the high-level task, the edited image, and the generated checklist, and determines whether each checklist item is satisfied or not. The final score for an image is computed as the fraction of checklist items that are satisfied. We then average these scores across the dataset to obtain the reported results.

\paragraph{Checklist Generation}
In order to generate the checklist, we use the following system prompt: 

\begin{tcolorbox}[colback=gray!5!white, colframe=black, boxrule=0.5pt,
                  left=2pt,right=2pt,top=2pt,bottom=2pt]
\scriptsize

\textbf{Checklist Specification Generator Prompt}

\textbf{System Instruction:}
You are a specification generator for image editing tasks. Your goal is to produce a \textit{final-state constraint list} describing what must be true after all edits are completed.

\textbf{Inputs:}
\begin{itemize}
\item Input image
\item Task description
\item Step-by-step editing plan
\end{itemize}

\textbf{Output Format:}
Return a valid Python list of strings.  
Each string must represent exactly \textbf{one atomic final-state constraint}.  
Use the following prefixes:
\begin{itemize}
\item \textbf{Preserve:}
\item \textbf{Remove:}
\item \textbf{Replace:}
\item \textbf{Add:}
\item \textbf{Constraint:}
\end{itemize}

\textbf{Rules:}
\begin{itemize}
\item Analyze the input image to identify salient and structurally important elements.
\item If an important element is \textbf{not modified by the plan}, include a \textbf{Preserve} constraint.
\item For replacements, use the format: \texttt{Replace: <original> -> <new>}.
\item Use \textbf{Remove} only when an element must explicitly disappear.
\item Use \textbf{Add} only for new elements introduced by the plan (include placement or attributes when relevant).
\item Use \textbf{Constraint} only when the plan specifies positional, visibility, or relational requirements (e.g., top-right corner, left of product, below logo).
\item Do \textbf{not} include aesthetic or realism requirements unless explicitly stated in the plan.
\item If multiple valid realizations exist, include all possible outcomes.
\item Each list entry must represent exactly \textbf{one atomic execution condition}.
\item Be concise.
\item No explanations.
\item No markdown.
\item Output only the Python list.
\end{itemize}

\textbf{Example}

Image contains:
\begin{itemize}
\item Product pack
\item Brand logo
\item Blue background
\item Wooden surface
\end{itemize}

Plan:
\begin{enumerate}
\item Replace background with a festive Diwali scene
\item Add fireworks
\end{enumerate}

Output:

\begin{verbatim}
[
  "Preserve: product pack design",
  "Preserve: brand logo",
  "Preserve: wooden surface",
  "Replace: background -> Diwali festive scene",
  "Add: fireworks"
]
\end{verbatim}

\end{tcolorbox}

This prompt helps us to generate a dense checklist. Now we use this dense checklist to score the final edit. The system prompt for that is:

\begin{tcolorbox}[colback=gray!5!white, colframe=black, boxrule=0.5pt,
                  left=2pt,right=2pt,top=2pt,bottom=2pt]
\scriptsize

\textbf{Checklist Verification Prompt}

\textbf{System Instruction:}
You are a strict evaluator for image editing execution. Your task is to determine whether the edited image satisfies each constraint in a provided checklist.

\textbf{Inputs:}
\begin{itemize}
\item Original (input) image
\item Editing task / instruction
\item Checklist of final-state constraints
\item Edited (output) image
\end{itemize}

\textbf{Checklist Format:}
Each checklist item begins with one of the following prefixes:
\begin{itemize}
\item \textbf{Preserve:}
\item \textbf{Remove:}
\item \textbf{Replace:}
\item \textbf{Add:}
\item \textbf{Constraint:}
\end{itemize}

\textbf{Evaluation Rules:}
\begin{itemize}
\item \textbf{Preserve:} Mark Y if the referenced element remains present and meaningfully unchanged. Mark N if it is altered, removed, or obscured.
\item \textbf{Remove:} Mark Y if the element is no longer visible. Mark N if it still appears.
\item \textbf{Replace:} Mark Y only if the original element is gone and the specified new element is present. Otherwise mark N.
\item \textbf{Add:} Mark Y if the specified new element clearly exists. Mark N if it is missing or incorrect.
\item \textbf{Constraint:} Mark Y if the stated positional, relational, or visibility condition is satisfied. Otherwise mark N.
\end{itemize}

\textbf{General Guidelines:}
\begin{itemize}
\item Judge strictly based on visible evidence in the edited image.
\item Do not assume intent or infer missing details.
\item If the result is unclear or ambiguous, mark N.
\item Ignore aesthetic quality unless it is explicitly part of the constraint.
\item Provide only a very short justification phrase (e.g., ``logo still visible'', ``background changed'', ``color incorrect'').
\end{itemize}

\textbf{Output Format:}
Return a JSON array where each checklist item corresponds to one entry containing a short reason and a binary result.

\begin{verbatim}
[
  { "reason": "<short reason>", "result": "Y" },
  { "reason": "<short reason>", "result": "N" },
  ...
]
\end{verbatim}

Output only the JSON array and no additional text.

\end{tcolorbox}

For visual quality, we use the same evaluation as we discussed in section \ref{subsubsec:wholeediteval}. 

\subsubsection{Plan Comparison}
In order to perform the evaluation in section \ref{supp:clist_comp}. We use the following prompt:

\begin{tcolorbox}[colback=gray!5!white, colframe=black, boxrule=0.5pt,
                  left=2pt,right=2pt,top=2pt,bottom=2pt]
\scriptsize

\textbf{Plan Preference Evaluation Prompt}

\textbf{System Instruction:}
You are an expert evaluator of image editing plans. Your task is to compare two candidate plans and determine which one better solves the given editing task.

\textbf{Inputs:}
\begin{itemize}
\item Input image
\item Editing task / instruction
\item Plan A
\item Plan B
\end{itemize}

\textbf{Evaluation Criteria:}
Choose the plan that demonstrates stronger understanding of the image and the editing task.

\textbf{Strong plans typically:}
\begin{itemize}
\item correctly identify the important elements in the image that must be modified
\item recognize which elements should be preserved
\item demonstrate practical knowledge of how edits would realistically be executed
\item propose non-trivial yet practical changes that lead to a meaningful edit
\item provide clear and direct instructions with minimal ambiguity
\end{itemize}

\textbf{Weak plans often:}
\begin{itemize}
\item give generic instructions without understanding the structure of the image
\item rely on vague wording indicating limited knowledge of the task
\item suggest trivial changes that do not meaningfully improve the image
\item contain vague steps (e.g., ``update the text'' without specifying how)
\end{itemize}

\textbf{Plan Consistency Rules:}
\begin{itemize}
\item Plans are sequential; later steps operate on the result of earlier steps.
\item Early steps should not remove or modify elements in a way that makes later steps impossible.
\item Intermediate steps need not make sense in isolation, only within the overall plan.
\end{itemize}

\textbf{Output Format:}
First briefly explain your reasoning (2--4 sentences) referencing specific steps or phrases from the plans. Then output the final choice as JSON.

\begin{verbatim}
{
  "reasoning": "<brief reasoning>",
  "choice": "A" or "B"
}
\end{verbatim}

\end{tcolorbox}

\subsection{Inference through Tree Search}

Each tool invocation is represented using a structured format
\[
\{\texttt{"tool"}: t,\ \texttt{"arguments"}: a\},
\]
where $t$ denotes the tool name and $a$ specifies the corresponding arguments (e.g., region selection). This representation mirrors the execution interface of our editing framework, allowing predicted tool calls to be directly executed without requiring the model to generate intermediate code.

Importantly, this structured representation also defines a discrete and enumerable space of candidate actions. In typical tool-calling setups where the model generates free-form code or API calls, the output space is effectively unbounded, making it difficult to evaluate likelihoods over all possible actions. In contrast, our formulation specifies a finite set of candidate tool--region pairs $(a,r)\in\mathcal{C}$ for each sub-task. This allows us to explicitly evaluate how likely the orchestrator considers each candidate action.

Let $y_{a,r}=(y_1,\dots,y_L)$ denote the token sequence corresponding to a candidate tool invocation. For each candidate action we compute a length-normalized log-likelihood score under the orchestrator policy $\pi_\phi$:
\[
\mathrm{score}_{a,r} =
\frac{1}{L}\sum_{i=1}^{L}
\log \pi_\phi(y_i \mid \hat{x}, s_t, y_{<i}).
\]
This quantity corresponds to the average token log-likelihood (equivalently, the negative log-perplexity up to a constant) assigned by the orchestrator to the candidate action. Because the candidate action space is explicitly enumerated, the orchestrator can score and rank all possible tool selections in this manner.

These scores are then used to select the most promising candidate actions, which are executed and evaluated by the reward model as described in Algorithm~\ref{alg:orchestrator}.
\begin{algorithm}[t]
\caption{Reward-Guided Tool Selection}
\label{alg:orchestrator}

\KwIn{Input image $x$, plan $\mathcal{P}=\{s_1,\dots,s_T\}$, orchestrator $\pi_\phi$, reward model $R$, candidate tool--region set $\mathcal{C}$, beam size $K$}
\KwOut{Edited image $\hat{x}$}

$\hat{x} \leftarrow x$

\For{each sub-task $s_t \in \mathcal{P}$}{
    
    \For{each candidate $(a,r) \in \mathcal{C}$}{
        Let $y_{a,r}=(y_1,\dots,y_L)$ be the token sequence for $(a,r)$\;
        $\mathrm{score}_{a,r} \leftarrow 
        \frac{1}{L}\sum_{i=1}^{L}
        \log \pi_\phi(y_i \mid \hat{x}, s_t, y_{<i})$\;
    }

    $\mathcal{B}_t \leftarrow \mathrm{TopK}_{(a,r)\in\mathcal{C}}(\mathrm{score}_{a,r}, K)$\;

    \For{each $(a,r) \in \mathcal{B}_t$}{
        $\tilde{x}_{a,r} \leftarrow f_{a,r}(\hat{x})$\;
        $R_{a,r} \leftarrow R(\tilde{x}_{a,r}, x)$\;
    }

    $(a_t,r_t) \leftarrow \arg\max_{(a,r)\in\mathcal{B}_t} R_{a,r}$\;

    $\hat{x} \leftarrow f_{a_t,r_t}(\hat{x})$\;
}

%\textbf{Global refinement}\;

%$\hat{x} \leftarrow f_{\mathrm{ref}}(\hat{x})$\;

\Return $\hat{x}$

\end{algorithm}

\end{document}